\documentclass[a4paper,fleqn]{cas-dc}
\usepackage{graphicx}%
\usepackage{multirow}%
\usepackage{amsmath,amssymb,amsfonts}%
\usepackage{amsthm}%
\usepackage{mathrsfs}%
\usepackage[title]{appendix}%
\usepackage{xcolor}%
\usepackage{textcomp}%
\usepackage{manyfoot}%
\usepackage{rotating}
\usepackage{booktabs}%
\usepackage{algorithm}%
\usepackage{algorithmicx}%
\usepackage{algpseudocode}%
\usepackage{listings}%
\usepackage[table]{xcolor} 
\usepackage{bbding}
\usepackage[percent]{overpic}
\usepackage{pifont}
\usepackage{makecell}
\usepackage[most]{tcolorbox}
\usepackage{longtable}
\usepackage{threeparttable}
\definecolor{myRed}{RGB}{220,50,50}   
\definecolor{myBlue}{RGB}{135, 206, 235} 
\usepackage{subcaption}
\usepackage[percent]{overpic}
\usepackage[numbers,sort&compress]{natbib}
\def\tsc#1{\csdef{#1}{\textsc{\lowercase{#1}}\xspace}}
\tsc{WGM}
\tsc{QE}

\begin{document}
\let\WriteBookmarks\relax
\def\floatpagepagefraction{1}
\def\textpagefraction{.001}

\shorttitle{}    

\shortauthors{}  

\title [mode = title]{Standardizing Longitudinal Chest X-ray Report Evaluation via Large Language Model Annotation}  

%

\author[1]{Xinyi Wang}

\author[2]{Grazziela Figueredo}

\author[1]{Ruizhe Li}

\author[1]{Xin Chen}
\ead{Xin.Chen@nottingham.ac.uk}
\cormark[1]
\cortext[1]{Corresponding author}



\begin{abstract}
Longitudinal information in radiology reports refers to the sequential tracking of findings across multiple imaging-based examinations over time, which is crucial for monitoring disease progression and guiding clinical decisions. Many recent automated radiology report generation methods are designed to capture longitudinal information; however, validating their performance is challenging. There is no proper tool to consistently label temporal changes in both ground-truth and model-generated texts for meaningful comparisons. Large language models (LLMs) offer a promising annotation, as they are capable of capturing nuanced linguistic patterns and semantic similarities without extensive manual intervention. In this study, we therefore propose an LLM-based pipeline to automatically annotate longitudinal information in radiology reports. The pipeline first identifies sentences containing relevant information and then extracts disease progression information expressed in text. We evaluate and compare five mainstream LLMs on these two tasks using 500 manually annotated reports. Considering both efficiency and performance, Qwen2.5-32B was subsequently selected and used to annotate another 95,169 reports from the public MIMIC-CXR dataset. Our Qwen2.5-32B-annotated dataset provided us with a standardized benchmark for evaluating report generation models. Using this new benchmark, we assessed seven state-of-the-art report generation models on their ability to producing longitudinal information. Our LLM-based annotation methods outperform existing annotation solutions, improving F1 scores from 82.7\% to 94.0\% (+11.3\%) for longitudinal information detection and from 78.5\% to 83.8\% (+5.3\%) for disease progression information extraction. The LLMs show relatively stable gains in the no change category, but performs less effectively on the worsened category. In conclusion, this work demonstrates the potential of LLMs for efficient and effective medical report annotation and establishes a standardized evaluation framework for longitudinal chest X-ray report generation. The source code is available at
\url{https://github.com/wxinyi1996/Standardizing-Longitudinal-Chest-X-ray-Report-Evaluation-via-Large-Language-Model-Annotation.git}.
\end{abstract}






\maketitle

\section{Introduction}\label{sec1}

Radiological imaging is a key non-invasive tool used in routine clinical practice to assess patient condition and guide clinical decisions \cite{liao2023deep}. In clinical practice, serial images of a specific anatomical region are often acquired at multiple time points to monitor known conditions or detect new findings. Information derived from comparing such images over time is called longitudinal information. It is essential for tracking disease or injury progression and for evaluating treatment response \cite{kalil2016management, ImaGenomewu2021chest, li2020siamese,13zhang-etal-2025-libra,2hou2023recap}. 

\begin{figure*}[!htbp]
\centering
\includegraphics[width=1.8\columnwidth]{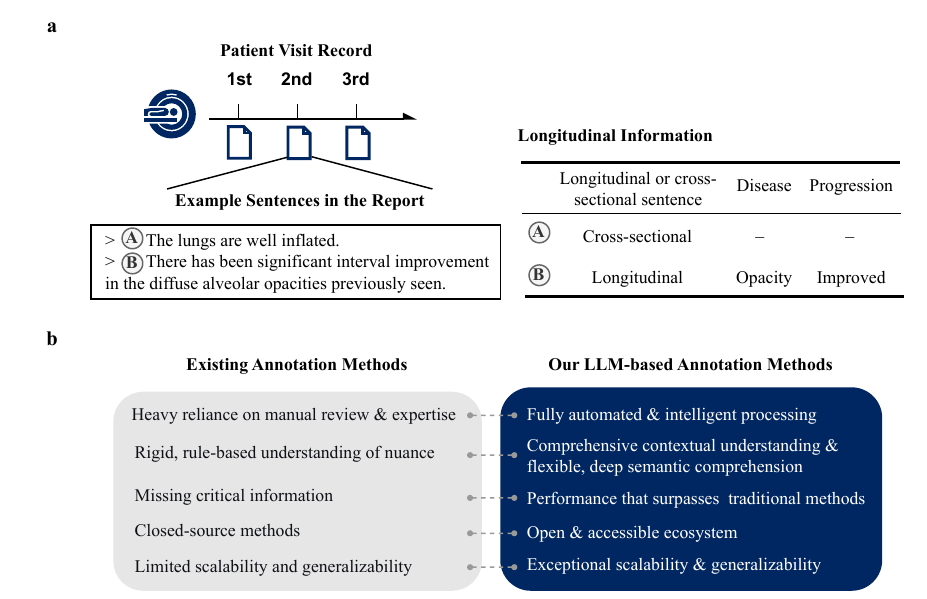}
\caption{\label{fig:Longitudinal_example} The longitudinal information in the report and the advantages of LLM-based annotation. a. Take sentences from the second-visit report of a patient to illustrate cross-sectional sentences and longitudinal information in free-text reports. b. Comparing existing longitudinal information annotation methods with large model-based annotation methods. } 
\end{figure*}

Radiologists often summarize longitudinal information in unstructured free-text reports. This leads to follow-up radiology reports containing two sentence types. Cross-sectional statements describe findings derived from a single time-point visit, while longitudinal sentences integrate information across sequential studies. Figure \ref{fig:Longitudinal_example}a illustrates a follow-up report that includes a cross-sectional sentence on lung inflation (i.e., sentence A) and a longitudinal sentence documenting the improvement in opacity compared to the initial study (i.e., sentence B).

Given the substantial workload associated with manual radiology report writing, the development of automatic report generation models has been widely researched \cite{wang2025survey}. Early report generation models primarily processed single time-point images, overlooking the sequential changes intrinsic to longitudinal image series \cite{wang2025survey,edirisinghe2025chest,r1hou2025mambagen,r3hou2023mkcl,r2hou2025recalibrated}. Recent automated models capture this longitudinal information by leveraging multiple sequential images or reports \cite{13zhang-etal-2025-libra,10nicolson2024longitudinal,4wang2024hergen,8dalla2023controllable,15zhou2024medversa,14bannur2024maira,1bannur2023learning,5liu2025enhanced,6wang2025llm,3zhu2023utilizing,2hou2023recap,7liu2025hc,14bannur2024maira,yang2025spatio,r4song2026mixture}. There is however a gap in the current literature, which is the lack of a standardized framework for assessing longitudinal report generation \cite{7liu2025hc}. This gap precludes systematic comparison of models and their constituent methodological components.

 Quantitative and reproducible assessment requires automated annotation tools that can reliably annotate changes over time in both ground-truth and model-generated reports for valid comparisons. Specifically, the coherency of longitudinal descriptions can be evaluated by extracting and comparing longitudinal sentences in clinical and generated reports, while the capturing of disease changes can be assessed by extracting disease progression labels. Such assessments are very common for cross-sectional content. For instance, prevalent Clinical Efficacy metrics employ automated labelers (e.g., CheXbert \cite{01smit2020chexbert}) to detect the presence of 14 specific diseases in both reference and generated reports for comparison \cite{wang2025survey}. However, the assessment of longitudinal information has received scant attention, largely because existing tools are inadequate for establishing a standardized evaluation framework. In addition, the benefits of an efficient longitudinal annotation tool are not limited to evaluation. It can help structure reports and provide annotated training data for deep learning models particularly in medical image analysis. 

 Existing longitudinal annotation tools mainly include MS-CXR-T \cite{1bannur2023mscxr}, ImaGenome silver \cite{ImaGenomewu2021chest}, MIMIC-Diff-VQA \cite{07hu2023expert}, and the Temporal Entity Matching (TEM) score \cite{1bannur2023learning}. Among them, MS-CXR-T, ImaGenome silver and MIMIC-Diff-VQA refer to the methods used in annotating these three datasets. MS-CXR-T is manually annotated, making it infeasible to keep pace with the newly generated reports continuously produced by report generation models. ImaGenome silver provides structured annotations for objects, attributes, and disease status changes in reports. However, it only releases the annotation results on MIMIC-CXR \cite{MIMICjohnson2019mimic} and does not open-source the underlying methodology. Consequently, the tool cannot be applied to annotate generated or other reports, which severely limits its usability. As for MIMIC-Diff-VQA \cite{07hu2023expert}, it presents information in unstructured question-answer pairs  (e.g., Question: “What has changed in the right lung area?” Answer: “The level of pleural effusion has changed from small to moderate”), which lacks the structured labels necessary for automated evaluation.

The TEM score \cite{1bannur2023learning} evaluates generated reports by matching temporal expressions against ground-truth reports. It extracts these expressions using both temporal lexicon matching and a deep-learning-based entity recognition model. A key limitation of this approach is its focus on extracting temporal expressions (TE) rather than summarizing disease progression (DP). While TE captures discrete descriptions of medical states (e.g., “nodule\_increased”), DP requires the analytical synthesis of such information to characterize clinical change (e.g., “nodule\_worsened”). Consequently, it penalizes semantically equivalent expressions that differ lexically (e.g., “enlarged” vs. “increased in size”). Furthermore, temporal keyword matching is too rigid to capture all longitudinal descriptions. For example, the changing positions of support devices across sequential imaging cannot be captured by the TEM lexicon.

Beyond the limitations in evaluating generated reports, current longitudinal information annotation approaches also face scalability and generalizability issues because they depend on either manual annotation \cite{1bannur2023mscxr} or lexicon-based rule-driven methods \cite{ImaGenomewu2021chest, 07hu2023expert, 1bannur2023learning}. Both manual annotation and the development of lexicons/rules require extensive specialized human effort from medically trained annotators, making the process costly, time-consuming, and difficult to adapt to other diseases. For instance, the ImaGenome silver constructed an initial vocabulary of 8,752 phrases for chest X-rays  \cite{05wu2021ai,ImaGenomewu2021chest}, which need to be refined through a semi-automatic filtering process.

Recent advances in large language models (LLMs) offer a scalable alternative for automated radiology report annotation \cite{flanders2025evolution}. Unlike previous methods, LLMs are capable of both summarizing disease progression and generalizing across diverse linguistic expressions without predefined lexicons or task-specific rules. These capabilities make LLMs particularly suitable for performing longitudinal annotation of reports and, subsequently, facilitating the evaluation of generated reports based on these annotations. Prior studies have applied LLMs to transfer free-text reports into structured templates for cross-sectional findings \cite{08woznicki2025automatic,011hein2025iterative,010grothey2025comprehensive} or specific disease-diagnosis tasks \cite{09kim2025benchmarking,012le2024performance,013nowak2025privacy}, but their potential for extracting longitudinal clinical information remains unexplored. 

To address the limited flexibility of existing annotation tools for longitudinal information and the shortcomings of the current evaluation frameworks for longitudinal chest X-ray report generation, this study investigates the following research questions:

RQ1: Can LLMs reliably identify longitudinal sentences and disease progression information in radiology reports compared with traditional annotation methods?

RQ2: Can an LLM-based annotation framework be used to construct a standardized benchmark and evaluation framework for assessing longitudinal information in chest X-ray report generation?

RQ3: How effectively do current automated report generation models capture longitudinal clinical information and disease progression when evaluated under the proposed framework?

Specifically, we propose an LLM-based framework for extracting longitudinal information from radiology reports. The framework comprises two main tasks: (1) longitudinal sentence identification, which determines whether a sentence contains longitudinal information, and (2) extracting disease progression information expressed in reports, which categorizes the change in the medical condition described in longitudinal sentences as ``improved”, ``no change”, or ``worsened”. Figure \ref{fig:Longitudinal_example}b illustrates the superiority of the LLM-based approach over existing methods, and Table \ref{table:motivation} presents a detailed comparison between our annotation method and existing annotation methods.

\begin{table*}[!htbp]
\centering
\renewcommand{\arraystretch}{1.2}
\caption{Our longitudinal information annotation method compared with other annotation methods. Form refers to how the method is presented: as a labeled dataset (D) or a general labeling method (M). TDM: traditional deep-learning model; TE: temporal expressions; DP: disease progression. Eval refers to whether this method can be used for generated report evaluation. }\label{table:motivation}
\begin{tabular}{%
    >{\raggedright\arraybackslash}m{3cm}%
    >{\centering\arraybackslash}m{2.5cm}%
    >{\centering\arraybackslash}m{2cm}%
    >{\centering\arraybackslash}m{1cm}%
    >{\centering\arraybackslash}m{0.8cm}%
    >{\centering\arraybackslash}m{0.8cm}%
    >{\centering\arraybackslash}m{1.3cm}%
    >{\centering\arraybackslash}m{0.8cm}%
}
\toprule
\multirow{2}{*}{Annotation method} 
& \multicolumn{4}{c}{Method category}
& \multirow{2}{*}{Form}
& \multirow{2}{*}{TE or DP}
& \multirow{2}{*}{Eval} \\
\cline{2-5}
& Lexicon and rules 
& Manual review
& TDM
& LLM
& 
& \\
\midrule
TEM & \ding{51} &--&\ding{51}&--& M & TE &\ding{51}\\
MIMIC-Diff-VQA & \ding{51} &--&--&--&D&TE &--\\
MS-CXR-T   & -- &\ding{51}&--&-- & D& DP&--\\
ImaGenome silver & \ding{51} &--&--&--& D& DP&--\\
Ours & -- &--&--&\ding{51}&M,D &DP&\ding{51}\\
\bottomrule
\end{tabular}
\end{table*}

To find the most suitable LLMs for this given task, we compare the performance of five mainstream LLMs on a subset of the Chest ImaGenome dataset \cite{ImaGenomewu2021chest}. The selected models vary in size and application scenario, including both general-purpose models and those fine-tuned for medical applications: MedGemma-27B \cite{sellergren2025medgemma}, MedResearcher-R1-32B \cite{yu2025medreseacher}, Qwen2.5-32B \cite{qwen2.5}, LLama3.3-70B \cite{dubey2024llama}, and Qwen2.5-72B \cite{qwen2.5}. The evaluation subset comprises 1,975 manually annotated sentences extracted from 500 reports. Subsequently, to establish an evaluation benchmark for report generation models, we annotate the follow-up reports in the MIMIC-CXR dataset \cite{MIMICjohnson2019mimic,johnson2019mimicjpg} using Qwen2.5-32B based on its annotation effectiveness and efficiency. We termed the resulting annotated dataset L-MIMIC. Details of the model selection and new benchmark dataset construction processes are provided in Appendix \ref{sup:LLM_select} and Section \ref{sec:data}, respectively.

Based on the LLM annotation pipeline, we developed an evaluation framework to assess whether report generation models can effectively capture longitudinal information. This evaluation examines performance along two key aspects: (1) the coherency of longitudinal descriptions, and (2) the capability to accurately capture disease progression. Finally, we evaluated seven mainstream report generation models (i.e., R2Gen \cite{chen2020generating}, MedVersa \cite{15zhou2024medversa}, L\_R2Gen \cite{3zhu2023utilizing}, MLRG \cite{5liu2025enhanced}, HC-LLM \cite{7liu2025hc}, Maira2 \cite{14bannur2024maira}, and Libra \cite{13zhang-etal-2025-libra}) on the L-MIMIC test dataset using the evaluation framework. 

In summary, our work makes three key contributions:

\begin{itemize}
    \item \textbf{First LLM-based annotation framework for longitudinal sentence identification and disease progression information extraction:} Distinct from the traditional annotation framework in Table \ref{table:motivation}, we leverage LLMs to automatically identify longitudinal content and disease progression in radiology reports. We compare five mainstream LLMs and show substantial improvements over traditional methods through comprehensive validation.

    \item \textbf{A standardized evaluation framework for longitudinal chest X-ray report generation:} We created a longitudinal annotation for the MIMIC-CXR dataset and named this newly annotated dataset L-MIMIC, which serves as a benchmark for evaluating how well report generation models capture longitudinal information. Based on the LLM-driven annotation framework, we also propose a set of metrics specifically designed for this assessment.

   \item \textbf{Evaluation of automated report generation models:} We assessed seven representative models to showcase the effectiveness of our proposed evaluation framework in identifying their strengths and limitations in capturing longitudinal information.
\end{itemize}

\section{Method}
This section describes the proposed pipeline for using LLMs to automatically annotate radiology reports, including LLM model selection, prompt design, and evaluation methodology. Based on a comparative evaluation, the best-performing LLM is selected to generate a large benchmark dataset, referred to as L-MIMIC. The second part of the method outlines the evaluation pipeline and metrics used to assess the performance of seven state-of-the-art radiology report generation models on the L-MIMIC dataset.

\subsection{LLM-based longitudinal annotation}

\subsubsection{Large language model selection}
Our model selection follows Kim et al. \cite{09kim2025benchmarking}, who benchmarked fifteen open-source LLMs for medical image diagnosis. We observe that models from the LLaMA, Qwen, and Gemma families perform strongly on medical diagnostic tasks. Accordingly, we select representative models from these families, covering both medical-specialized and general-purpose variants:
MedGemma-27B (\href{https://huggingface.co/google/medgemma-27b-text-it}{link}),
MedResearcher-R1-32B (\href{https://huggingface.co/AQ-MedAI/MedResearcher-R1-32B}{link}),
Qwen2.5-32B (\href{https://huggingface.co/Qwen/Qwen2.5-32B-Instruct}{link}),
Llama3.3-70B (\href{https://huggingface.co/meta-llama/Llama-3.3-70B-Instruct}{link}),
and Qwen2.5-72B (\href{https://huggingface.co/Qwen/Qwen2.5-72B-Instruct}{link}). MedGemma-27B and MedResearcher-R1-32B represent clinically adapted models trained on substantial medical corpora, whereas Qwen2.5-32B, Llama3.3-70B, and Qwen2.5-72B serve as state-of-the-art general LLM baselines. All models are released as open-source and freely available. All experiments used deterministic greedy decoding and were run on NVIDIA A100 GPUs.

\subsubsection{Report annotation pipeline and LLM prompt design}
\begin{figure*}[!htbp]
\centering
\includegraphics[width=2\columnwidth]{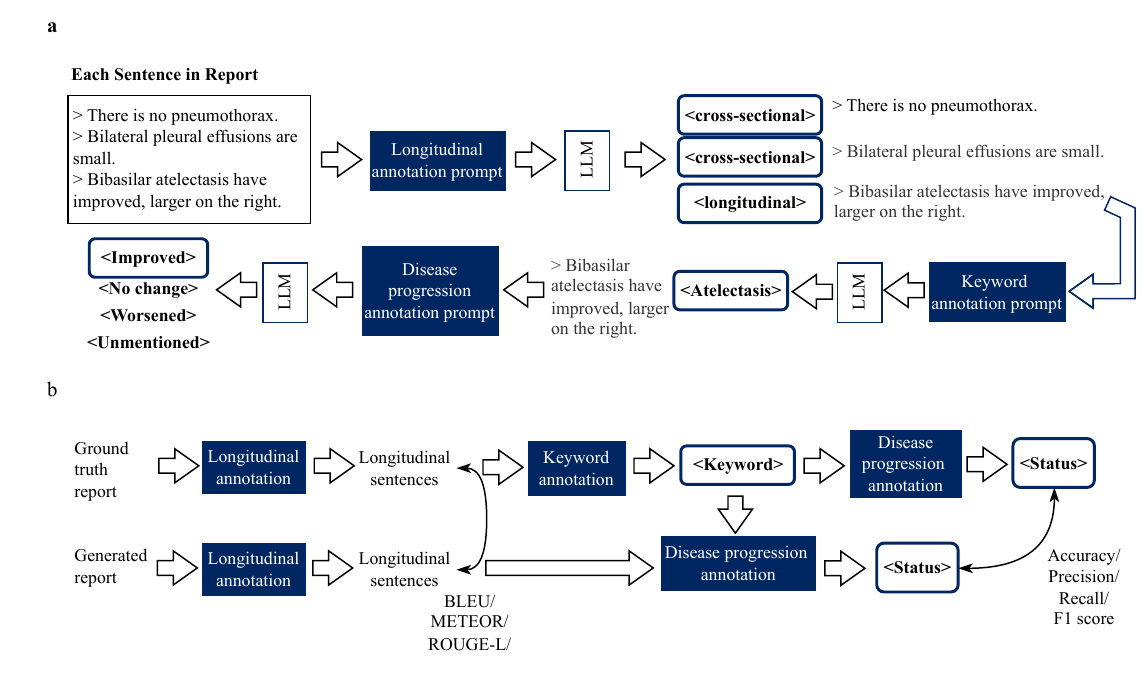}
\caption{\label{fig:LLM-Pipeline-onlylong} Annotation and evaluation pipeline. a. Longitudinal, keyword, and disease progression annotations. Longitudinal annotation identifies sentences containing longitudinal information; keyword annotation captures the main content; disease progression annotation captures changes in status (i.e., improved, no change, worsened, or unmentioned) related to the main content. b. An evaluation framework for the report generation model based on the proposed annotation method.} 
\end{figure*}

Reports were first segmented into sentences using Stanza \cite{qi2020stanza}. Each sentence was subsequently classified by an LLM into either a cross-sectional sentence, describing the patient’s static condition without temporal comparison, or a longitudinal sentence. For sentences identified as longitudinal, the model extracted disease-related keywords (e.g., atelectasis) and assigned a progression label. The design of the progression label follows previous work \cite{13zhang-etal-2025-libra,ImaGenomewu2021chest,1bannur2023mscxr,1bannur2023learning} and annotates three categories, i.e., improved, no change, and worsened. An additional unmentioned label is introduced to account for irrelevant content potentially generated by generative models. The overall processing workflow is summarized in Fig. \ref{fig:LLM-Pipeline-onlylong}a.

Following previous works \cite{zhuo2024prosa, 013nowak2025privacy}, prompts were designed by combining concise task descriptions with illustrative examples. The prompts used for longitudinal classification, keyword extraction, and progression labeling are provided below. Bold text indicates placeholders that are replaced with task-specific content. 

The example prompts below have been applied to all models in our framework, except that MedGemma requires “Please output \textless0\textgreater~or \textless1\textgreater.” at the end of the prompt when performing longitudinal annotation to enforce binary outputs. The disease list design is provided in Appendix \ref{disease_list}. Additionally, we investigated the sensitivity of LLMs to the number of prompt examples (denoted as ``x"-shot) provided for longitudinal and disease progression annotation. The detailed settings are provided in Section \ref{sec:prompt}.

\begin{quote}
\textit{``You are a medical AI assistant. Given a radiology report sentence \textbf{sentence}, determine if it compares the current image with prior studies (e.g., remain, compare, similar, stable, increased, still, new, again). If yes, return \textless1\textgreater. If no, return \textless0\textgreater. Examples: ``Cardiac and mediastinal silhouettes are stable.'' returns \textless1\textgreater, ``No larger pleural effusions.'' returns \textless1\textgreater,``Increased retrocardiac opacity may reflect atelectasis.'' returns \textless1\textgreater.''}
\end{quote}

\begin{quote}
\textit{``Which item in the list most related to the following sentence \textbf{sentence}? The list is \textbf{disease list}. Please reply with the answer enclosed in \textless\textgreater, for example \textless spinal fracture\textgreater. Return \textless support devices\textgreater~for any changes in support device positions.''}
\end{quote}

\begin{quote}
\textit{``The following sentence \textbf{sentence} describes the status of \textbf{disease} as \textless no change\textgreater (e.g., similar, unchanged), \textless improved\textgreater (e.g., resolved), or \textless worsened\textgreater. Return the answer in the form of \textless no change\textgreater, \textless improved\textgreater , or \textless worsened\textgreater. If this condition is not mentioned, return \textless unmentioned\textgreater. Note that if `change' or `new' is mentioned along with a newly appearing lesion, it usually implies worsening. Example: in cases of `pleural effusion', `New small bilateral pleural effusions.' means \textless worsened\textgreater. In cases of `low lung volumes', `lower' means \textless worsened\textgreater; `increased' means \textless improved\textgreater.'' }
\end{quote}

\subsubsection{Dataset and evaluation metrics \label{sec:data}}
The test set for assessing LLMs annotation performance was derived from ImaGenome \cite{wu2021chest}. Longitudinal annotations in ImaGenome were provided in two forms: manually curated (i.e., gold) and automatically generated (i.e., silver). Gold dataset annotations were performed by physicians with 2–10 years of clinical experience. Critical steps used dual annotation with consensus-based conflict resolution, while simpler tasks were single-annotated. Final outputs were reviewed to assess inter-annotator agreement \cite{wu2021chest}. In addition, the longitudinal annotations reflect only the findings documented in the reference radiology reports and may not represent the complete clinical reality, as radiology reports may omit stable, minor, or clinically less relevant abnormalities. Therefore, progression labels extracted by the LLM should be interpreted as reflecting documented radiological findings rather than absolute clinical truth.

The gold subset included only longitudinal sentences from second-visit reports of 500 patients, whereas the silver annotations covered the majority of sentences across all reports. We constructed a combined dataset of 1,975 sentences sourced from 500 reports. Specifically, this dataset comprises all sentences from the gold subset, augmented with non-longitudinal sentences sampled from the same reports in the silver annotations. It is worth noting that ImaGenome Gold is a randomly sampled subset of the MIMIC-CXR dataset \cite{wu2021chest}. This shared provenance supports its use for evaluating LLM-based annotation and its applicability to longitudinal report annotation in MIMIC-CXR. Furthermore, to demonstrate the representativeness of the 500 reports, we provide additional analyses of disease distribution in Appendix \ref{app:rep}. In addition, ImaGenome’s silver annotations, which were used as the baseline for comparison in this study, were also validated using the 500 gold-annotated samples. The silver annotations have been widely adopted in progression-related tasks \cite{ko2026temporal,yang2025tempa,chen2025coca}, which further supports the reliability and validity of the 500 gold annotations.

Model performance for both longitudinal annotation and disease-progression annotation was quantified using standard evaluation metrics, including accuracy, precision, recall, and the F1 score.  
Moreover, our evaluation of LLM-based annotation performance is conducted at the level of the full annotation framework (Figure \ref{fig:LLM-Pipeline-onlylong}), reflecting end-to-end usage scenarios in which errors introduced during the longitudinal annotation stage propagate to downstream disease annotation tasks. For zero-shot and few-shot prompting comparisons, as well as error analysis, we restrict disease progression annotation evaluation to sentences labeled as longitudinal sentences in the ground-truth annotations. This allows us to more precisely assess LLM performance on the disease progression annotation task in isolation.

\subsubsection{Decoding strategies and statistical analysis \label{sec:med_decoding}}
We conduct experiments using two decoding strategies: greedy decoding and stochastic sampling, to evaluate model performance under deterministic and stochastic settings, respectively. Unless otherwise specified, all experiments use greedy decoding following prior work \cite{09kim2025benchmarking} to ensure reproducibility.

For stochastic sampling, we apply temperature (0.7), top-p (0.9), and top-k (20) across all models. Since there is no standard consensus on hyperparameter settings for stochastic sampling in prior work, our hyperparameter settings are inspired by previous studies \cite{newSgilardi2023chatgpt,newSwang2022self,newSwang2024human}, which investigate the impact of sampling hyperparameters on performance. We adopt a hybrid top-k/top-p sampling strategy to reduce sampling variance while maintaining output diversity. Specifically, temperature rescales logits (dividing by T) before softmax, top-k restricts the candidate set to the K most probable tokens, and top-p (nucleus sampling) selects the smallest set of tokens whose cumulative probability exceeds P. Each model is evaluated over five independent runs with different random seeds. We report the mean and standard deviation of F1 scores across five runs for both tasks.

LLMs demonstrated notable improvements in the longitudinal annotation task (Table \ref{tab:results1}). To assess whether the performance gains of LLM-based annotation compared with traditional approaches were statistically significant in disease progression annotation, we conducted paired non-parametric bootstrap analyses for all evaluation metrics, including the estimation of bootstrap confidence intervals and paired bootstrap comparisons. The detailed definitions of these statistical analyses follow the same procedure described in Section \ref{sec:stat}.

\subsubsection{Comparison of zero-shot and few-shot prompting \label{sec:prompt}}
For prompt design, a key issue is whether few-shot or zero-shot prompting is more effective. Zero-shot uses only instructions without examples, while few-shot provides one or more examples for the LLMs to learn from. However, it remains unclear which approach performs better across tasks and instruction designs \cite{zhuo2024prosa, 013nowak2025privacy}.
Focusing on longitudinal annotation and disease progression annotation tasks, we evaluate model performance under both zero-shot and few-shot settings to assess robustness to prompt design and the impact of exemplar inclusion.

As shown in Figure \ref{fig:LLM-Pipeline-onlylong}, the framework consists of three prompts: longitudinal, keyword, and disease progression annotation prompts. Given the extensive disease list in the keyword prompt, we use a zero-shot setting without examples. For the longitudinal annotation prompt, examples can be added in two places: change indicators (e.g., ``remain'') and task examples. We evaluated 0, 3, and 5 task examples, and 0, 4, 8, and 12 change indicators. When varying one component, the other was held constant: 8 indicators were used for the task-example experiments, while 3 examples were used for the indicator experiments. For the disease progression annotation prompt, we experimented with 0, 2, and 4 task examples. Full prompt designs are provided in Appendix \ref{sec:zeroshot}.

\subsection{Evaluation of report generation models}

\subsubsection{L-MIMIC Dataset}
The L-MIMIC dataset was constructed by annotating radiology reports from the MIMIC-CXR database using Qwen2.5-32B, which was chosen for its favorable balance between performance and computational efficiency (Appendix \ref{sup:LLM_select}). Following Zhu et al. \cite{3zhu2023utilizing}, we selected reports containing a ``Findings" section from patients’ second or subsequent visits. Unlike them, we retained all images for each patient across different scan positions. In addition, we segmented the reports into "Indication," "Comparison," "Findings," "Impression," "Technique," "History," and "Examination" sections, using the official MIMIC-CXR code.

Each sentence in the “Findings” section was then annotated with Qwen2.5-32B to identify longitudinal information, relevant clinical keywords, and indications of disease progression. Overall, the L-MIMIC dataset consists of 73,093 training, 588 validation, and 1,863 test reports, which can be used as a benchmark for method comparison. 

\subsubsection{Radiology report generation models}

We assessed seven radiology report generation models, including two baselines: R2Gen, representing conventional deep-learning approaches, and MedVersa, representing large-scale models ($\geq$7B parameters). Neither baseline incorporates prior clinical history. The other five models—L\_R2Gen, MLRG, HC-LLM, Maira2 and Libra—are designed to exploit patients’ historical information to generate longitudinally informed reports. Among them, L\_R2Gen and MLRG do not incorporate LLMs, whereas HC-LLM includes a frozen LLM; all remaining models are LLM-based. This set comprehensively spans models of varying scales, architectures, and capacities to leverage longitudinal clinical data. Full model configurations are provided in Appendix \ref{model_config}.

\subsubsection{Evaluation pipeline and metrics}
Figure \ref{fig:LLM-Pipeline-onlylong}b summarizes the evaluation framework for assessing whether the report-generation models faithfully capture longitudinal disease progression. Using the generated L-MIMIC dataset as an example, ground-truth reports were first annotated with longitudinal sentences, disease-specific keywords, and progression labels. Generated reports were then processed to extract longitudinal sentences, which were matched to their ground-truth counterparts. Linguistic similarity was quantified using standard natural-language metrics \cite{wang2025survey}—BLEU-1–4, ROUGE-L, and METEOR—to evaluate the fluency and lexical correspondence of the generated longitudinal descriptions. 

For each disease instance with a reference progression label, progression status was inferred from the generated longitudinal sentences. Notably, progression labels excluded support devices. Given that changes in support devices are predominantly positional, their progression trends (improved, worsened or no change) cannot be reliably classified, and no corresponding annotations were provided by ImaGenome. In addition, as illustrated in Figure \ref{fig:LLM-Pipeline-onlylong}a, the annotation pipeline includes an additional “unmentioned” category to prevent the model from guessing when the text lacks relevant evidence. During evaluation, the predictions assigned to this category are post-processed as “disease absent”. If the gold standard labels for these instances are marked as certain disease progressions, these post-processed predictions are counted as incorrect. Table \ref{tab:D_error} in Appendix \ref{app:error_example} presents an example. Model performance in predicting progression was then evaluated using accuracy, precision, recall, and F1-score against the ground-truth annotations.

\subsubsection{Error analysis of LLM-based annotation}

Annotation errors generated by automated systems may propagate bias to downstream tasks, particularly evaluation. To better understand these effects, we analyze the errors made by Qwen2.5-32B and assess their impact.

Klie et al. \cite{klie2023annotation} compared noisy and human-corrected annotations, identifying four error categories: true errors, inconsistencies, incorrect corrections, and hallucinated entities. As part of their noisy annotations originated from single-annotator labeling, which shares characteristics with LLM-generated annotations, we adopted this taxonomy as the basis of our error analysis. We further refined the category definitions to better capture the characteristics of longitudinal and disease progression annotation tasks. A more detailed taxonomy of each error type and illustrative examples are presented in Figure \ref{fig:error}, while concrete examples are provided in Appendix \ref{app:error_example}. We manually examined all erroneous annotations in the 1,975-sentence test set across both tasks, assigned each error to a category and analyzed its impact. See Section \ref{sec:error_ana} for details on the causes and impacts of errors.

\begin{figure*}[!htbp]
\centering
\includegraphics[width=2\columnwidth]{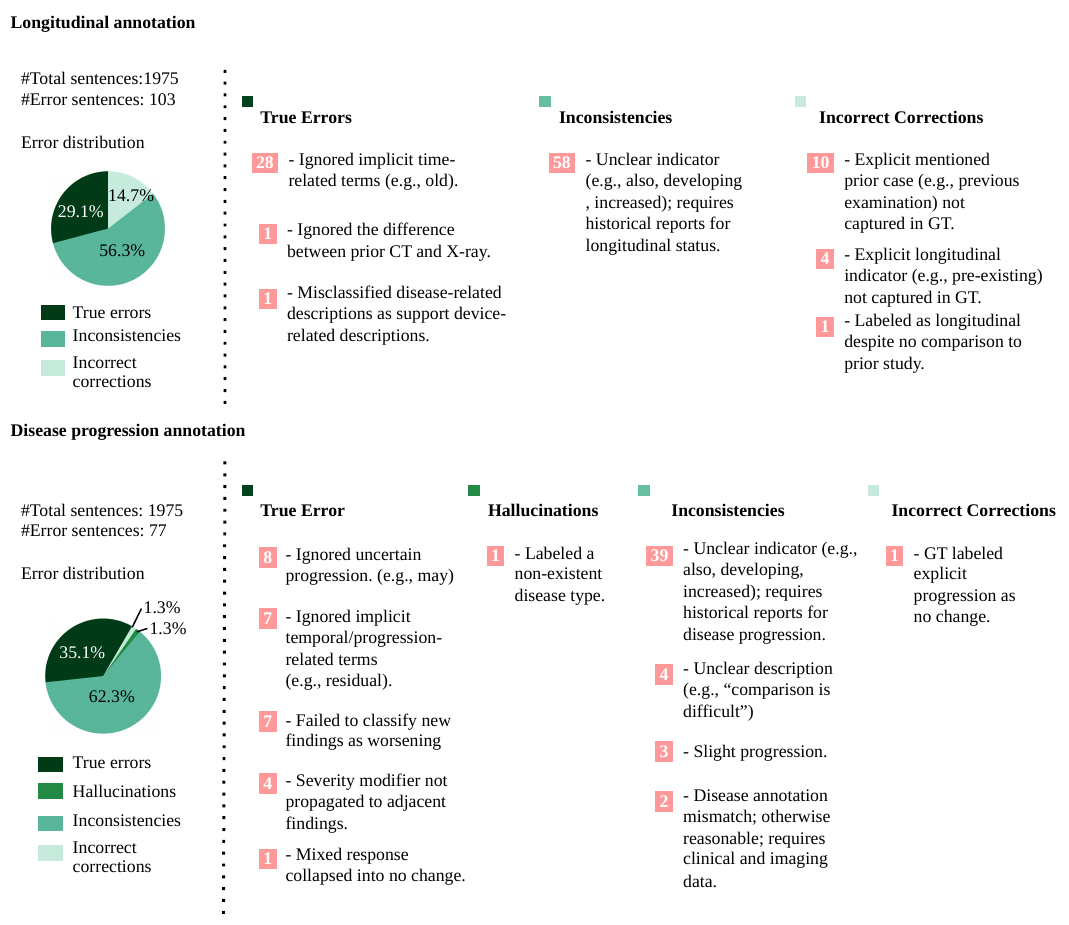}
\caption{\label{fig:error} Error analysis of Qwen2.5-32B in longitudinal annotation and disease progression annotation tasks based on the 500 reports (1975 sentences) with gold annotation. The numbers in red indicate the number of errors.}
\end{figure*}

Specifically, for longitudinal annotation task, as this is a binary classification task, hallucination errors are not applicable. The other error types are defined as follows:
\begin{itemize}
    \item True errors fell into two categories. The first included cases with explicit comparisons to prior examinations that were incorrectly judged by the model. The second included cases where comparisons were implicit and inferred from clinical wording referring to prior findings.
    \item Inconsistency errors occur when similar or identical expressions receive different GT annotations because they are interpreted in the context of different findings described in patients' prior reports. This category can be confused with the second true-error category; therefore, we apply two strict criteria: (1) similar expressions must be annotated inconsistently within the gold-standard dataset, and (2) the expression must not contain an explicit longitudinal cue.
    \item Incorrect corrections refer to clear annotation errors in the manual labels. As the ImaGenome gold dataset was built through inter-annotator agreement, such cases are identified only when strong evidence is available, e.g., explicit references to prior reports or clear comparative expressions.
\end{itemize}

For disease progression, the error categories were defined similarly to the categories used for longitudinal annotation task. Newly observed notable true-error cases included: (1) failures to recognize newly emerging diseases as ``worsened'' progression; and (2) difficulties in interpreting modifiers in complex sentences. Newly observed inconsistency errors involved ambiguous wording and descriptions of mild interval changes, such as ``unchanged to slightly smaller''. Hallucination errors were also observed in this task, where the model inferred disease progression that was not explicitly stated or supported by the source text.

\begin{figure*}[!htbp]
\centering
\includegraphics[width=2\columnwidth]{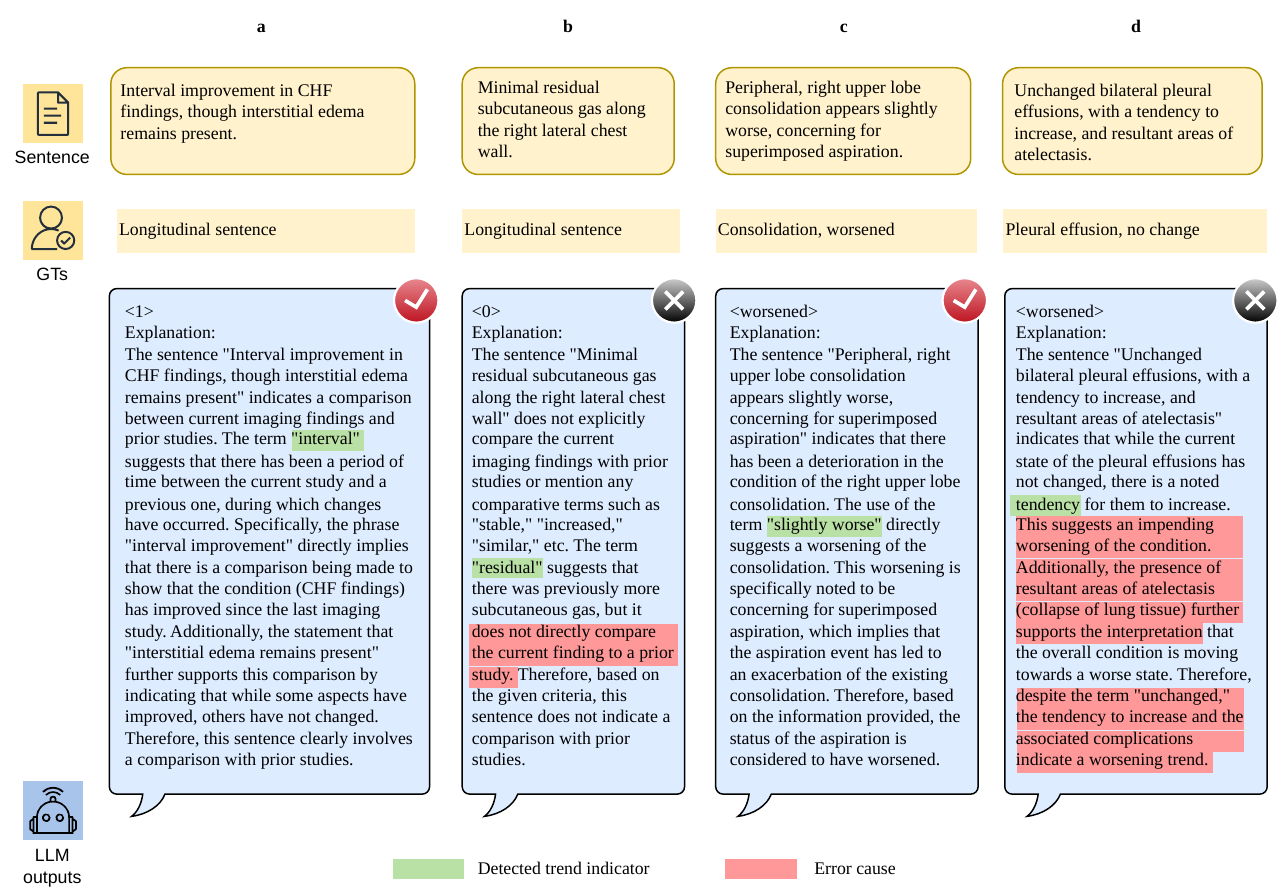}
\caption{\label{fig:explainability} Interpretability of Qwen2.5-32B on longitudinal and disease progression annotation tasks. (a, b) longitudinal annotation (one correct, one incorrect); (c, d) disease progression annotation (one correct, one incorrect). We present the annotated sentences, human labels, and model outputs, with model decision cues (i.e., detected trend indicator) and error sources highlighted in green and red, respectively.} 
\end{figure*}

When classifying these errors, in addition to manual categorization, we use LLM-based interpretability analysis to investigate the underlying causes of model failures. Specifically, based on Qwen2.5-32B, we appended the following instruction to the original prompt: ``Provide your answer first, then explain in detail why you gave it.'' This enabled the model to generate explicit explanations alongside its predictions. Figure \ref{fig:explainability} presents both successful and failed cases from the longitudinal annotation and disease progression annotation tasks, together with the corresponding explanations. We highlight the trend-related terms identified by the model and the factors it identified as contributing to its errors.

\subsubsection{Statistical Analysis \label{sec:stat}}

To assess whether the observed differences among report generation models were statistically meaningful, we performed a paired non-parametric bootstrap analysis on the evaluation set.

Although the L-MIMIC test set contains 1,863 reports, the evaluated models have different input requirements. To ensure a fair head-to-head comparison, all statistical analyses were performed on the common subset of 1,755 reports with corresponding frontal chest radiographs, which can be processed by all seven report generation models (see Appendix Table \ref{tab:model_sources} for details). For completeness and reproducibility, the performance of models that support the full test set of 1,863 reports is additionally reported in Appendix \ref{app:full_performance}. We estimated the uncertainty of each evaluation metric by generating $B=1000$ bootstrap replicates through sampling the $N=1,755$ evaluation cases with replacement while preserving the original sample size. 

To quantify the uncertainty of model evaluation results and assess the robustness of model comparisons, we performed bootstrap analysis based on two aspects: (1) confidence intervals of individual model performance metrics, and (2) statistical comparison between pairs of models. For each bootstrap iteration $b \in \{1,2,\ldots,B\}$, we generated a bootstrap replicate $D^{(b)}$ by sampling $N$ studies from the original evaluation dataset $D=\{x_1,x_2,\ldots,x_N\}$ with replacement:
$D^{(b)}=\{x^{(b)}_1,x^{(b)}_2,\ldots,x^{(b)}_N\}$.

All candidate models were evaluated on the same resampled dataset to preserve the paired correspondence among models, and all evaluation metrics were recomputed for every bootstrap replicate. Let $s_{m,k}^{(b)}$ denote the score of model $m$ on metric $k$ for bootstrap replicate $b$. The two bootstrap-based analyses are defined as follows:

\begin{itemize}

\item \textbf{Bootstrap confidence intervals.}
For each model and evaluation metric, the empirical bootstrap distribution
$
\{s_{m,k}^{(1)},\ldots,s_{m,k}^{(B)}\}
$
was used to estimate the $95\%$ confidence interval using the percentile bootstrap:
\begin{equation}
CI_{95\%}
=
\left[
Q_{0.025}\!\left(\{s_{m,k}^{(b)}\}_{b=1}^{B}\right),
Q_{0.975}\!\left(\{s_{m,k}^{(b)}\}_{b=1}^{B}\right)
\right],
\end{equation}
where $Q_p(\cdot)$ denotes the empirical $p$-th quantile of the bootstrap
distribution.

\item \textbf{Paired bootstrap comparison.}
For each evaluation metric $k$ and each pair of models $(i,j)$, where $i$ and $j$ denote two candidate models, we computed the bootstrap distribution of the paired performance difference:
\begin{equation}
\Delta_{i,j,k}^{(b)}
=
s_{i,k}^{(b)}
-
s_{j,k}^{(b)},
\end{equation}
where $s_{i,k}^{(b)}$ and $s_{j,k}^{(b)}$ denote the bootstrap scores of models $i$ and $j$, respectively, on evaluation metric $k$ for bootstrap replicate $b$.

The mean performance difference was estimated as
\begin{equation}
\bar{\Delta}_{i,j,k}
=
\frac{1}{B}
\sum_{b=1}^{B}
\Delta_{i,j,k}^{(b)},
\end{equation}
together with its percentile-based $95\%$ confidence interval. We also computed the empirical probability that model $i$ outperformed model $j$:
\begin{equation}
P(i>j)
=
\frac{1}{B}
\sum_{b=1}^{B}
\mathbf{1}
\left(
\Delta_{i,j,k}^{(b)}>0
\right),
\end{equation}
where $\mathbf{1}(\cdot)$ is the indicator function, which equals $1$ if the condition is true and $0$ otherwise. This quantity represents the proportion of bootstrap replicates in which model $i$ achieved a higher score than model $j$ on evaluation metric $k$.


\end{itemize}

\section{Results}\label{sec:results}

\subsection{LLM-based annotation performance}

\subsubsection{Greedy decoding performance\label{result:Greedy decoding}}

We compared the annotation performance of the five LLMs (i.e.  MedGemma-27B \cite{sellergren2025medgemma}, MedResearcher-R1-32B \cite{yu2025medreseacher}, Qwen2.5-32B \cite{qwen2.5}, LLama3.3-70B \cite{dubey2024llama}, and Qwen2.5-72B \cite{qwen2.5}) with ImaGenome silver \cite{ImaGenomewu2021chest} (traditional rule-based method used as baseline), as shown in Table \ref{tab:results1} and \ref{tab:results2}. On the task of identifying sentences containing longitudinal information, all LLMs substantially outperformed ImaGenome silver. LLaMA3.3-70B achieved the highest F1 score of 94.0\%, representing an 11.3\% improvement over ImaGenome silver (82.7\%). MedGemma-27B and Qwen2.5-32B also showed significant gains, with F1 scores of 93.2\% (+10.5\%) and 91.8\% (+9.1\%), respectively, while MedResearcher-R1-32B and Qwen2.5-72B scored 90.5\% (+7.8\%) and 89.5\% (+6.8\%) respectively. 

The gains in F1 were primarily driven by increased recall rather than precision. The rule-based method exhibited very high precision (98.5\%) but extremely low recall (71.2\%), resulting in many missed positive cases. In contrast, LLaMA3.3-70B achieved a recall of 96.8\% (+25.6\%) while maintaining high precision (91.4\%). Similar recall improvements were observed across other LLMs, including MedGemma-27B (93.7\%, +22.5\%) and Qwen2.5-32B (88.8\%, +17.6\%), showing that LLMs improve sensitivity without sacrificing precision to a degree that harms overall performance. Consistent with this trend, LLM-based models also showed higher accuracy than the rule-based method (90.2\%), with LLaMA3.3-70B reaching 95.9\% (+5.7\%) and MedGemma-27B 95.5\% (+5.3\%).

\begin{table*}[!htbp]
\centering
\caption{Performance of LLMs on longitudinal annotation. For all large language models, the decoding method is greedy decoding.\label{tab:results1}}

\begin{tabular}{|>{\raggedright\arraybackslash}m{1.8cm}|
                >{\raggedright\arraybackslash}m{2.6cm}|
                >{\raggedright\arraybackslash}m{1.5cm}|
                >{\raggedright\arraybackslash}m{2cm}|
                >{\raggedright\arraybackslash}m{2cm}|
                >{\raggedright\arraybackslash}m{2cm}|
                >{\raggedright\arraybackslash}m{2cm}|}
\hline
\multicolumn{2}{|c|}{}&Model size&Accuracy&Precision&Recall&F1 score\\
\hline
Traditional method&ImaGenome silver&--&90.2&98.5&71.2&82.7\\
\hline
\multirow{5}{*}{\shortstack[l]{LLM-based\\methods}}
& MedGemma & 27B
& 95.5 (\textcolor{myRed}{5.3$\uparrow$})
& 92.7 (\textcolor{myBlue}{5.8$\downarrow$})
& 93.7 (\textcolor{myRed}{22.5$\uparrow$})
& 93.2 (\textcolor{myRed}{10.5$\uparrow$})\\

& MedResearcher-R1 & 32B
& 94.0 (\textcolor{myRed}{3.8$\uparrow$})
& 95.4 (\textcolor{myBlue}{3.1$\downarrow$})
& 86.0 (\textcolor{myRed}{14.8$\uparrow$})
& 90.5 (\textcolor{myRed}{7.8$\uparrow$})\\

& Qwen2.5 & 32B
& 94.8 (\textcolor{myRed}{4.6$\uparrow$})
& 95.1 (\textcolor{myBlue}{3.4$\downarrow$})
& 88.8 (\textcolor{myRed}{17.6$\uparrow$})
& 91.8 (\textcolor{myRed}{9.1$\uparrow$})\\

& LLama3.3 & 70B
& 95.9 (\textcolor{myRed}{5.7$\uparrow$})
& 91.4 (\textcolor{myBlue}{7.1$\downarrow$})
& 96.8 (\textcolor{myRed}{25.6$\uparrow$})
& 94.0 (\textcolor{myRed}{11.3$\uparrow$})\\

& Qwen2.5 & 72B
& 93.4 (\textcolor{myRed}{3.2$\uparrow$})
& 94.4 (\textcolor{myBlue}{4.1$\downarrow$})
& 85.1 (\textcolor{myRed}{13.9$\uparrow$})
& 89.5 (\textcolor{myRed}{6.8$\uparrow$})\\
\hline
\end{tabular}
\end{table*}

\begin{table*}[!htbp]
\centering
\caption{Performance of LLMs on disease progression annotation. The micro-average metric is computed over the three classes: no change, improved, and worsened. For all large language models, the decoding method is greedy decoding. ImaG. silver and MedR-R1 refer to ImaGenome silver and MedResearcher-R1, respectively. * indicates statistically significant improvement over ImaGenome Silver (95\% CI of the difference does not contain zero).\label{tab:results2}}
\renewcommand{\arraystretch}{1.3}
\begin{tabular}{|>{\raggedright\arraybackslash}m{1.6cm}|
                >{\centering\arraybackslash}m{0.8cm}|
                >{\centering\arraybackslash}m{1.8cm}|
                >{\centering\arraybackslash}m{1.8cm}|
                >{\centering\arraybackslash}m{1.8cm}|
                >{\centering\arraybackslash}m{1.8cm}|
                >{\centering\arraybackslash}m{1.9cm}|
                >{\centering\arraybackslash}m{1.8cm}|}
\hline
&
\multirow{2}{*}{\shortstack[l]{Model\\ size}}&
\multirow{2}{*}{Micro Acc.}&
\multirow{2}{*}{Micro F1}&
\multirow{2}{*}{Macro F1}&
\multicolumn{3}{c|}{Per-class F1}\\
\cline{6-8}
&&&&&
No change&Improved&Worsened\\
\hline
ImaG. silver & --
& \makecell[c]{88.7\\[-2pt] \scriptsize[87.3, 90.0]}
& \makecell[c]{78.5\\[-2pt] \scriptsize[75.7, 81.1]}
& \makecell[c]{77.1\\[-2pt] \scriptsize[73.8, 80.1]}
& \makecell[c]{82.1\\[-2pt] \scriptsize[78.8, 85.0]}
& \makecell[c]{76.6\\[-2pt] \scriptsize[69.1, 82.5]}
& \makecell[c]{72.5\\[-2pt] \scriptsize[66.7, 77.8]} \\
\hline
MedGemma & 27B
& 91.4* (\textcolor{myRed}{2.7$\uparrow$})\newline \scriptsize[90.2, 92.6]
& 83.7* (\textcolor{myRed}{5.2$\uparrow$})\newline \scriptsize[81.4, 86.2]
& 81.2* (\textcolor{myRed}{4.1$\uparrow$})\newline \scriptsize[78.3, 84.0]
& 87.9* (\textcolor{myRed}{5.8$\uparrow$})\newline \scriptsize[85.4, 90.4]
& 76.0 (\textcolor{myBlue}{0.6$\downarrow$})\newline \scriptsize[68.5, 82.4]
& 79.6* (\textcolor{myRed}{7.1$\uparrow$})\newline \scriptsize[74.9, 83.8] \\

MedR-R1 & 32B
& 90.1 (\textcolor{myRed}{1.4$\uparrow$})\newline \scriptsize[88.8, 91.5]
& 79.5 (\textcolor{myRed}{1.0$\uparrow$})\newline \scriptsize[76.7, 82.4]
& 76.6 (\textcolor{myBlue}{0.5$\downarrow$})\newline \scriptsize[73.0, 80.0]
& 84.6 (\textcolor{myRed}{2.5$\uparrow$})\newline \scriptsize[81.8, 87.6]
& 74.9 (\textcolor{myBlue}{1.7$\downarrow$})\newline \scriptsize[67.6, 81.3]
& 70.4 (\textcolor{myBlue}{2.1$\downarrow$})\newline \scriptsize[64.4, 75.9] \\

Qwen2.5 & 32B
& 91.9* (\textcolor{myRed}{3.2$\uparrow$})\newline \scriptsize[90.6, 93.1]
& 83.2* (\textcolor{myRed}{4.7$\uparrow$})\newline \scriptsize[80.5, 85.8]
& 82.1* (\textcolor{myRed}{5.0$\uparrow$})\newline \scriptsize[79.2, 85.0]
& 87.0* (\textcolor{myRed}{4.9$\uparrow$})\newline \scriptsize[84.4, 89.7]
& 87.1* (\textcolor{myRed}{10.5$\uparrow$})\newline \scriptsize[82.1, 91.6]
& 72.2 (\textcolor{myBlue}{0.3$\downarrow$})\newline \scriptsize[66.5, 77.6] \\

LLama3.3 & 70B
& 90.5* (\textcolor{myRed}{1.8$\uparrow$})\newline \scriptsize[89.2, 91.8]
& 78.9 (\textcolor{myRed}{0.4$\uparrow$})\newline \scriptsize[75.9, 81.7]
& 78.9 (\textcolor{myRed}{1.8$\uparrow$})\newline \scriptsize[75.9, 81.7]
& 79.8 (\textcolor{myBlue}{2.3$\downarrow$})\newline \scriptsize[76.1, 83.2]
& 79.4 (\textcolor{myRed}{2.8$\uparrow$})\newline \scriptsize[73.2, 84.3]
& 77.4 (\textcolor{myRed}{4.9$\uparrow$})\newline \scriptsize[73.0, 81.2] \\

Qwen2.5 & 72B
& 91.2* (\textcolor{myRed}{2.5$\uparrow$})\newline \scriptsize[89.9, 92.6]
& 83.8* (\textcolor{myRed}{5.3$\uparrow$})\newline \scriptsize[81.2, 86.2]
& 82.5* (\textcolor{myRed}{5.4$\uparrow$})\newline \scriptsize[79.2, 85.3]
& 87.7* (\textcolor{myRed}{5.6$\uparrow$})\newline \scriptsize[85.0, 90.1]
& 83.7 (\textcolor{myRed}{7.1$\uparrow$})\newline \scriptsize[77.2, 88.7]
& 76.2 (\textcolor{myRed}{3.7$\uparrow$})\newline \scriptsize[70.9, 81.1] \\

\hline
\end{tabular}
\end{table*}

In disease progression prediction, LLMs consistently outperformed rule-based ImaGenome silver. Overall, micro-average accuracy ranged from 90.1–91.9\%, representing improvements of 1.4–3.2\% over ImaGenome Silver (88.7\%). Qwen2.5-32B achieved the highest accuracy (91.9\%). Micro-average F1 scores ranged from 78.9–83.8\%, corresponding to gains of 0.4–5.3\% over ImaGenome Silver (78.5\%). Qwen2.5-72B (83.8\%) showed the strongest performance. Macro-average F1 scores showed a similar trend to micro-average F1, ranging from 76.6–82.5\%, corresponding to improvements of 1.8–5.4\% over ImaGenome Silver (77.1\%). The only exception was MedResearcher-R1, which was 0.5\% lower than ImaGenome Silver. Qwen2.5-72B achieved the best macro-average F1 (82.5\%).

Different models excelled in different categories. In the ``no change" category, MedGemma-27B achieved an F1 score of 87.9\%, outperforming ImaGenome by 5.8\% (82.1\%), while Qwen2.5-32B reached 87\%, an improvement of 4.9\%. In the ``improved" category, Qwen2.5-32B performed the best, achieving 87.1\% and leading ImaGenome by 10.5\%. In the ``worsened" category, Qwen2.5-32B, which performed well in the first two categories, is comparable to ImaGenome, whereas MedGemma-27B outperformed ImaGenome by 7.1\%.

Beyond the point estimates, we assessed the robustness of the observed performance differences using bootstrap confidence intervals and paired bootstrap comparisons. All significance tests compared each LLM-based method against the ImaGenome Silver baseline. MedGemma, Qwen2.5-32B, and Qwen2.5-72B achieved statistically significant improvements over ImaGenome Silver on all overall evaluation metrics (micro-average accuracy, micro-average F1, and macro-average F1), with 95\% confidence intervals of the paired differences excluding zero. MedResearcher-R1 did not show statistically significant improvements, whereas Llama3.3 achieved a significant improvement only in micro-average accuracy. For class-wise performance, the selected Qwen2.5-32B model significantly outperformed the silver baseline on the improved and no change classes, while achieving comparable performance to the silver annotations on the worsened class.

\subsubsection{Sensitivity to sampling variability\label{result:sample}}

\begin{figure*}[htbp]
    \centering
    \vspace*{2em}
    \begin{subfigure}[b]{0.45\textwidth}
        \centering
        \begin{overpic}[width=\textwidth]{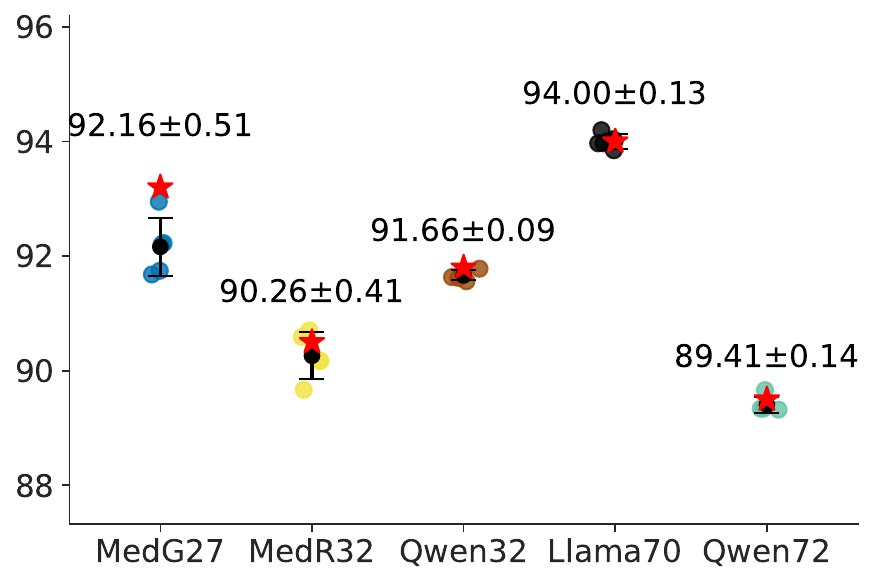}
            \put(-2,70){a}
        \end{overpic}
    \end{subfigure}
    \hspace{0.01\textwidth}
    \begin{subfigure}[b]{0.45\textwidth}
        \centering
        \begin{overpic}[width=\textwidth]{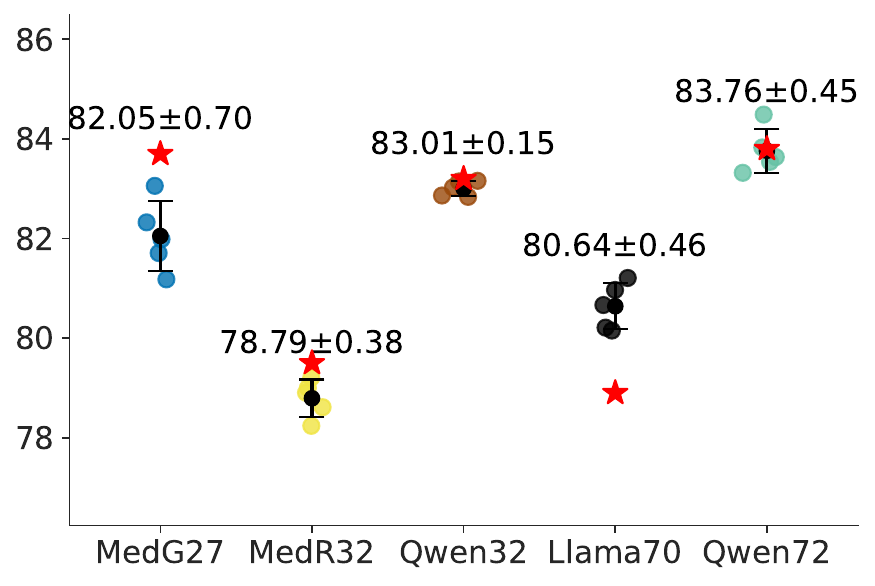}
            \put(-2,70){b}
        \end{overpic}
    \end{subfigure}

    \caption{Performance of five large language models on two tasks using stochastic decoding
    ($T=0.7$, top\_p$=0.9$, top\_k$=20$). Each model was evaluated across five
    independent runs. Dots represent individual runs; error bars indicate mean
    $\pm$ SD; red stars denote greedy decoding performance.
    (a) Longitudinal annotation F1(\%) score.
    (b) Disease progression annotation micro F1(\%).}
    \label{fig:samp}
\end{figure*}

As shown in Figure~\ref{fig:samp}, the performance of F1 measure was highly stable across five independent runs for all models on both tasks, suggesting that the evaluation was robust to sampling randomness. The standard deviations ranged from 0.09\% to 0.51\% for longitudinal annotation and from 0.15\% to 0.70\% for disease progression.

Among all models, Qwen2.5-32B exhibited the lowest run-to-run variability. Its mean performance under stochastic sampling was very close to that obtained with greedy decoding, achieving $91.66\% \pm 0.09\%$ versus $91.8\%$ on the longitudinal annotation task and $83.01\% \pm 0.15\%$ versus $83.2\%$ on the disease progression task.

Qwen2.5-72B, Llama3.3-70B, and MedResearcher-R1-32B showed moderate variability across runs, whereas MedGemma-27B exhibited the largest standard deviations (0.51\% and 0.70\% on the longitudinal annotation and disease progression tasks, respectively). MedGemma-27B was also the most sensitive to decoding strategy: greedy decoding outperformed the five-run stochastic sampling average by 1.04\% and 1.65\% on the longitudinal annotation and disease progression tasks, respectively. In contrast, Llama3.3-70B showed a slight advantage under stochastic sampling, achieving a higher average score (80.64\%) compared to greedy decoding (78.9\%) on the disease progression task.

\subsubsection{Performance of zero-shot and few-shot prompting\label{result:zeroshot}}

\begin{table}[htbp]
\centering
\caption{Zero-shot and few-shot prompt performance of Qwen2.5-32B on longitudinal and disease progression annotation. Bold text indicates the best result.}
\label{tab:example}

\textbf{Panel A. Longitudinal annotation with task examples.}

\vspace{2mm}

\begin{tabular}{llll}
\toprule
Setting&Accuracy&F1 score\\
\midrule
0-shot&91.8&87.5\\
3-shot&\textbf{94.8}&\textbf{91.8}\\
5-shot&92.4&88.4\\
\bottomrule
\end{tabular}

\vspace{4mm}

\textbf{Panel B. Longitudinal annotation with change-indicator examples.}

\vspace{2mm}

\begin{tabular}{llll}
\toprule
Setting&Accuracy&F1 score\\
\midrule
0-shot&88.1&80.1\\
4-shot&86.4&76.6\\
8-shot&\textbf{94.8}&\textbf{91.8}\\
12-shot&93.0&89.4\\
\bottomrule
\end{tabular}

\vspace{4mm}

\textbf{Panel C. Disease progression annotation with task examples.}

\vspace{2mm}

\begin{tabular}{lll}
\toprule
Setting&Micro Acc.&Micro F1.\\
\midrule
0-shot&95.9&86.2\\
2-shot&\textbf{96.0}&\textbf{86.4}\\
4-shot&95.9&86.2\\
\bottomrule
\end{tabular}

\end{table}

As shown in Table \ref{tab:example}, few-shot Qwen2.5-32B consistently outperformed zero-shot prompting on both tasks. However, the effect of example quantity differed between tasks.

Longitudinal annotation showed greater sensitivity to the number of task examples than disease progression annotation. For longitudinal annotation with different task examples (Panel A), accuracy ranged from 91.8\% to 94.8\% and F1 scores from 87.5\% to 91.8\%, with the best performance achieved at 3 shots (94.8\% accuracy, 91.8\% F1). In contrast, disease progression annotation (Panel C) was largely unaffected by the number of task examples, with accuracy ranging from 95.9\% to 96.0\% and F1 from 86.2\% to 86.4\%.

Longitudinal annotation is more sensitive to change-indicator examples than to task examples, though this effect weakens at higher shot counts (Panel B). Accuracy ranged from 86.4\% to 94.8\% and F1 from 76.6\% to 91.8\%. Performance improved substantially from 0 to 8 shots and then plateaued or slightly declined at higher shot counts. The best performance was achieved at 8 shots (94.8\% accuracy, 91.8\% F1).

\subsection{Errors in LLM-based annotation \label{result:error}}

The error distribution is shown in Figure \ref{fig:error}. Overall, we identified errors in 103 sentences for longitudinal annotation and 77 sentences for disease progression annotation.  Inconsistencies are the dominant error type across both annotation tasks, accounting for 56.3\% and 62.3\% of errors in longitudinal and disease progression annotations, respectively. In longitudinal annotations, true errors (29.1\%) and incorrect corrections (14.7\%) constitute the remaining error types. In disease progression annotations, true errors represent 35.1\% of cases, with hallucinations and incorrect corrections each accounting for 1.3\%.

Analysis of error subcategories showed that, in the longitudinal annotation task, most true errors (28/30) were attributed to ignored implicit time-related terms (e.g., old). No cases of ignored explicit comparisons to prior examinations were observed among the 1,975 evaluated sentences. All inconsistency errors were associated with indicators that required comparison with prior reports for accurate interpretation. Incorrect corrections mainly stem from the neglect of clearly identified longitudinal contexts.

In the disease progression annotation task, true errors were more heterogeneous. Of the 27 true errors, 8 involved failures to detect uncertain progression indicators, 7 involved failures to detect implicit temporal/progression cues, 7 involved misclassifying newly identified findings as worsening, and 4 involved failures to propagate severity modifiers from one finding to adjacent findings in long sentences. One additional error involved a mixed progression description, in which the model predicted no change instead of detecting concurrent improvement and worsening. For inconsistency errors, the majority were caused by unclear indicators (39/47), followed by unclear descriptions (4), slight progression (3), and disease annotation mismatches (2). Hallucinations and incorrect corrections were rare (1), accounting for only a small proportion of the overall errors.

\subsection{L-MIMIC for report generation evaluation\label{result:reportgeneration}}

\begin{figure*}[!htbp]
\centering
\includegraphics[width=2\columnwidth]{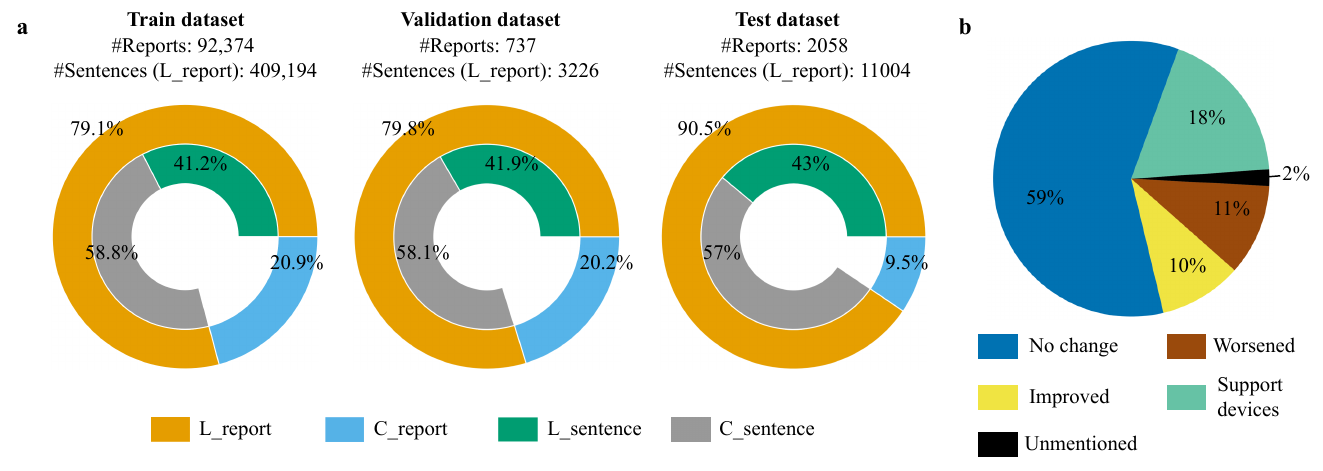}
\caption{Annotation statistics for follow-up reports in MIMIC-CXR. a. Proportion of sentences and reports with longitudinal information. L\_report/L\_sentence: reports/sentences with longitudinal information. C\_report/C\_sentence: reports/sentences that convey only cross-sectional information and exclude any longitudinal information. b. Proportion of disease progression categories in L-MIMIC.}
\label{res:data_annotation}
\end{figure*}

The follow-up reports in the MIMIC-CXR dataset comprise 92,374 reports for training, 737 for validation, and 2,058 for testing, following its official split. Figure \ref{res:data_annotation} shows its annotation statistics. While most follow-up reports include disease progression information. About 20\% of training/validation and 10\% of test reports lack such details (Figure \ref{res:data_annotation}a). Among disease progression, 59\%, 10\%, and 11\% indicate ``no change'', ``improved'', and ``worsened'', respectively. Furthermore, variations in the support device, such as its position, accounted for 18\% (Figure \ref{res:data_annotation}b). We extracted cases whose reports were labeled as containing longitudinal information, forming the L-MIMIC dataset with 73,093 training, 588 validation, and 1,863 test cases. The training and validation sets include longitudinal annotations to support the development of medical image analysis models, and are suitable for tasks such as pre-training, hyperparameter setting, and fine-tuning. The test set is designed for evaluating model performance particularly in report generation and disease progression information extraction. Detailed application scenarios for longitudinal annotation are discussed in Section \ref{sec:scenario}.

\vspace{0.5em}

\begin{figure*}[htbp]
    \centering

    \begin{subfigure}[b]{0.48\textwidth}
        \centering
        \begin{overpic}[width=\linewidth]{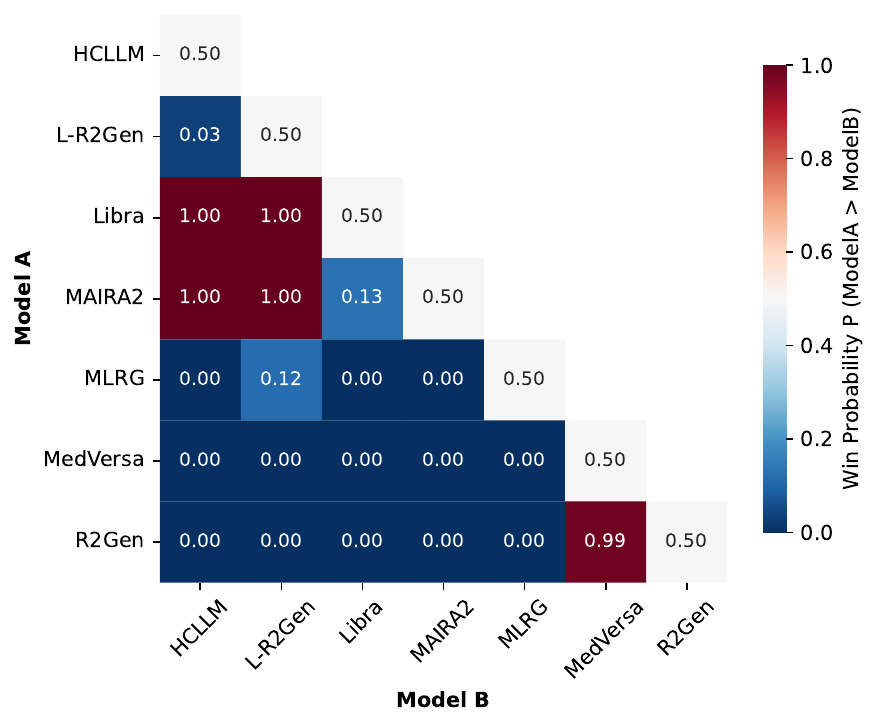}
            \put(-2,70){\textbf{a}}
        \end{overpic}
    \end{subfigure}
    \hfill
    \begin{subfigure}[b]{0.48\textwidth}
        \centering
        \begin{overpic}[width=\linewidth]{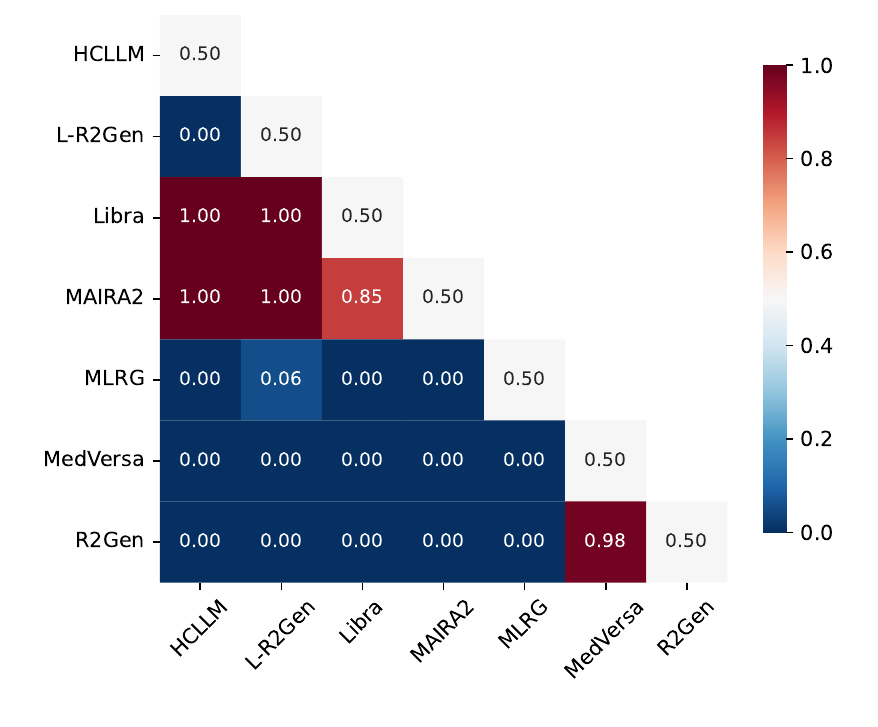}
            \put(-2,70){\textbf{b}}
        \end{overpic}
    \end{subfigure}

    \vspace{0.5em}

    \begin{subfigure}[b]{0.48\textwidth}
        \centering
        \begin{overpic}[width=\linewidth]{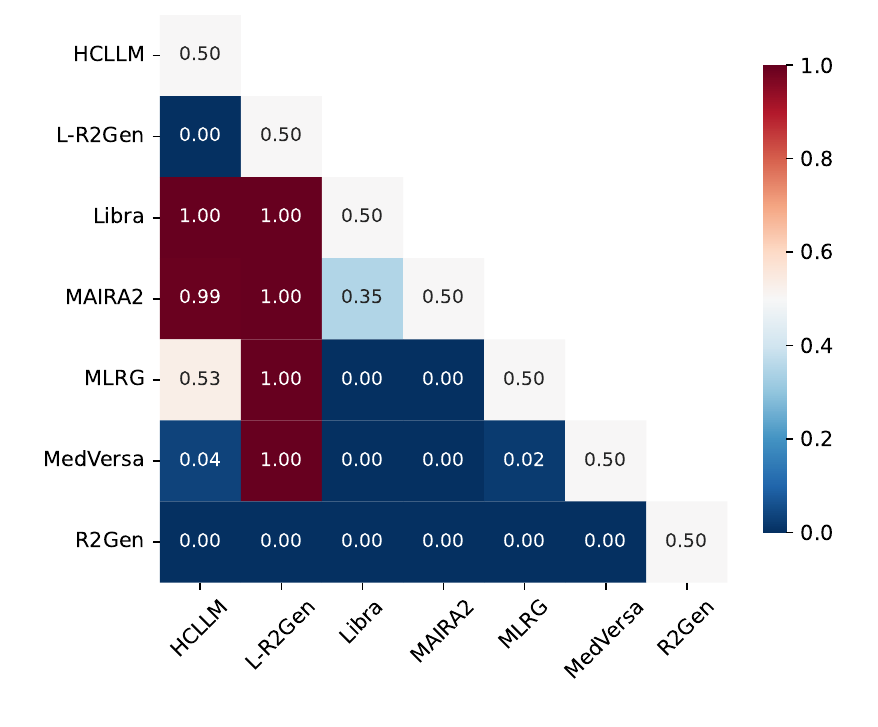}
            \put(-2,70){\textbf{c}}
        \end{overpic}
    \end{subfigure}
    \hfill
    \begin{subfigure}[b]{0.48\textwidth}
        \centering
        \begin{overpic}[width=\linewidth]{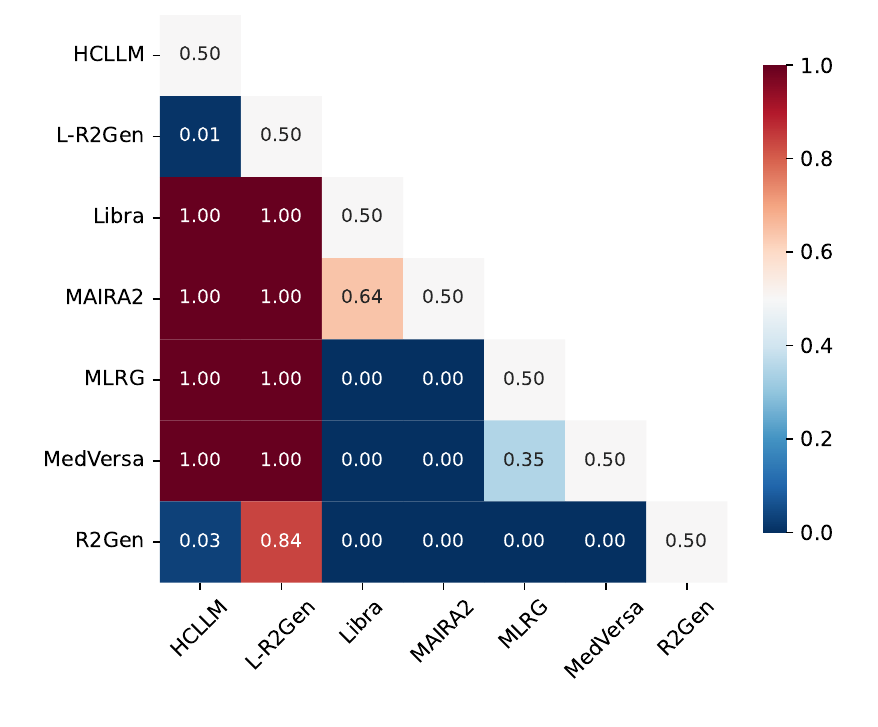}
            \put(-2,70){\textbf{d}}
        \end{overpic}
    \end{subfigure}

    \caption{Pairwise empirical win probability among seven automated report generation models. (a) micro-averaged F1. (b) F1 score for the no change class. (c) F1 score for the improved class. (d) F1 score for the worsened class.}
    \label{fig:winprob_heatmaps}
\end{figure*}

We evaluated seven report-generation systems (i.e., R2Gen \cite{chen2020generating}, MedVersa \cite{15zhou2024medversa}, L\_R2Gen \cite{3zhu2023utilizing}, MLRG \cite{5liu2025enhanced}, HC-LLM \cite{7liu2025hc}, Maira2 \cite{14bannur2024maira}, and Libra \cite{13zhang-etal-2025-libra})  on the L-MIMIC test set to assess their ability to capture longitudinal information, therefore demonstrating the advantages of using L-MIMIC for method evaluation. Table \ref{table:boot_evaluation} summarizes the key metrics: BLEU-4  \cite{BLEUpapineni2002bleu} and METEOR \cite{Methorbanerjee2005meteor} for language quality, micro-average F1 for disease diagnosis, and both micro-average and class-level F1 for disease-progression categories. 
Regarding the statistical analysis described in Section \ref{sec:med_decoding}, Table \ref{table:boot_evaluation} reports model performance together with 95\% bootstrap confidence intervals. Statistical significance is assessed based on whether the confidence interval of the performance difference excludes zero. We further report the significance of the best-performing model against the second-best model for each evaluation metric. Libra achieves significantly better performance than competing models on language-based metrics for both whole-report and longitudinal-sentence evaluations. In contrast, no model significantly outperforms the others on medical correctness metrics, including our proposed progression F1 metrics. To further assess the statistical reliability of the observed rankings, Figure \ref{fig:winprob_heatmaps} presents pairwise empirical win probabilities between the seven report generation models based on progression micro-F1 and category-wise F1 scores. The pairwise comparisons based on mean performance differences show consistent trends with the empirical win probability analysis, with additional results provided in Appendix Figure \ref{appfig:diff_heatmaps}.

\begin{table*}[!htbp]
\centering
\small 
\setlength{\tabcolsep}{2.5pt} 
\caption{Model performance on the L-MIMIC test set. Values are reported as point estimates with 95\% bootstrap confidence intervals, computed from 1,000 paired bootstrap resamples. A\_F1, micro-average F1 score; N\_F1, I\_F1, W\_F1, F1 scores for no-change, improved, and worsened classes,  respectively. Higher values are better for all indicators. \textbf{Bold} indicates the best performance in each column. $^*$ denotes that the best result is statistically significantly better than the second-best model (95\% CI of the difference does not contain zero).}\label{table:boot_evaluation}

\begin{tabular}{ >{\raggedright\arraybackslash}p{1.3cm} 
>{\centering\arraybackslash}p{1.6cm} 
>{\centering\arraybackslash}p{1.6cm} 
>{\centering\arraybackslash}p{1.6cm} 
>{\centering\arraybackslash}p{1.6cm} 
>{\centering\arraybackslash}p{1.6cm} 
>{\centering\arraybackslash}p{1.6cm} 
>{\centering\arraybackslash}p{1.6cm} 
>{\centering\arraybackslash}p{1.6cm} 
>{\centering\arraybackslash}p{1.6cm} 
}
\toprule
\multirow{3}{*}{Models} 
& \multicolumn{4}{c}{Language-based metrics} 
& \multicolumn{5}{c}{Medical correctness-based metrics (\%)} \\
\cmidrule(lr){2-5}\cmidrule(lr){6-10}
 & \multicolumn{2}{c}{Whole report} & \multicolumn{2}{c}{Longitudinal sentences} 
& Diagnosis & \multicolumn{4}{c}{Progression} \\
\cmidrule(lr){2-3} \cmidrule(lr){4-5} \cmidrule(lr){6-6} \cmidrule(lr){7-10}
& BLEU-4 & METEOR & BLEU-4 & METEOR & F1 & A\_F1 & N\_F1 & I\_F1 & W\_F1 \\
\midrule
R2Gen
& \begin{tabular}[c]{@{}c@{}}9.4\\ \scriptsize[9.0, 9.9]\end{tabular}
& \begin{tabular}[c]{@{}c@{}}13.4\\ \scriptsize[13.1, 13.6]\end{tabular}
& \begin{tabular}[c]{@{}c@{}}2.3\\ \scriptsize[2.0, 2.6]\end{tabular}
& \begin{tabular}[c]{@{}c@{}}5.8\\ \scriptsize[5.5, 6.1]\end{tabular}
& \begin{tabular}[c]{@{}c@{}}33.2\\ \scriptsize[31.8, 34.6]\end{tabular}
& \begin{tabular}[c]{@{}c@{}}45.3\\ \scriptsize[43.8, 46.7]\end{tabular}
& \begin{tabular}[c]{@{}c@{}}54.5\\ \scriptsize[52.9, 56.0]\end{tabular}
& \begin{tabular}[c]{@{}c@{}}0.0\\ \scriptsize[0.0, 0.0]\end{tabular}
& \begin{tabular}[c]{@{}c@{}}5.2\\ \scriptsize[3.2, 7.3]\end{tabular} \\
\addlinespace[2pt]

MedVersa
& \begin{tabular}[c]{@{}c@{}}15.0\\ \scriptsize[14.4, 15.6]\end{tabular}
& \begin{tabular}[c]{@{}c@{}}16.5\\ \scriptsize[16.2, 16.8]\end{tabular}
& \begin{tabular}[c]{@{}c@{}}2.8\\ \scriptsize[2.4, 3.3]\end{tabular}
& \begin{tabular}[c]{@{}c@{}}7.2\\ \scriptsize[6.8, 7.6]\end{tabular}
& \begin{tabular}[c]{@{}c@{}}53.1\\ \scriptsize[51.9, 54.4]\end{tabular}
& \begin{tabular}[c]{@{}c@{}}43.1\\ \scriptsize[41.6, 44.6]\end{tabular}
& \begin{tabular}[c]{@{}c@{}}52.2\\ \scriptsize[50.3, 53.8]\end{tabular}
& \begin{tabular}[c]{@{}c@{}}12.0\\ \scriptsize[9.0, 14.8]\end{tabular}
& \begin{tabular}[c]{@{}c@{}}17.3\\ \scriptsize[14.0, 20.3]\end{tabular} \\
\addlinespace[2pt]

L\_R2Gen
& \begin{tabular}[c]{@{}c@{}}9.1\\ \scriptsize[8.7, 9.6]\end{tabular}
& \begin{tabular}[c]{@{}c@{}}13.0\\ \scriptsize[12.7, 13.2]\end{tabular}
& \begin{tabular}[c]{@{}c@{}}3.2\\ \scriptsize[2.8, 3.6]\end{tabular}
& \begin{tabular}[c]{@{}c@{}}6.4\\ \scriptsize[6.1, 6.7]\end{tabular}
& \begin{tabular}[c]{@{}c@{}}42.4\\ \scriptsize[40.9, 43.8]\end{tabular}
& \begin{tabular}[c]{@{}c@{}}49.6\\ \scriptsize[48.1, 51.1]\end{tabular}
& \begin{tabular}[c]{@{}c@{}}59.3\\ \scriptsize[57.7, 60.8]\end{tabular}
& \begin{tabular}[c]{@{}c@{}}4.7\\ \scriptsize[1.6, 7.1]\end{tabular}
& \begin{tabular}[c]{@{}c@{}}4.0\\ \scriptsize[1.9, 5.8]\end{tabular} \\
\addlinespace[2pt]

MLRG
& \begin{tabular}[c]{@{}c@{}}15.0\\ \scriptsize[14.4, 15.6]\end{tabular}
& \begin{tabular}[c]{@{}c@{}}16.8\\ \scriptsize[16.5, 17.2]\end{tabular}
& \begin{tabular}[c]{@{}c@{}}6.8\\ \scriptsize[6.2, 7.4]\end{tabular}
& \begin{tabular}[c]{@{}c@{}}9.6\\ \scriptsize[9.1, 10.0]\end{tabular}
& \begin{tabular}[c]{@{}c@{}}53.1\\ \scriptsize[51.8, 54.3]\end{tabular}
& \begin{tabular}[c]{@{}c@{}}48.5\\ \scriptsize[47.2, 49.9]\end{tabular}
& \begin{tabular}[c]{@{}c@{}}57.7\\ \scriptsize[56.2, 59.2]\end{tabular}
& \begin{tabular}[c]{@{}c@{}}15.8\\ \scriptsize[12.3, 19.3]\end{tabular}
& \begin{tabular}[c]{@{}c@{}}18.2\\ \scriptsize[15.1, 21.1]\end{tabular} \\
\addlinespace[2pt]

HC-LLM
& \begin{tabular}[c]{@{}c@{}}11.7\\ \scriptsize[11.3, 12.1]\end{tabular}
& \begin{tabular}[c]{@{}c@{}}15.6\\ \scriptsize[15.4, 15.9]\end{tabular}
& \begin{tabular}[c]{@{}c@{}}5.4\\ \scriptsize[4.9, 5.9]\end{tabular}
& \begin{tabular}[c]{@{}c@{}}9.0\\ \scriptsize[8.6, 9.3]\end{tabular}
& \begin{tabular}[c]{@{}c@{}}43.4\\ \scriptsize[42.0, 44.7]\end{tabular}
& \begin{tabular}[c]{@{}c@{}}51.3\\ \scriptsize[49.9, 52.8]\end{tabular}
& \begin{tabular}[c]{@{}c@{}}62.2\\ \scriptsize[60.7, 63.7]\end{tabular}
& \begin{tabular}[c]{@{}c@{}}15.6\\ \scriptsize[12.4, 18.6]\end{tabular}
& \begin{tabular}[c]{@{}c@{}}8.5\\ \scriptsize[5.9, 11.0]\end{tabular} \\
\addlinespace[2pt]

Maira2
& \begin{tabular}[c]{@{}c@{}}19.1\\ \scriptsize[18.4, 19.8]\end{tabular}
& \begin{tabular}[c]{@{}c@{}}19.5\\ \scriptsize[19.2, 19.9]\end{tabular}
& \begin{tabular}[c]{@{}c@{}}9.5\\ \scriptsize[8.8, 10.1]\end{tabular}
& \begin{tabular}[c]{@{}c@{}}12.4\\ \scriptsize[11.9, 12.8]\end{tabular}
& \begin{tabular}[c]{@{}c@{}}\textbf{58.2}\\ \scriptsize[56.9, 59.3]\end{tabular}
& \begin{tabular}[c]{@{}c@{}}56.3\\ \scriptsize[55.2, 57.5]\end{tabular}
& \begin{tabular}[c]{@{}c@{}}\textbf{67.3}\\ \scriptsize[66.0, 68.6]\end{tabular}
& \begin{tabular}[c]{@{}c@{}}21.6\\ \scriptsize[18.2, 24.6]\end{tabular}
& \begin{tabular}[c]{@{}c@{}}\textbf{24.9}\\ \scriptsize[21.8, 27.9]\end{tabular} \\
\addlinespace[2pt]

Libra
& \begin{tabular}[c]{@{}c@{}}\textbf{23.6}$^*$\\ \scriptsize[22.9, 24.4]\end{tabular}
& \begin{tabular}[c]{@{}c@{}}\textbf{22.6}$^*$\\ \scriptsize[22.3, 22.9]\end{tabular}
& \begin{tabular}[c]{@{}c@{}}\textbf{12.7}$^*$\\ \scriptsize[12.0, 13.4]\end{tabular}
& \begin{tabular}[c]{@{}c@{}}\textbf{13.9}$^*$\\ \scriptsize[13.4, 14.4]\end{tabular}
& \begin{tabular}[c]{@{}c@{}}57.6\\ \scriptsize[56.4, 58.9]\end{tabular}
& \begin{tabular}[c]{@{}c@{}}\textbf{57.1}\\ \scriptsize[55.9, 58.4]\end{tabular}
& \begin{tabular}[c]{@{}c@{}}66.5\\ \scriptsize[65.2, 67.7]\end{tabular}
& \begin{tabular}[c]{@{}c@{}}\textbf{22.4}\\ \scriptsize[18.9, 25.8]\end{tabular}
& \begin{tabular}[c]{@{}c@{}}24.3\\ \scriptsize[20.9, 27.5]\end{tabular} \\
\bottomrule
\end{tabular}
\end{table*}

All report-generation models exhibited considerable scope for improvement in capturing longitudinal information. Libra outperformed counterparts on language-based metrics, with BLEU-4 and METEOR scores of 23.6 (95\% CI, 22.9--24.4) and 22.6 (22.3--22.9) for full reports, and 12.7 (12.0--13.4) and 13.9 (13.4--14.4) for longitudinal sentences, respectively; these improvements over the second-best model were statistically significant. Maira2 achieved the highest diagnostic F1 (58.2\%, 95\% CI, 56.9--59.3) and Libra attained the best overall progression performance (micro-average F1 57.1\%, 95\% CI, 55.9--58.4). However, the observed differences in diagnostic and progression performance were not statistically significant.

Models with longitudinal priors achieved higher performance on longitudinal information generation than models without priors. As detailed in Appendix~\ref{model_config}, the model input configurations vary regarding prior context (text reports and/or images). Specifically, for the model versions evaluated in our study, R2Gen and MedVersa use neither prior images nor prior reports. In contrast, L\_R2Gen, MLRG, HC-LLM, and Maira2 incorporate both, while Libra utilizes prior images alone. R2Gen and MedVersa, which lack input of prior information, achieved longitudinal BLEU-4 scores of 2.3 (95\% CI: 2.0--2.6) and 2.8 (2.4--3.3), METEOR scores of 5.8 (5.5--6.1) and 7.2 (6.8--7.6), and diagnosis F1 scores of 33.2\% (31.8--34.6) and 53.1\% (51.9--54.4), respectively. Methods incorporating prior information consistently achieved stronger longitudinal performance, with BLEU-4 scores ranging from 3.2 to 12.7, METEOR scores from 6.4 to 13.9, and diagnosis F1 scores from 42.4\% to 58.2\%. Compared with R2Gen, the longitudinally enhanced variant L\_R2Gen showed lower performance on whole-report language generation metrics. However, it achieved improved longitudinal sentence generation performance (BLEU-4: 3.2 [2.8--3.6] vs. 2.3 [2.0--2.6]; METEOR: 6.4 [6.1--6.7] vs. 5.8 [5.5--6.1]), as well as higher diagnosis F1 (42.4\% [40.9--43.8] vs. 33.2\% [31.8--34.6]) and progression micro-F1 (49.6\% [48.1--51.1] vs. 45.3\% [43.8--46.7]), suggesting the effectiveness of incorporating longitudinal information.

Across the three progression categories, all models exhibit substantial class imbalance, with the no-change category consistently achieving markedly higher performance than the improved and worsened categories. R2Gen and L\_R2Gen show the most pronounced disparity, with no-change F1 scores of 54.5\% (95\% CI: 52.9--56.0) and 59.3\% (57.7--60.8), respectively, whereas their improved and worsened F1 scores remain below 5\%. More recent models alleviate this imbalance to varying degrees, yielding noticeable gains on the minority progression classes while largely preserving strong no-change performance. For example, Maira2 achieves improved and worsened F1 scores of 21.6\% (18.2--24.6) and 24.9\% (21.8--27.9), respectively, together with a no-change F1 of 67.3\% (66.0--68.6). Libra further improves the improved class to 22.4\% (18.9--25.8) while maintaining a comparable no-change F1 of 66.5\% (65.2--67.7) and a worsened F1 of 24.3\% (20.9--27.5).

Figure \ref{fig:winprob_heatmaps} illustrates the robustness of model rankings using pairwise empirical win probabilities. Across all four progression-related metrics, most pairwise comparisons yield empirical win probabilities close to 0 or 1, indicating that the relative rankings of most model pairs remain stable under bootstrap resampling. For the micro-averaged F1 score (Figure \ref{fig:winprob_heatmaps}a) and the F1 score of the no change class (Figure \ref{fig:winprob_heatmaps}b), Libra and MAIRA2 consistently emerge as the top-performing models, substantially outperforming all remaining baselines. This suggests that their superiority is robust, whereas the performance difference between the two is comparatively small. Longitudinal-R2Gen and MLRG, which occupy the lower end of the rankings, also exhibit relatively lower ranking certainty than most other model pairs. The other models all exhibit very high ranking stability.

Ranking uncertainty is relatively more pronounced for the F1 scores of the improved and worsened classes (Figure \ref{fig:winprob_heatmaps}c--d). In particular, MLRG and HCLLM show nearly indistinguishable performance on the improved class, with an empirical win probability of 0.53, suggesting no clear advantage between the two models. Similarly, Libra and MAIRA2 exhibit comparable performance, while both consistently outperform the remaining models. A similar pattern is observed for WF1, where the relative rankings of Libra versus MAIRA2, Longitudinal-R2Gen versus R2Gen, and MLRG versus MedVersa remain uncertain. However, consistent with the findings from the micro-averaged F1 score and the F1 score of the no change class metrics, all remaining models demonstrate high ranking stability.

\section{Discussion}\label{sec:discussion}

In this study, we benchmark LLMs' ability to annotate longitudinal information in radiology reports, and further scale this annotation pipeline to annotate a large public dataset. Building on this validated LLM-based annotation methodology, we develop a systematic evaluation framework tailored to assessing the capacity of radiology report generation models to capture longitudinal information, and apply this framework to evaluate seven mainstream report generation models, revealing their limitations. In the following discussion, we elaborate the advantages of LLM-based annotation, the influence of LLM scale and training dataset, potential application scenarios of automated longitudinal information extraction, and the performance of report generation models. Finally, we discuss the limitations and future work of this study.

\subsection{Advantages of LLM-based longitudinal information  annotation}

We evaluated five mainstream LLMs on their ability to identify longitudinal descriptions and track disease progression. The results are presented in Section \ref{result:Greedy decoding}. In the task of identifying longitudinal descriptions, all LLMs significantly outperformed the conventional rule-based method (i.e., ImaGenome silver), with MedGemma-27B, LLaMA3.3-70B, and Qwen2.5-32B achieving the highest performance. For disease progression, MedGemma-27B, Qwen2.5-72B, and Qwen2.5-32B demonstrated a clear advantage over traditional approaches, whereas other models achieved only modest improvements. 

In terms of stability, across five runs with stochastic sampling, all evaluated models showed very low performance variance, indicating strong robustness to decoding randomness, as presented in Section \ref{result:sample}. This suggests that the superior performance of large language models on the two tasks is largely driven by their underlying task competence rather than by favorable decoding outcomes. Among the evaluated models, Qwen2.5-32B exhibited the lowest variance and thus the highest stability. Considering both annotation accuracy, efficiency, and stability, Qwen2.5-32B was selected as a practical tool for large-scale annotation of the MIMIC-CXR dataset (see Appendix \ref{sup:LLM_select}), facilitating the construction of a longitudinally labeled resource for downstream studies of longitudinal description identification and disease progression.

Zero-shot and few-shot prompt sensitivity analysis examined the extent to which model performance depends on contextual guidance versus intrinsic capability. The results are presented in Section \ref{result:zeroshot}. The two tasks behaved differently. Disease progression annotation showed stable performance across varying numbers of in-context examples, indicating reliance on pretrained knowledge with limited gains from additional demonstrations. In contrast, longitudinal annotation was more sensitive to prompting, with performance improving from zero-shot to few-shot settings and reaching its best results when sufficient task-specific or change-indicator examples were included.

Both the pairwise comparisons and the bootstrap analyses consistently demonstrate that LLM-based methods, particularly Qwen2.5-32B, substantially outperform traditional models. Although their ability to distinguish the worsened category remains comparable to that of traditional models, they achieve significantly better performance across the other outcome categories. Importantly, the value of LLMs in medical annotation extends beyond predictive performance, offering additional practical advantages. The most direct limitation of ImaGenome Silver annotations is their lack of an open, reusable annotation pipeline for efficiently labeling new data, which limits their applicability in evaluation settings.

Conventional annotation methods typically rely on extensive manually curated vocabularies, which require labor-intensive human verification. Meanwhile, these vocabularies are often disease-specific, which limits their general applicability. While LLMs enable automatic and scalable annotation by interpreting context-rich clinical narratives without reliance on predefined lexicons, thereby reducing manual effort and enhancing adaptability across diverse clinical domains.

Some existing approaches, such as TEM and MIMIC-Diff-VQA, extract temporal descriptive terms directly from radiology reports, but do not summarize information on disease progression. This limitation restricts their utility across diverse application scenarios. In contrast, LLMs, with their strong language comprehension and flexibility in understanding clinical narratives, offer a promising avenue for summarizing disease progression from radiology reports.

In conclusion, our experiments show that LLMs significantly outperform traditional rule-based methods in identifying longitudinal descriptions and disease progression in radiology reports. Unlike conventional approaches that require laborious construction and maintenance of large domain-specific lexicons, LLMs reduce human resource costs and enhance task scalability, making them a promising and robust tool for this task.

\subsection{Influence of LLMs scale and training dataset \label{sec:influenceLLMscale}}

Larger LLMs do not necessarily outperform smaller ones in terms of both longitudinal annotation and disease progression. Within the Qwen2.5 series, Qwen2.5-32B and Qwen2.5-72B achieved comparable performance, while across series, LLaMA3.3-70B slightly underperformed compared with Qwen2.5-32B and MedGemma-27B. These findings show that 32B-scale LLMs are sufficient to meet the demands of specific clinical scenarios, offering strong performance alongside greater ease of deployment and lower GPU requirements.

Consistent with the observations of Kim et al. \cite{09kim2025benchmarking} in disease diagnosis, medically fine-tuned language models did not outperform their base models in our annotation tasks. MedResearcher-R1-32B, a medically fine-tuned variant of Qwen2.5-32B, showed lower performance than its base model, suggesting that domain-specific fine-tuning does not always enhance task-specific capabilities. This may reflect limitations in the design of fine-tuning tasks or the coverage of domain-specific data. In addition, the annotation and evaluation frameworks proposed in this study are model-agnostic and not constrained by model size. This enables the incorporation of LLMs beyond the five evaluated in this work.

Rather than larger and more capable commercial models (e.g., ChatGPT and Gemini), this study focuses on open-source models mainly for three reasons:
\begin{itemize}
    \item Reproducibility. Open-source models can be deployed locally at fixed versions, enabling stable and reproducible evaluation settings. In contrast, commercial models are typically proprietary and continuously updated, with earlier versions often deprecated, leading to temporal variability in outputs that may affect reproducibility and comparability.
    \item Accessibility and experimental efficiency. Commercial APIs are often subject to usage quotas; for example, Gemini 3 Flash limits standard paid accounts to 10,000 requests per day. Annotating 1,863 reports at sentence level for longitudinal and disease progression tasks would therefore require multiple API days per account, limiting iterative experimentation. In contrast, locally deployed open-source models avoid external rate limits; Qwen2.5-32B completes full dataset inference in \textasciitilde 1 hour on our hardware (see Appendix \ref{sup:LLM_select}).
    \item Privacy and compliance. Unlike locally deployed open-source models, commercial models are typically accessed via cloud-hosted APIs, thereby introducing additional privacy and regulatory compliance challenges for clinical data management.
\end{itemize}

Furthermore, we conducted preliminary experiments using Gemini 3 Flash API \footnote{https://ai.google.dev/gemini-api/docs} to evaluate the performance of a more powerful commercial model on this task (see Appendix \ref{app:gemini}). While Gemini 3 Flash outperform Qianwen2.5-32B, we observe that for certain error categories (particularly inconsistency errors, see Figure \ref{fig:error}), Gemini 3 Flash frequently relies on experience-based inferences that go beyond the information explicitly stated in the input text. Although such behavior may improve agreement with human annotations, its practical utility in real-world deployment may be limited.

In contrast, to better support the use of LLMs for longitudinal information annotation, an alternative to solely improving performance on existing free-text radiology reports may be to standardize and explicitly structure longitudinal statements and their associated categories within radiology reports. This would reduce the need for models to infer unstated information. We further discuss this issue in Section \ref{sec:limitations}.

\subsection{Impact of annotation errors on model evaluation \label{sec:error_ana}}

Automated annotation tools inevitably contain errors that may affect model evaluation. Among existing approaches, CheXpert is widely used to extract disease labels for 14 findings from both ground-truth and generated reports \cite{wang2025survey}, with F1 scores ranging from 46.2\% to 97.3\% across different findings \cite{johnson2019mimicjpg}. The recently popular RadGraph achieves 94\% entity recognition F1 and 82.3\% relation extraction F1 \cite{zhang2024rexrank, 02jain2021radgraph}. It serves as the foundation for the increasingly popular evaluation metric, RadGraph F1. \cite{zhang2024rexrank, 02jain2021radgraph}. Similarly, our Qwen2.5-32B-based annotation pipeline achieves an F1 score of 91.8\% for longitudinal annotation and 83.2\% for disease progression extraction (Tables \ref{tab:results1} and \ref{tab:results2}). In this section, we analyze the causes of annotation errors, examine their impact on evaluation results, and discuss strategies to mitigate their effects.

Overall, the results are presented in Section \ref{result:error}, and Figure \ref{fig:error} illustrates the distribution of four error types: true errors, hallucinations, inconsistencies, and incorrect corrections. We argue that the impact of these error types on downstream tasks varies, with their severity following the order: true errors > hallucinations > inconsistencies > incorrect corrections. True errors and hallucinations are the most detrimental, while incorrect corrections have a negligible effect. Inconsistencies introduce noise, but have a limited impact. Hallucinations account for only a negligible proportion (one case) and are thus omitted from this section. Nevertheless, as an inherent challenge of large language models, their mitigation is discussed in Section \ref{sec:limitations}.

As shown in Figure \ref{fig:error}, true errors identified in the longitudinal annotation task were primarily associated with expressions containing implicit semantics, accounting for only 1.4\% of the entire dataset. The true errors identified in the disease progression task are more diverse, but they also represent a small proportion of the entire dataset (1.3\%). Although improved handling of such cases may further reduce residual errors and benefit downstream applications, the current error rate remains acceptable. These errors may potentially be mitigated through further improvements in LLM capabilities. Section \ref{sec:limitations} discusses future work on improving the annotation capabilities of LLMs in detail.

Notably, Qwen2.5-32B made no misclassifications in longitudinal annotation when references to prior examinations were explicitly provided. Similarly, most errors in disease progression annotation arose from ambiguous wording rather than failures in temporal reasoning. These findings indicate that the model can reliably interpret longitudinal information when temporal relationships are clearly specified, underscoring the importance of structured reporting practices that explicitly document prior examinations and comparative findings. Such practices may enhance longitudinal information extraction and reasoning by large language models.

Inconsistencies errors account for a large proportion in both tasks. However, we believe they have a limited impact on model evaluation, mainly because Qwen2.5-32B classifies almost all of them as non-longitudinal sentences (see Table~\ref{tab:L_error}). In this case, treating the ground-truth reports and generated reports consistently ensures a fair evaluation.

The sentences associated with inconsistency errors can often be interpreted from a single image rather than requiring longitudinal comparison. For example, ``Vague opacity at the right base may be due to atelectasis or developing infiltrate.'' Such sentences mainly describe diagnostic information rather than disease progression. Furthermore, some inconsistency errors arise from limitations of the dataset itself rather than the capability of the LLM. For example, phrases such as ``comparison is difficult'' provide limited clinical information and do not directly reflect longitudinal changes.

Since ImaGenome Gold underwent inter-annotator agreement verification, the occurrence of incorrect corrections warrants further investigation. We analyzed the causes of the 15 incorrect corrections identified during the longitudinal annotation process, which accounted for 0.7\% of the 1,975 sentences. All reports in the gold dataset were obtained from patients’ second visits in MIMIC-CXR, and annotators were instructed to use the corresponding first-visit reports as references when assigning longitudinal labels. Due to incomplete prior documentation and variations in examination focus, certain medical findings may not be reported consistently across visits. Consequently, findings described as changed relative to prior examinations in the current report may not have been labeled as progression if the corresponding abnormality was not documented in the first-visit report.

We further analyzed the annotation results generated by Qwen2.5-32B on 500 manually annotated reports by disease type. Some disease descriptions may contain relatively fixed expressions, which could introduce apparent disease-specific patterns in the annotation results to some extent. However, Figure \ref{fig:error} shows that most errors are strongly associated with change indicators rather than specific disease categories. Therefore, the disease-wise results are provided in Appendix \ref{app:diseaseperformance} for reference only.

To assess the impact of LLM stochasticity on evaluation reliability, we conducted five rounds of evaluations on seven automated report generation models using the same stochastic decoding configuration (temperature=0.7, top-p=0.9, and top-k=20) described in Section \ref{sec:med_decoding}. Detailed results are presented in Appendix \ref{app:rankings}. Kendall’s coefficient of concordance and Spearman’s rank correlation coefficient both indicate high consistency across repeated evaluations, suggesting that the stochasticity introduced by this sampling configuration has limited impact on model rankings. While stochastic sampling is useful for robustness analysis, we recommend greedy decoding for practical evaluations to ensure reproducibility.

In addition, we recognize that using Qwen2.5-32B to annotate both the GT reports and the generated reports may introduce a potential evaluator-induced bias. To assess the extent to which such bias may affect the reliability of model rankings, we conducted two additional robustness experiments. Detailed descriptions and results are provided in Appendix \ref{app:rankings}. First, motivated by previous findings that commercial LLM-based evaluations can achieve strong agreement with human judgments in open-ended text generation tasks \cite{zheng2023judging}, together with the superior performance of Gemini 3 Flash on the longitudinal annotation task (Appendix \ref{app:gemini}), we replaced Qwen2.5-32B with Gemini 3 Flash as the annotator for reference reports. The resulting model rankings remained highly consistent according to statistical rank correlation measures, suggesting that the relative performance rankings are robust to noise introduced by different LLM annotators. Second, we evaluated the agreement between LLM-based annotations and human annotations on a randomly selected subset of 100 samples from the test set. The rankings produced by the LLM evaluator showed strong and statistically significant correlations with human annotations, supporting the reliability of the proposed evaluation framework.


In summary, we believe that the errors introduced by using Qwen2.5-32B remain within an acceptable range and have a limited impact on the overall evaluation of report generation models.

\subsection{Evaluation of medical report generation models}

We evaluate the progress and limitations of current report generation models in longitudinal information generation by validating seven mainstream models on the L-MIMIC dataset, based on our evaluation framework (Table \ref{table:boot_evaluation}). The results are presented in Section \ref{result:reportgeneration}. Overall, recent models have incorporated prior image and text data via diverse strategies, achieving notable progress in longitudinal information capture. However, considerable room for enhancement remains, specifically in the coverage and fluency of longitudinal sentences, and the ability to identify disease improvement and deterioration.

Our evaluation framework considers language-based metrics and disease progression–based metric for longitudinal information. For language-based metrics, we use BLEU, ROUGE-L, and METEOR, which are widely used in radiology report generation and image captioning \cite{wang2025survey, ghandi2023deep}. These overlap-based measures capture surface-level lexical similarity between generated and reference texts, approximating content coverage. However, relying solely on these metrics has two main limitations. First, minor lexical variations can lead to substantial score fluctuations, especially in short texts. Second, these metrics fail to capture clinically meaningful semantics. For example, ``no radiographic evidence of interval progression'' and ``stable chest findings'' are clinically equivalent but receive low similarity scores. Conversely, ``no interval change in cardiac silhouette size'' and ``interval increase in cardiac silhouette size'' convey opposite meanings yet may still receive high scores.

Despite these limitations, language generation metrics have not been deprecated and remain useful for evaluating fluency and content coverage \cite{r1hou2025mambagen,r2hou2025recalibrated,r3hou2023mkcl,r4song2026mixture}. The recent composite metric RadCliQ-v1 \cite{yu2023evaluating} also incorporates BLEU as a component. In our experiments, full reports and longitudinal sentences exhibit clear performance differences under language-based metrics. Although longitudinal sentences are typically shorter and more sensitive to minor lexical variations, the observed gap still suggests that existing models struggle to adequately express longitudinal information.

To address the limitation that language-based metrics fail to capture clinically meaningful disease progression, we introduce disease progression–based evaluation metrics derived from disease progression annotations. These metrics assess whether generated reports reflect clinically important disease progression. We recommend jointly using both language-based and disease progression–based metrics for a more comprehensive evaluation of automated report generation models.

Report generation models typically exhibit three types of errors when handling longitudinal information: missing comparison, redundant comparison, and wrong trend prediction. A missing comparison occurs when documented lesion changes are omitted; a redundant comparison introduces unreported alterations; and a wrong trend prediction misrepresents the direction of disease progression. Our evaluation framework can detect all these errors. Specifically, progression metrics provide an intuitive reflection of wrong trend errors and missing comparison errors. Language-based metrics for longitudinal sentences can distinguish linguistic descriptive discrepancies, thereby capturing redundant comparisons and missing comparison errors, as well as partially indicating wrong trend errors. 

Trend errors stem from the limitations of the model's capabilities, whereas redundant and missing comparisons arise when the model fails to recognize the target disease. Even radiologists interpreting identical imaging data may produce reports with inconsistent documentation priorities without explicit guidance—this underscores the necessity of integrating keyword guidance into report generation models to delineate required information for extraction. While keyword-driven information retrieval has been embedded in general report generation frameworks \cite{tanida2023interactive, muller2024chex}, analogous research focusing on longitudinal information capture remains sparse.

Our evaluation metrics are highly sensitive to missing comparisons, with both language-based and disease progression–based metrics substantially degraded by such errors. This is expected, as missing comparisons directly obscure longitudinal disease progression, which is the primary capability that our evaluation is designed to assess.

In contrast, the metrics are relatively insensitive to redundant longitudinal comparisons. Their impact is mainly limited to language-based metrics, as these metrics depend on the coverage of comparison statements. Disease progression-based metrics remain unaffected, as they do not evaluate redundant disease status. Accurately quantifying redundant longitudinal comparisons is challenging, even with human evaluation, due to the lack of ground-truth longitudinal comparisons and expert radiologist annotations.

The sensitivity to trend errors falls between these two cases and is mainly reflected in disease progression-based metrics. As discussed in Section \ref{sec:error_ana}, true errors account for only 1.4\% of cases in the longitudinal annotation stage. Therefore, the likelihood of missing disease progression due to incorrect identification of longitudinal sentences is low. However, during the disease progression annotation stage, Qwen2.5-32B tends to assign the no change label in ambiguous cases. As a result, an automated report generation model that frequently produces statements such as “all findings remain unchanged” may achieve higher performance on the no change metric. However, these types of errors will be reflected in both the improved and worsened categories; therefore, it is necessary to consider the model’s performance across each category.

Report generation models show a significant performance gap in longitudinal metrics compared with language-based and diagnostic metrics for full reports. This suggests that the models can capture clinical findings; however, they struggle to correlate these with patients’ historical records. Previous report generation models (e.g., R2Gen and MedVersa) typically take single-time-point images as input, which do not contain longitudinal information. Longitudinal descriptions generated by these models are hallucinations that lack clinical value.

Recent models for medical report generation incorporate prior images and reports as inputs, achieving improved performance on longitudinal metrics compared to baseline approaches. However, existing fusion strategies remain rudimentary. Most methods rely on cross-attention mechanisms to retrieve relevant information in prior inputs \cite{3zhu2023utilizing,13zhang-etal-2025-libra,5liu2025enhanced}. Yet, none of these methods introduce  constraints when associating current observations with historical patient data. Liu et al. \cite{7liu2025hc} attempt to compare the current image with prior images, as well as the current report with prior reports, by imposing consistency constraints, but they keep the LLM frozen and only train a linear projection layer, potentially limiting expressive power. Furthermore, we observed a substantial gap in the report generation model’s performance on generating improved and worsened sentences compared with unchanged sentences. This discrepancy may be attributed to data imbalance, which could potentially be mitigated through data augmentation techniques such as resampling.

In conclusion, our evaluation of the longitudinal information generation capability refines the assessment framework for report generation models and identifies the limitations of current models. Incorporating additional constraints, such as keywords, to better integrate prior images and reports into the training process represents a promising direction for future report generation research.

\subsection{Other application scenarios of automated longitudinal information annotation in radiology reports \label{sec:scenario}}

Besides the evaluation scenario, automated longitudinal information extraction has two primary application scenarios that support both clinical practice and research. First, it enables structured characterization of disease trajectories from free-text radiology reports by converting narrative descriptions into progression categories (e.g., improved, no change, worsened). Structuring longitudinal and disease-tracking information can support quantification of population-level trends, identification of clinically meaningful temporal changes, and retrieval of cases for large-scale analysis. 

Second, automatic longitudinal annotation can enhance the training of automated report generation models by providing additional supervisory signals that indicate whether prior examinations should be referenced, more accurately reflecting real-world clinical workflows. Current models treat longitudinal and cross-sectional sentences identically, using the same input. In reality, longitudinal sentences require reference to previous images and reports, whereas cross-sectional sentences can be generated from single time-point images. Zhu et al. \cite{3zhu2023utilizing} selected longitudinal reports from patients with more than two visits, assuming these contain longitudinal information. However, our annotations (Figure \ref{res:data_annotation}) reveal that 10–20\% of such reports lack longitudinal content and around 58\% of sentences in longitudinal reports are cross-sectional sentences. Therefore, explicit differentiation of cross-sectional from true longitudinal sentences is critical to enabling models to prioritize high-value input data.

Disease progression labels (e.g., no change, improved, worsened) can also provide valuable supervisory signals for automatic report generation models, enabling more accurate longitudinal report generation. Prior studies have introduced auxiliary tasks based on disease diagnosis labels derived from automated labeling tools such as CheXpert and incorporated them into model training, achieving promising performance improvements \cite{edirisinghe2025chest,wang2025survey}. These results suggest that disease progression labels could similarly benefit longitudinal report generation, although their potential has not yet been thoroughly explored.

Additionally, automatic longitudinal annotation supplies high-quality training data for model pre-training, which is crucial for deep learning, especially for large-scale models. Pre-training transforms model parameters from random initialization into a semantically informed starting point, laying a robust foundation for effective fine-tuning. For example, Zhang et al. \cite{13zhang-etal-2025-libra} utilized VQA annotations from MIMIC-Diff-VQA \cite{07hu2023expert} to pre-train their report generation model.

\subsection{Limitations and future works \label{sec:limitations}}

We analyzed the limitations and future directions from three perspectives: the dataset, prompt design, and clinical feasibility.

Although our evaluation follows the Imagenome benchmark protocol and includes 500 reports with multi-annotator annotations, the test set size may still limit the coverage of diverse clinical findings, particularly rare abnormalities. This scale is comparable to those used in prior validation studies. For example, the CheXpert-based labeling approach applied to MIMIC-CXR was validated on 687 manually annotated reports \cite{johnson2019mimicjpg, wang2025survey}, and the recently popular annotation tool RadGraph was validated on 100 manually annotated reports \cite{02jain2021radgraph}. Future work should curate and validate larger manually annotated datasets to improve the representation of rare findings and further strengthen evaluation reliability.

From a prompt design perspective, this study developed a prompt configuration that yielded strong performance on the gold test set, suggesting the feasibility of using LLMs for longitudinal sentence identification and disease progression annotation. During prompt design, to prevent data leakage, we restricted prompt development to iterative manual refinement using an independently sampled subset of MIMIC-CXR. This process focused on rephrasing instructions to minimize ambiguity and reduce erroneous interpretations by the LLM. The use of an independently sampled subset ensured that the gold test set remained completely untouched until the final evaluation. We then report the final performance on the untouched gold test set to evaluate the robustness, effectiveness and generalizability of the prompt. However, we note that no dedicated validation set is available for prompt development. Therefore, we are unable to report quantitative results for prompt tuning and directly provide the prompt used in our experiments. This limitation is not unique to longitudinal information extraction but is also prevalent in many LLM-based tasks, such as diagnosis \cite{09kim2025benchmarking} and structured reporting \cite{08woznicki2025automatic}. Future research could investigate more effective iterative prompt refinement strategies that allow for quantitative evaluation beyond interaction-based prompt optimization.

From a clinical perspective, the annotation framework proposed in this study has been applied to evaluate automated report generation models, thereby supporting their future clinical translation and implementation. In addition, if statistical analyses are required to compare report generation model performance, it is important to consider that the primary outputs are free-text reports. In our evaluation framework, Qwen2.5-32B was used to extract information from both ground-truth and generated reports, producing categorical progression labels (no-change, improved, or worsened) in text format. Since the extracted outcomes are text-based categorical labels rather than continuous scores or probability estimates, conventional parametric and non-parametric tests, such as the T-test or Wilcoxon signed-rank test, cannot be applied directly to these classification outputs. We therefore recommend utilize bootstrap resampling to assess statistical significance when comparing different report generation models on L-MIMIC.

If further adapted for structured report generation in clinical practice, several key issues warrant additional investigation.

First, this study relies on publicly available datasets, whereas real-world clinical deployment is more complex. Clinical reports from different sources often vary in terminology, style, and level of detail, which may affect the robustness of LLMs. Future work should address this by building multi-institutional real-world datasets for training and evaluation. In addition, standardizing clinical narratives by mapping free-text reports to structured representations or controlled vocabularies (e.g., OMOP \cite{de2025llm}) could reduce inter-institutional variability, ease language understanding, and improve model stability.

Second, the validation in this study is limited to chest X-ray reports and does not include other imaging modalities such as CT or MRI. Although the proposed LLM-based annotation framework operates on textual reports rather than imaging data, its performance may still be affected by modality-specific reporting styles and clinical contexts. Differences in disease prevalence, imaging indications, and reporting granularity across modalities may introduce distribution shifts that have not been systematically evaluated. Moreover, high-quality expert-annotated datasets spanning multiple modalities and anatomical regions remain limited. Therefore, constructing such datasets and further evaluating the robustness of LLM-based annotations across diverse clinical settings are important next steps.

Third, although hallucination errors are relatively infrequent in our error analysis, they remain a non-negligible limitation of LLMs, particularly in domains such as rare diseases where training data are scarce. Recent advances, including enabling LLMs to access the internet, have helped mitigate hallucinations and outdated knowledge. A promising research direction is therefore the development of LLM-based annotation methods that support online querying or real-time knowledge updates. In addition, incorporating interpretability signals (Figure \ref{fig:explainability}) and adopting human-in-the-loop workflows may further reduce hallucination risks. As shown in Figure \ref{fig:explainability}, large models can already generate explanations, which may improve user understanding and facilitate clinical adoption.

Fourth, in terms of performance, as shown in the error analysis in Section \ref{sec:error_ana}, many errors appear to be associated with unclear change-indicating prompts. This limitation may be mitigated through two potential strategies: (1) future work could explore more nuanced prompt engineering, for example by incorporating prior clinical reports to provide richer contextual information; (2) structuring longitudinal information in clinical reports, explicitly specifying visit record identifiers and the corresponding changes, may help reduce reasoning errors arising from ambiguous longitudinal context, thereby improving clinical usability.

Finally, real-world clinical deployment continues to face several practical constraints, including computational cost, regulatory approval, and data privacy concerns. Open-source models deployed locally may offer enhanced privacy guarantees compared to proprietary commercial APIs.

\section{Conclusions}
Recent automated report generation methods have increasingly focused on designing specialized components to capture longitudinal information; however, appropriate evaluation tools remain scarce. In this study, we present an LLM-based approach for annotating longitudinal information in radiology reports, enabling automated extraction and monitoring of disease progression. Evaluation across five LLMs demonstrates substantial improvements over traditional rule-based methods, significantly reducing manual effort while enhancing scalability. Considering efficiency and performance, we selected Qwen-32B to annotate the MIMIC-CXR dataset and, based on this, developed a standardized framework to evaluate chest X-ray report generation models on their ability to capture longitudinal information. Applying this framework to seven representative models, we analyzed their strengths and limitations, offering insights to guide future model development and evaluation strategies.


\section*{Data Availability Statement}
Data Usage and Sources: The study utilizes two primary datasets which were derived from the public MIMIC-CXR dataset: https://physionet.org/content/mimic-cxr/2.1.0/. 

(1) Chest ImaGenome: A subset of this dataset, comprising 1,975 manually annotated sentences from 500 reports, was used to evaluate and compare the performance of various Large Language Models (LLMs).

(2) MIMIC-CXR: This public dataset served as the source for 95,169 radiology reports that were subsequently annotated using the Qwen2.5-32B model to create a new benchmark called L-MIMIC.

Newly Generated Data: The researchers created the L-MIMIC dataset, which provides standardized longitudinal annotations for the MIMIC-CXR reports. It includes 73,093 training cases, 588 validation cases, and 1,863 test cases.

Data Availability and Access: Both the code and the L-MIMIC dataset will be made publicly available and updated on GitHub upon acceptance of the paper.

All LLMs used for the annotation (MedGemma-27B, MedResearcher-R1-32B, Qwen2.5-32B, Llama3.3-70B, and Qwen2.5-72B) are open-source and freely available.

\section*{Conflict of Interest}
The authors declare no competing interests.

\printcredits


\bibliography{cas-refs}


\begin{appendices}

\section{Details on longitudinal information annotation with LLMs}
\subsection{\label{disease_list} Design of the disease list for prompt construction}

The disease list was generated by deduplicating all disease annotations in the ImaGenome gold set, yielding 51 unique terms. The terms “normal” and “abnormal” were excluded because they do not correspond to specific disease entities. A category for support devices was subsequently added. Changes in support devices, particularly positional changes, do not indicate disease progression and are therefore not annotated as disease status changes in ImaGenome. Nevertheless, information on changes in support devices often requires reference to prior cases, justifying its inclusion.

The final list contains 50 terms, as follows:

\textit{[`support devices',`airspace opacity', `alveolar hemorrhage', `aspiration', `atelectasis', `bone lesion', `bronchiectasis', `calcified nodule', `clavicle fracture', `consolidation', `copd/emphysema', `costophrenic angle blunting', `elevated hemidiaphragm', `enlarged cardiac silhouette', `enlarged hilum', `fluid overload/heart failure', `goiter', `granulomatous disease', `hernia', `hydropneumothorax', `increased reticular markings/ild pattern', `infiltration', `interstitial lung disease', `lobar/segmental collapse', `low lung volumes', `lung cancer', `lung lesion', `lung opacity', `mass/nodule (not otherwise specified)', `mediastinal displacement', `mediastinal widening',  `opacity', `pericardial effusion', `pleural effusion', `pleural/parenchymal scarring', `pneumonia', `pneumothorax', `pulmonary edema/hazy opacity', `rib fracture', `scoliosis', `shoulder osteoarthritis', `spinal degenerative changes', `spinal fracture', `sub-diaphragmatic air', `subcutaneous air', `superior mediastinal mass/enlargement', `tortuous aorta', `vascular calcification', `vascular congestion', and `vascular redistribution']}.

The design of this keyword list is guided by two considerations. First, it maintains consistency with the ImaGenome annotation scheme to ensure comparability, as identical descriptions may reflect either improvement or deterioration across different clinical findings. Second, as detailed in Appendix \ref{app:diseaseperformance}, these keywords cover 10 clinically relevant chest X-ray categories: (1) pulmonary parenchymal/airspace abnormalities, (2) interstitial and chronic lung disease, (3) lung volume loss/atelectasis, (4) pleural disease and pneumothorax, (5) cardiac and pericardial abnormalities, (6) vascular changes, (7) mediastinal/hilar/thyroid abnormalities, (8) nodules/masses/tumors, (9) bone and chest wall findings, and (10) other, thereby providing comprehensive coverage of radiographic manifestations. However, we acknowledge that it may remain insufficiently systematic and comprehensive for structured reporting. We encourage further development of standardized reporting templates and high-quality corresponding annotations.

\subsection{Analysis of large models selection}\label{sup:LLM_select}

Deep learning models typically require large-scale datasets for effective training and evaluation. Therefore, when selecting a LLM for report annotation, it is essential to balance accuracy and inference efficiency: accuracy determines the reliability of annotations, whereas efficiency affects the feasibility of deployment in real-world clinical workflows.

As shown in Table \ref{tab:results1} and Table \ref{tab:results2}, Qwen2.5-32B and MedGemma-27B outperform other evaluated models as described in Section \ref{sec:results}. Regarding stability, as shown in Figure \ref{fig:samp}, although all models demonstrate relatively consistent performance across five random samplings, Qwen2.5-32B exhibits the highest stability among all evaluated models. In this section, we benchmarked their inference speeds using the longitudinal annotation task as an example (Table \ref{tab:inference_speed}). Specifically, 100 sentences were randomly sampled from the test set, and all experiments were conducted on NVIDIA A100 GPUs with FlashAttention2 acceleration under identical configurations.

The results indicate that MedGemma-27B generates substantially longer reasoning chains per query, resulting in an overall runtime 15.7× longer than that of Qwen2.5-32B. This finding suggests that Qwen2.5-32B offers a more favorable trade-off between accuracy and efficiency, making it the more practical choice for real-world applications.

\begin{table}[!t]

\caption{\label{tab:inference_speed}Inference speed of each model on the longitudinal annotation task. Time and tokens are averages per query.}
\centering
\begin{tabular}{>{\raggedright\arraybackslash}m{2cm}
            >{\centering\arraybackslash}m{0.5cm}
            >{\centering\arraybackslash}m{1.5cm}
            >{\centering\arraybackslash}m{1cm}
            >{\centering\arraybackslash}m{1cm}}
\toprule
Model & GPU\# & Speed (tokens/s) & Tokens\# & Time (s) \\
\midrule
MedGemma-27B & 1 & 11.34 & 358.58 & 31.62 \\
\midrule
MedResearcher-R1-32B & 1 & 19.38 & 67.79 & 3.50 \\
\midrule
Qwen2.5-32B & 1 & 18.45 & 38.23 & 2.07 \\
\midrule
\multirow{2}{*}{LLama3.3-70B} & 1 & 0.42 & \multirow{2}{*}{276.73} & 658.88 \\
                               & 2 & 24.53 &                          & 11.28 \\
\midrule
\multirow{2}{*}{Qwen2.5-72B} & 1 & 0.13 & \multirow{2}{*}{56.02} & 430.9 \\
                               & 2 & 9.40 &                          & 5.96 \\
\bottomrule
\end{tabular}
\end{table}

\subsection{Zero-shot and few-shot prompting} \label{sec:zeroshot}

In this section, we describe the prompts used to evaluate Qwen2.5-32B on longitudinal and disease progression annotation tasks under zero-shot and few-shot settings. In the zero-shot setting, the model generates annotations from instructions without any task-specific examples. In the few-shot setting, we vary the number of task examples and change-indicator examples. For longitudinal annotation, we independently vary the number of task examples (0, 3, 5) and change-indicator examples (0, 4, 8, 12). When varying one factor, the other is fixed to its best-performing setting (task examples: 3; change-indicator examples: 8), identified through preliminary experiments. For disease progression annotation, we use 0, 2, and 4 task examples. Detailed prompt designs are provided in Table \ref{apptab:example}.

\begin{table*}[!htbp]
\centering
\caption{Prompts used for zero-shot and few-shot evaluation of Qwen2.5-32B on longitudinal annotation and disease progression annotation tasks. * indicates the final prompt used.}
\label{apptab:example}

\newlength{\colA}
\newlength{\colB}
\newlength{\colC}

\setlength{\colA}{2.7cm} 
\setlength{\colB}{1.3cm} 
\setlength{\colC}{12cm} 

\begin{tabular}{|>{\raggedright\arraybackslash}p{\colA}|
                >{\raggedright\arraybackslash}p{\colB}|
                >{\raggedright\arraybackslash}p{\colC}|}
\hline

Task & Setting & Prompt \\ \hline

\multirow{3}{\colA}{Longitudinal annotation prompts using different numbers of task examples}
& 0-shot & You are a medical AI assistant. Given a radiology report sentence \textbf{sentence}, determine if it compares the current image with prior studies (e.g., remain, compare, similar, stable, increased, still, new, again). If yes, return \textless1\textgreater. If no, return \textless0\textgreater. \\ \cline{2-3}
& 3-shot* & You are a medical AI assistant. Given a radiology report sentence \textbf{sentence}, determine if it compares the current image with prior studies (e.g., remain, compare, similar, stable, increased, still, new, again). If yes, return \textless1\textgreater. If no, return \textless0\textgreater. Examples: ``Cardiac and mediastinal silhouettes are stable.'' returns \textless1\textgreater, ``No larger pleural effusions.'' returns \textless1\textgreater,``Increased retrocardiac opacity may reflect atelectasis.'' returns \textless1\textgreater. \\ \cline{2-3}
& 5-shot & You are a medical AI assistant. Given a radiology report sentence \textbf{sentence}, determine if it compares the current image with prior studies (e.g., remain, compare, similar, stable, increased, still, new, again). If yes, return \textless1\textgreater. If no, return \textless0\textgreater. Examples: "Cardiac and mediastinal silhouettes are stable." returns \textless1\textgreater, ''No larger pleural effusions.'' returns \textless1\textgreater,''Increased retrocardiac opacity may reflect atelectasis.'' returns \textless1\textgreater. ''Previously questioned area of right upper lobe pneumonia is still abnormal.'' returns \textless1\textgreater. ''Lower lung volumes seen on the current exam.'' returns \textless0\textgreater. \\ \hline

\multirow{4}{\colA}{Longitudinal annotation prompts using different numbers of change-indicator examples}
& 0-shot & You are a medical AI assistant. Given a radiology report sentence \textbf{sentence}, determine if it compares the current image with prior studies. If yes, return \textless1\textgreater. If no, return \textless0\textgreater. Examples: ``Cardiac and mediastinal silhouettes are stable.'' returns \textless1\textgreater, ``No larger pleural effusions.'' returns \textless1\textgreater,``Increased retrocardiac opacity may reflect atelectasis.'' returns \textless1\textgreater. \\ \cline{2-3}
& 4-shot & You are a medical AI assistant. Given a radiology report sentence \textbf{sentence}, determine if it compares the current image with prior studies. If yes, return \textless1\textgreater. If no, return \textless0\textgreater. Examples: ``Cardiac and mediastinal silhouettes are stable.'' returns \textless1\textgreater, ``No larger pleural effusions.'' returns \textless1\textgreater,``Increased retrocardiac opacity may reflect atelectasis.'' returns \textless1\textgreater. \\ \cline{2-3}
& 8-shot* & You are a medical AI assistant. Given a radiology report sentence \textbf{sentence}, determine if it compares the current image with prior studies (e.g., remain, compare, similar, stable, increased, still, new, again). If yes, return \textless1\textgreater. If no, return \textless0\textgreater. Examples: ``Cardiac and mediastinal silhouettes are stable.'' returns \textless1\textgreater, ``No larger pleural effusions.'' returns \textless1\textgreater,``Increased retrocardiac opacity may reflect atelectasis.'' returns \textless1\textgreater. \\ \cline{2-3}
& 12-shot & You are a medical AI assistant. Given a radiology report sentence \textbf{sentence}, determine if it compares the current image with prior studies (e.g., remain, compare, similar, stable, increased, still, new, again, better, worsening, developing, change). If yes, return \textless1\textgreater. If no, return \textless0\textgreater. Examples: ``Cardiac and mediastinal silhouettes are stable.'' returns \textless1\textgreater, ``No larger pleural effusions.'' returns \textless1\textgreater,``Increased retrocardiac opacity may reflect atelectasis.'' returns \textless1\textgreater. \\ \hline

\multirow{3}{\colA}{Disease progression annotation prompts using different numbers of task examples}
& 0-shot & The following sentence \textbf{sentence} describes the status of \textbf{disease} as \textless no change\textgreater (e.g., similar, unchanged), \textless improved\textgreater (e.g., resolved), or \textless worsened\textgreater. Return the answer in the form of \textless no change\textgreater, \textless improved\textgreater , or \textless worsened\textgreater. If this condition is not mentioned, return \textless unmentioned\textgreater. Note that if `change' or `new' is mentioned along with a newly appearing lesion, it usually implies worsening. \\ \cline{2-3}
& 2-shot* & The following sentence \textbf{sentence} describes the status of \textbf{disease} as \textless no change\textgreater (e.g., similar, unchanged), \textless improved\textgreater (e.g., resolved), or \textless worsened\textgreater. Return the answer in the form of \textless no change\textgreater, \textless improved\textgreater , or \textless worsened\textgreater. If this condition is not mentioned, return \textless unmentioned\textgreater. Note that if `change' or `new' is mentioned along with a newly appearing lesion, it usually implies worsening. Example: in cases of `pleural effusion', `New small bilateral pleural effusions.' means \textless worsened\textgreater. In cases of `low lung volumes', `lower' means \textless worsened\textgreater; `increased' means \textless improved\textgreater. \\ \cline{2-3}
& 4-shot & The following sentence \textbf{sentence} describes the status of \textbf{disease} as \textless no change\textgreater (e.g., similar, unchanged), \textless improved\textgreater (e.g., resolved), or \textless worsened\textgreater. Return the answer in the form of \textless no change\textgreater, \textless improved\textgreater , or \textless worsened\textgreater. If this condition is not mentioned, return \textless unmentioned\textgreater. Note that if `change' or `new' is mentioned along with a newly appearing lesion, it usually implies worsening. Example: in cases of `pleural effusion', `New small bilateral pleural effusions.' means \textless worsened\textgreater. In cases of `low lung volumes', `lower' means \textless worsened\textgreater; `increased' means \textless improved\textgreater. In case of \'pneumonia\', \'pneumonia is still abnormal\' means <no change>. In case of \'copd/emphysema\', \'Severe emphysema worse in the right lower lobe is better evaluated in prior CT.\'  means <worsened>. \\ \hline

\end{tabular}

\end{table*}

\subsection{Performance of Gemini on longitudinal and disease progression annotation\label{app:gemini}}

We evaluate Gemini 3 Flash, a representative proprietary large language model, on a manually annotated dataset comprising 1,975 sentences sampled from 500 reports. All evaluations are conducted via the official API (\url{https://ai.google.dev/gemini-api/docs}). The results are reported in Tables~\ref{gemini1} and~\ref{gemini2}. Despite the favorable results, this work focuses on open-source large language models for practical applicability, as discussed in Section \ref{sec:influenceLLMscale}.

\begin{table*}[!htbp]
\centering
\caption{Performance of Gemini 3-flash on longitudinal annotation.}\label{gemini1}

\begin{tabular}{|>{\raggedright\arraybackslash}m{1.8cm}|
                >{\raggedright\arraybackslash}m{2.6cm}|
                >{\raggedright\arraybackslash}m{2cm}|
                >{\raggedright\arraybackslash}m{2cm}|
                >{\raggedright\arraybackslash}m{2cm}|
                >{\raggedright\arraybackslash}m{2cm}|}
\hline
\multicolumn{2}{|c|}{}&Accuracy&Precision&Recall&F1 score\\
\hline
Traditional method&ImaGenome silver&90.2&98.5&71.2&82.7\\
\hline
Proprietary LLM & Gemini3-flash 
& 95.8 (\textcolor{myRed}{5.6$\uparrow$})
& 90.9 (\textcolor{myBlue}{7.6$\downarrow$})
& 96.9 (\textcolor{myRed}{25.7$\uparrow$})
& 93.8 (\textcolor{myRed}{11.1$\uparrow$})\\
\hline
\end{tabular}
\end{table*}

\begin{table*}[!htbp]
\centering
\caption{Performance of Gemini 3-flash on disease progression annotation.}\label{gemini2}

\begin{tabular}{|>{\raggedright\arraybackslash}m{1.6cm}|
                >{\raggedright\arraybackslash}m{2.6cm}|
                >{\raggedright\arraybackslash}m{1.75cm}|
                >{\raggedright\arraybackslash}m{1.75cm}|
                >{\raggedright\arraybackslash}m{1.75cm}|
                >{\raggedright\arraybackslash}m{1.75cm}|
                >{\raggedright\arraybackslash}m{1.75cm}|}
\hline
\multicolumn{2}{|c|}{}&\multirow{2}{*}{Micro Acc.}&\multirow{2}{*}{Micro F1}&\multicolumn{3}{c|}{Per-class F1}\\
\cline{5-7}
\multicolumn{2}{|c|}{}&&&No change&Improved&Worsened\\
\hline
Traditional method&ImaGenome silver&88.7&78.5&82.1&76.6&72.5\\
\hline
Proprietary LLM&Gemini3-flash
&94.5 (\textcolor{myRed}{5.8$\uparrow$})
&90.0 (\textcolor{myRed}{11.5$\uparrow$})
&90.7 (\textcolor{myRed}{8.6$\uparrow$})
&92.8 (\textcolor{myRed}{16.2$\uparrow$})
&87.3 (\textcolor{myRed}{14.8$\uparrow$})\\
\hline
\end{tabular}
\end{table*}

\subsection{Assessing the representativeness of the 500 manual annotated reports} \label{app:rep}

As noted in Section \ref{sec:data}, the ImaGenome Gold dataset \cite{ImaGenomewu2021chest} is a randomly sampled subset of MIMIC-CXR and therefore shares the same underlying data distribution as our proposed L-MIMIC dataset. This common provenance makes it an appropriate test dataset for evaluating LLM-generated annotations while supporting the generalizability of the evaluation results to the broader MIMIC-CXR cohort.

To further assess the representativeness of the 500-report subset used for the longitudinal information annotation task, we compared its distributions of longitudinal sentences, clinical keywords, and disease progression categories (no change, improved, and worsened), as annotated by Qwen2.5-32B, with those of the full L-MIMIC dataset. Longitudinal sentences constituted 33\% (653/1,975 sentences) of the evaluation subset. This is highly consistent with the proportion of longitudinal sentences in follow-up reports from MIMIC-CXR, as shown in Figure~\ref{res:data_annotation}. In the training dataset, longitudinal sentences (L\_sentences) account for $41\% \times 79\% = 32\%$ of all sentences.

\begin{sidewaysfigure*}
    \centering
    \includegraphics[width=0.95\textheight]{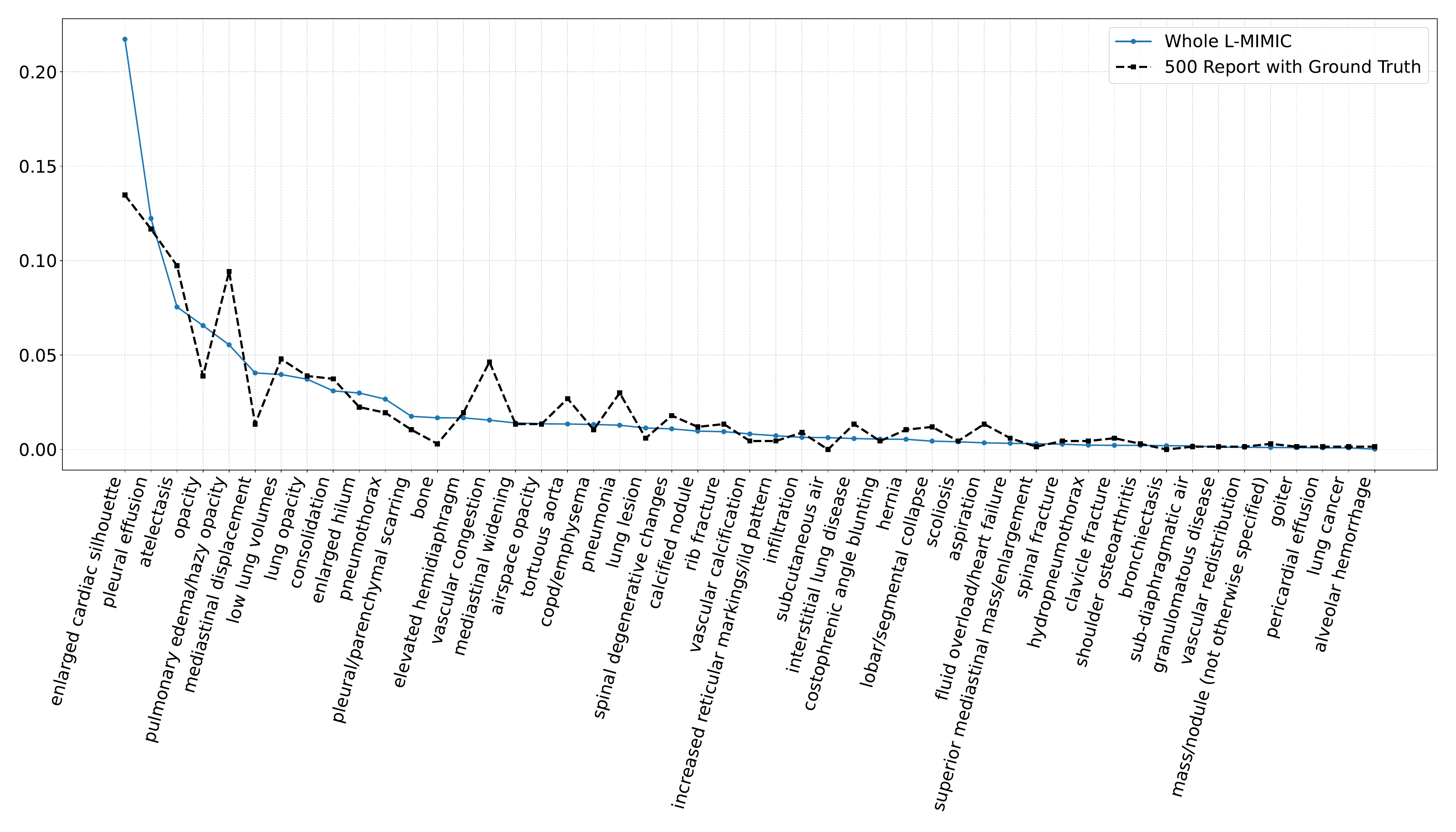}
    \caption{Distribution of finding-related keywords in longitudinal sentences from 500 manually annotated reports and the whole L-MIMIC dataset. All annotations were generated using Qwen2.5-32B. The x-axis lists the finding terms, and the y-axis shows their percentage of occurrences in the dataset.Qwen2.5-32B.} \label{appfig:distribution}
\end{sidewaysfigure*}

The distribution of clinical keywords was highly similar to that of the full dataset, with a Jensen--Shannon distance \cite{lin1991divergence, menendez1997jensen} of 0.18. The distance was computed using the jensenshannon function provided by the SciPy library. Since Jensen--Shannon divergence compares probability distributions rather than raw counts, keyword counts were normalized into probability distributions before comparison. This normalization also accounts for the substantial difference in sample size between the 500 sampled reports and the full L-MIMIC dataset and enables comparison of relative keyword frequencies. Figure \ref{appfig:distribution} illustrates the two distributions. The overall shape and peaks line up well, with only minor variations or slight shifts. Since the SciPy implementation defines Jensen--Shannon distance as the square root of Jensen--Shannon divergence, this corresponds to a Jensen--Shannon divergence of approximately $0.0324$ ($0.18^2$). Under the natural logarithm convention used by SciPy, the theoretical range of Jensen--Shannon divergence is $[0, \ln 2]$. The small divergence value suggests that the overall keyword distribution in the 500 sampled reports is highly similar to that of the full L-MIMIC dataset. For the three major disease progression categories, the 500-report subset contained 59\%, 14\%, and 20\% of no change, improved, and worsened cases, respectively, compared with 59\%, 10\%, and 11\% in the full dataset. The remaining cases mainly involved support devices, with a small number unmentioned, as shown in Figure \ref{res:data_annotation}b. These distributions were highly similar between the subset and the full dataset.

Overall, these findings suggest that the 500-report subset reasonably reflects the distributional characteristics of the full L-MIMIC dataset and provides a representative test dataset for evaluating annotation quality.

\subsection{Performance of disease progression annotation across different categories} \label{app:diseaseperformance}

We evaluated the performance of Qwen2.5-32B on the disease progression task by stratifying model outputs according to radiological finding types. To enable a more structured and clinically meaningful analysis, all findings were grouped into nine common chest radiography categories, along with an additional “Other” category. The nine predefined categories include: (1) Pulmonary parenchymal / airspace abnormalities, (2) Interstitial and chronic lung disease, (3) Lung volume loss / atelectasis, (4) Pleural and pneumothorax, (5) Cardiac and pericardial abnormalities, (6) Vascular changes, (7) Mediastinum / hilum / thyroid, (8) Nodules / masses / tumors, and (9) Bone and chest wall.

To quantify performance across different types of findings, we also report category-level weighted averages. The number of samples (n) in each category is determined through manual annotated label. The result is shown in Table \ref{app:tabledisease}. Nevertheless, our error analysis (Figure \ref{fig:error}) indicates that the annotations are more strongly associated with progression-related keywords than with specific diseases; thus, the table is for reference only.

\begin{table*}[htbp]
\centering
\caption{Performance of Qwen on disease progression annotation tasks across grouped radiological findings.} \label{app:tabledisease}

\begin{tabular}{p{4.cm} p{5cm} p{1.5cm} p{1.5cm} p{2.5cm}}
\toprule
Group & Findings & $n_1$ & F1$_1$ & Weighted average \\
\midrule

\multirow{8}{4.5cm}{Pulmonary parenchymal / airspace abnormalities}
& airspace opacity & 21 & 82.9 & \multirow{8}{*}{79.4} \\
& consolidation & 33 & 69.7 & \\
& pneumonia & 61 & 63.9 & \\
& infiltration & 6 & 66.7 & \\
& alveolar hemorrhage & 8 & 100.0 & \\
& aspiration & 19 & 57.1 & \\
& lung opacity & 409 & 83.1 & \\
& opacity & 1 & 100.0 & \\
\midrule

\multirow{6}{4.5cm}{Interstitial and chronic lung disease}
& interstitial lung disease & 2 & 100.0 & \multirow{6}{*}{88.3} \\
& increased reticular markings/ILD pattern & 5 & 100.0 & \\
& pulmonary edema/hazy opacity & 88 & 86.9 & \\
& pleural/parenchymal scarring & 14 & 85.7 & \\
& bronchiectasis & 2 & 100.0 & \\
& copd/emphysema & 5 & 100.0 & \\
\midrule

\multirow{4}{4.5cm}{Lung volume loss / atelectasis}
& atelectasis & 102 & 74.5 & \multirow{4}{*}{79.3} \\
& lobar/segmental collapse & 13 & 84.6 & \\
& low lung volumes & 44 & 82.8 & \\
& elevated hemidiaphragm & 13 & 100.0 & \\
\midrule

\multirow{4}{4.5cm}{Pleural and pneumothorax}
& pleural effusion & 98 & 91.8 & \multirow{4}{*}{91.8} \\
& costophrenic angle blunting & 2 & 100.0 & \\
& hydropneumothorax & 3 & 100.0 & \\
& pneumothorax & 20 & 90.0 & \\
\midrule

\multirow{3}{4.5cm}{Cardiac and pericardial}
& enlarged cardiac silhouette & 95 & 90.4 & \multirow{3}{*}{89.4} \\
& pericardial effusion & 1 & 100.0 & \\
& fluid overload/heart failure & 9 & 77.8 & \\
\midrule

\multirow{4}{4.5cm}{Vascular changes}
& vascular congestion & 38 & 84.2 & \multirow{4}{*}{90.3} \\
& vascular redistribution & 1 & 100.0 & \\
& vascular calcification & 5 & 100.0 & \\
& tortuous aorta & 18 & 100.0 & \\
\midrule

\multirow{5}{4.5cm}{Mediastinum / hilum / thyroid}
& enlarged hilum & 22 & 86.4 & \multirow{5}{*}{89.6} \\
& mediastinal widening & 17 & 90.9 & \\
& mediastinal displacement & 20 & 87.2 & \\
& superior mediastinal mass/enlargement & 13 & 96.0 & \\
& goiter & 1 & 100.0 & \\
\midrule

\multirow{5}{4.5cm}{Nodules / masses / tumors}
& lung cancer & 3 & 66.7 & \multirow{5}{*}{93.3} \\
& lung lesion & 15 & 93.3 & \\
& mass/nodule (not otherwise specified) & 15 & 93.3 & \\
& calcified nodule & 8 & 100.0 & \\
& granulomatous disease & 4 & 100.0 & \\
\midrule

\multirow{6}{4.5cm}{Bone and chest wall}
& clavicle fracture & 7 & 71.4 &  \multirow{6}{*}{89.2} \\
& rib fracture & 11 & 81.8 & \\
& spinal fracture & 11 & 100.0 & \\
& scoliosis & 4 & 100.0 & \\
& spinal degenerative changes & 2 & 100.0 & \\
& shoulder osteoarthritis & 2 & 100.0 & \\
\midrule

\multirow{3}{4.5cm}{Other}
& hernia & 5 & 100.0 & \multirow{3}{*}{83.3} \\
& subcutaneous air & 2 & 50.0 & \\
& sub-diaphragmatic air & 1 & 100.0 & \\
\bottomrule
\end{tabular}
\end{table*}

\subsection{Typical errors of annotation} \label{app:error_example}

Tables \ref{tab:L_error} and \ref{tab:D_error} provide typical examples of each type of error.

\begin{table*}[htbp]
\caption{Typical errors of Qwen2.5-32B in the longitudinal annotation task.}\label{tab:L_error}
\begin{tabular}{p{11cm}p{1cm}p{1cm}p{3cm}}
\hline
Sentence & Y\_pred & Y\_true & Error type \\ 
\hline
No larger pleural effusions. & 0 & 1 & Inconsistencies \\ \hline
The extent of the pre-existing pleural effusion is constant. & 1 & 0 & Incorrect corrections \\ \hline
Bibasilar linear atelectasis is also noted. & 0 & 1 & Inconsistencies \\ \hline
There is improved positioning and as result, the right hilum appears normal. & 0 & 1 & True error \\ \hline
Minimal residual subcutaneous gas along the right lateral chest wall. & 0 & 1 & True error \\ \hline
Widespread parenchymal opacities appear to be progressed that might be the contribution of pulmonary edema, in particular in the right lateral lower lung and left mid lower lung. & 0 & 1 & True error \\ \hline
In addition, better visible than on the previous examination, a linear opacity at the bases of the right upper lobe that is better visible on the frontal than on the lateral film. & 1 & 0 & Incorrect corrections \\ \hline
Some left pleural effusion may be present as well, but the major change is atelectasis. & 0 & 1 & True error \\ \hline
In addition, there is a retrocardiac atelectasis as well as a parenchymal opacity at the right lung base, combined to air bronchograms and thickening of the peribronchial connective tissue in the appropriate clinical setting this change could reflect pneumonia. & 0 & 1 & Incorrect corrections \\ \hline
Nevertheless, there is a small amount of increased opacification overlying the lower thoracic spine, raising the possibility of consolidation at the left base posteriorly. & 0 & 1 & Inconsistencies \\ \hline
Possibility of slight increase in the degree of pulmonary vascular plethora would be difficult to exclude. & 0 & 1 & Inconsistencies \\ \hline
IMPRESSION:  Resolution of mild interstitial pulmonary edema. & 0 & 1 & True error \\ \hline
Vague opacity at the right base may be due to atelectasis or developing infiltrate. & 0 & 1 & Inconsistencies \\ \hline
Perihilar opacities, left greater than the right, are re-demonstrated. & 1 & 0 & Inconsistencies \\ \hline
\end{tabular}
\end{table*}

\begin{table*}[htbp]
\caption{Typical errors of Qwen2.5-32B in the disease progression annotation task. Unfilled Pred\_keyword and True\_keyword fields mean that they match.}\label{tab:D_error} 
\begin{tabular}{p{5.5cm}p{2.5cm}p{2cm}p{1.5cm}p{1.5cm}p{2cm}}
\hline
Sentence & Pred\_keyword & True\_keyword &Pred\_status & True\_status & Error type \\ 
\hline
FINDINGS:  As compared to the previous radiograph, the lung volumes have increased.& copd/emphysema&low lung \newline volumes&Worsened&Improved&Inconsistencies \\ \hline
Minimal residual subcutaneous gas along the right lateral chest wall.&- &- &No change&Improved&True error \\ \hline
Some left pleural effusion may be present as well, but the major change is atelectasis.& --& --&No change&Worsened&Inconsistencies \\ \hline
Possibility of slight increase in the
 degree of pulmonary vascular plethora would be difficult to exclude.& --& --&No change&Worsened&Inconsistencies \\ \hline
CONSOLIDATION IN THE RIGHT LOWER LOBE DEVELOPED BETWEEN \_\_\_ AND, COULD BE ACUTE PNEUMONIA.&-&-&Worsened&No change& Incorrect\newline corrections\\ \hline
The pre-existing distention of the stomach is no longer
 present.&low lung\newline volumes &abnormal &Unmentioned&Improved&Hallucination \\ \hline
Prominent left hilum, consistent with known perihilar
 non-small-cell lung cancer and lymphadenopathy, is grossly unchanged to slightly smaller.&-&-&Improved&No change& Inconsistencies\\ \hline
\end{tabular}
\end{table*}

\section{Report generation model configurations and full performance results}\label{app:full_results}

\subsection{Model configurations}\label{model_config}
Generated reports used in this study were obtained in three ways: (1) officially provided generated outputs, (2) running inference from provided model checkpoints, and (3) reimplementing models using the provided code. Table \ref{tab:model_sources} lists the official links, the sources of generated reports, and details of the model inputs. All experiments were run on one NVIDIA A100 GPU.

\begin{table*}[!t]

\centering
\caption{Summary of models, links, generated-report sources, image inputs, and text inputs. * indicates that, given the model’s large and diverse training corpus, only sources relevant to report generation are listed.}
\begin{tabular}{m{1.5cm}m{1.5cm}m{4cm}m{4cm}m{2cm}}
\toprule
\textbf{Models} & \textbf{Links} & \textbf{Generated-report sources} & \textbf{Image inputs} & \textbf{Text inputs} \\
\midrule
R2Gen & \href{https://github.com/cuhksz-nlp/R2Gen}{GitHub} & Checkpoint &Current frontal or lateral view& --\\
\midrule
L\_R2Gen& \href{https://github.com/CelestialShine/Longitudinal-Chest-X-Ray}{GitHub} & Reproduced results&Current frontal or lateral view, \newline previous frontal or lateral view& Previous report\\
\midrule
MLRG & \href{https://github.com/mk-runner/MLRG}{GitHub} & Released results&Current frontal and lateral view, \newline previous frontal view&Previous report,\newline indication\\
\midrule
HC-LLM & \href{https://github.com/TengfeiLiu966/HC-LLM}{GitHub} & Reproduced results &Current frontal or lateral view, \newline previous frontal or lateral view& Previous report\\
\midrule
Libra & \href{https://github.com/X-iZhang/Libra}{GitHub} &Checkpoint&Current frontal view, \newline previous frontal view&Indication,\newline history,\newline comparison,\newline technology\\
\midrule
Maira2* & \href{https://huggingface.co/microsoft/maira-2}{Hugging Face} & Checkpoint &Current frontal and lateral view, \newline previous frontal view&Previous report,\newline indication,\newline comparison,\newline technology\\
\midrule
MedVersa* & \href{https://huggingface.co/hyzhou/MedVersa_Internal}{Hugging Face} & Checkpoint &Current frontal or lateral view&Indication,\newline comparison \\
\bottomrule
\end{tabular}
\label{tab:model_sources}
\end{table*}

Prior to language-based evaluation, data are typically preprocessed to remove irrelevant information, which can substantially affect evaluation outcomes. To ensure consistency across models, we adopted the preprocessing procedures used in R2Gen for all models, to enable fair comparisons. Detailed preprocessing steps are provided in Table \ref{preprocess}.

\begin{table*}[htbp]
\centering

\caption{\label{preprocess}Text Data Preprocessing Strategy in R2Gen.}
\begin{tabular}{>{\raggedright\arraybackslash}p{3cm} >{\raggedright\arraybackslash}p{4cm} >{\raggedright\arraybackslash}p{6.5cm}}
\toprule
\textbf{Processing Steps} & \textbf{Description} & \textbf{Example} \\
\midrule
Remove line breaks & Replace all newline characters with spaces to form a single continuous text. &
\textbf{Original:} \texttt{"heart size normal\textbackslash nlungs clear"}\newline \textbf{Processed:} \texttt{"heart size normal lungs clear"} \\[4pt]
\midrule
Normalize underscores & Replace consecutive underscores with a single underscore. &
\textbf{Original:} \texttt{"right\_\_lung"}\newline \textbf{Processed:} \texttt{"right\_lung"} \\[4pt]
\midrule
Collapse multiple spaces & Replace consecutive spaces with a single space. &
\textbf{Original:} \texttt{"heart~~size~~normal"}\newline \textbf{Processed:} \texttt{"heart size normal"} \\[4pt]
\midrule
Normalize periods & Replace consecutive periods with a single period. &
\textbf{Original:} \texttt{"opacity.. noted"}\\
& & \textbf{Processed:} \texttt{"opacity. noted"} \\[4pt]
\midrule
Remove numbering & Remove common section or list numbers (e.g., ``1.'', ``2.''). &
\textbf{Original:} \texttt{"1. lungs clear. 2. heart normal."}\newline\textbf{Processed:} \texttt{"lungs clear. heart normal."} \\[4pt]
\midrule
Lowercasing & Convert all characters to lowercase for normalization. &
\textbf{Original:} \texttt{"Heart Size Normal"}\newline\textbf{Processed:} \texttt{"heart size normal"} \\[4pt]
\midrule
Sentence segmentation & Split text into sentences using ``. '' as the delimiter. &
\textbf{Original:} \texttt{"heart size normal. lungs clear."}\newline\textbf{Processed:} \texttt{["heart size normal", "lungs clear"]} \\[4pt]
\midrule
Remove punctuation & Remove punctuation and special characters. &
\textbf{Original:} \texttt{"no acute disease!"}\newline\textbf{Processed:} \texttt{"no acute disease"} \\[4pt]
\midrule
Trim whitespace & Remove leading and trailing spaces in each sentence. &
\textbf{Original:} \texttt{"  lungs clear  "}\newline\textbf{Processed:} \texttt{"lungs clear"} \\[4pt]
\midrule
Reconstruct report & Rejoin cleaned sentences using periods. &
\textbf{Original:} \texttt{["lungs clear", "heart normal"]}\newline\textbf{Processed:} \texttt{"lungs clear . heart normal ."} \\[4pt]
\bottomrule
\end{tabular}
\label{tab:clean_report_steps}
\end{table*}

\subsection{Full performance results \label{app:full_performance}}

Table \ref{table:evaluation_alldata} follows the same format as Table \ref{table:boot_evaluation}, except that the results for R2Gen, MedVersa, L\_R2Gen, MLRG, HC-LLM, and Maira2 are reported on the complete set of 1,863 reports, since these methods are capable of taking lateral radiographs as input.

\begin{table*}[!htbp]
\centering
\caption{Performance of report generation models on the whole L-MIMIC test set. P\_I, prior image input; P\_T, prior text input; A\_F1, micro-average F1 score; N\_F1, I\_F1, W\_F1, F1 scores for no-change, improved, and worsened classes,  respectively. Higher values are better for all indicators.}\label{table:evaluation_alldata}

\begin{tabular}{
    p{1.2cm}  
    >{\centering\arraybackslash}p{0.5cm}
    >{\centering\arraybackslash}p{0.5cm}
    >{\centering\arraybackslash}p{1.2cm} 
    >{\centering\arraybackslash}p{1.5cm}  
    >{\centering\arraybackslash}p{1.2cm} 
    >{\centering\arraybackslash}p{1.5cm}  
    >{\centering\arraybackslash}p{1.3cm}  
    >{\centering\arraybackslash}p{0.8cm} 
    >{\centering\arraybackslash}p{0.8cm} 
    >{\centering\arraybackslash}p{0.8cm} 
    >{\centering\arraybackslash}p{0.8cm}  
}
\toprule
\multirow{3}{*}{Models} 
& \multicolumn{2}{c}{Inputs}
& \multicolumn{4}{c}{Language-based metrics} 
& \multicolumn{5}{c}{Medical correctness-based metrics (\%)} \\
\cmidrule(lr){2-3}\cmidrule(lr){4-7} \cmidrule(lr){8-12}
&\multirow{2}{*}{P\_I} 
&\multirow{2}{*}{P\_T} & \multicolumn{2}{c}{Whole report} & \multicolumn{2}{c}{Longitudinal sentences} 
& Diagnosis & \multicolumn{4}{c}{Progression} \\
\cmidrule(lr){4-5} \cmidrule(lr){6-7} \cmidrule(lr){8-8} \cmidrule(lr){9-12}
&&& BLEU-4 & METEOR & BLEU-4 & METEOR & F1 & A\_F1 & N\_F1 & I\_F1 & W\_F1 \\
\midrule
R2Gen       &\ding{55}&\ding{55}& 9.3  & 13.3 & 2.4  & 5.9  & 32.5 & 46.1 & 55.6 & 0.5 & 5.0  \\
MedVersa    &\ding{55}&\ding{55}& 15.0 & 16.5 & 2.9  & 7.3  & 52.8 & 43.0 & 52.1 & 12.9 & 16.9  \\
L\_R2Gen    &\ding{51}&\ding{51}& 9.0  & 12.9 & 3.4  & 6.5  & 42.1 & 50.4 & 60.3 & 4.3 & 3.8  \\
MLRG        &\ding{51}&\ding{51}& 15.0 & 16.8 & 6.9  & 9.7  & 52.8 & 48.4 & 57.7 & 17.2 & 17.9  \\
HC-LLM      &\ding{51}&\ding{51}& 11.6 & 15.6 & 5.4  & 9.0  & 43.4 & 51.0 & 62.0 & 16.4 & 8.6   \\
Maira2      &\ding{51}&\ding{51}& 18.9 & 19.5 & 9.6  & 12.5 & \textbf{58.2} & 56.0 & \textbf{67.1} & \textbf{23.2} & \textbf{24.4} \\
Libra       &\ding{51}&\ding{55}& \textbf{23.6} & \textbf{22.6} & \textbf{12.7} & \textbf{13.9} & 57.6 & \textbf{57.1} & 66.5 & 22.4 & 24.3 \\
\bottomrule
\end{tabular}
\end{table*}

The following tables provide detailed evaluation results of the report generation models on the L-MIMIC dataset. Table \ref{tab:language} presents comparisons based on language-based metrics, including BLEU1-4, ROUGE-L, and METEOR. Table \ref{tab:clinical} summarizes model performance according to clinical efficacy metrics, including precision, recall, and F1-score for both disease diagnosis and overall disease progression information extraction. Table \ref{tab:class} reports detailed results of disease progression information extraction across individual classes, corresponding to three progression states: improved, no change, and worsened, reflecting the models’ performance on different disease progression states. Together, these tables provide a comprehensive overview of model performance from both linguistic and clinical perspectives.

\begin{table*}[!htbp]
\centering
\caption{Comparisons of model performance on the L-MIMIC dataset based on the language-based metric.
B1, B2, B3, B4, RL, ME denote BLEU-1, BLEU-2, BLEU-3, BLEU-4, ROUGE-L and METEOR, respectively. }
\label{tab:language}
\begin{threeparttable}

\begin{tabular}{%
    p{1.3cm}
    >{\centering\arraybackslash}p{0.6cm}
    >{\centering\arraybackslash}p{0.6cm}
    >{\centering\arraybackslash}p{0.6cm}
    >{\centering\arraybackslash}p{0.6cm}
    >{\centering\arraybackslash}p{0.6cm}
    >{\centering\arraybackslash}p{0.6cm}
    |
    >{\centering\arraybackslash}p{0.6cm}
    >{\centering\arraybackslash}p{0.6cm}
    >{\centering\arraybackslash}p{0.6cm}
    >{\centering\arraybackslash}p{0.6cm}
    >{\centering\arraybackslash}p{0.6cm}
    >{\centering\arraybackslash}p{0.6cm}
}
\toprule
\multirow{2}{*}{Paper} & \multicolumn{6}{c}{Whole report} & \multicolumn{6}{c}{Longitudinal sentences} \\
\cmidrule(lr){2-7} \cmidrule(lr){8-13}
 & B1 & B2 & B3 & B4 & RL & ME & B1 & B2 & B3 & B4 & RL & ME \\
\midrule
R2Gen         & 35.1 & 20.6 & 13.3 & 9.3 & 25.8 & 13.3 & 9.3 & 4.9 & 3.3 & 2.4 & 17.0 & 5.9 \\
MedVersa      & 39.0 & 26.6 & 19.6 & 15.0 & 30.8 & 16.5 & 7.5 & 4.9 & 3.7 & 2.9 & 18.3 & 7.3 \\
L\_R2Gen      & 31.5 & 19.3 & 12.8 & 9.0 & 26.1 & 12.9 & 11.6 & 6.6 & 4.5 & 3.4 & 18.3 & 6.5 \\
MLRG          & 39.4 & 26.5 & 19.4 & 15.0 & 31.6 & 16.8 & 17.6 & 11.5 & 8.7 & 6.9 & 22.6 & 9.7 \\
HC-LLM        & 40.2 & 24.8 & 16.5 & 11.6 & 27.2 & 15.6 & 19.6 & 11.4 & 7.5 & 5.4 & 19.7 & 9.0 \\
Maira2        & 43.2 & 30.9 & 23.7 & 18.9 & 35.1 & 19.5 & 23.9 & 16.1 & 12.1 & 9.6 & 26.0 & 12.5 \\
Libra         & \textbf{49.7} & \textbf{36.9} & \textbf{29.0} & \textbf{23.6} & \textbf{38.1} & \textbf{22.6} & \textbf{28.6} & \textbf{20.2} & \textbf{15.7} & \textbf{12.7} & \textbf{28.1} & \textbf{13.9} \\
\bottomrule
\end{tabular}
\end{threeparttable}
\end{table*}

\begin{table*}[!htbp]
\centering
\caption{Comparisons of report generation model performance on the L-MIMIC dataset based on clinical efficacy metrics.}
\label{tab:clinical}
\begin{threeparttable}

\begin{tabular}{%
    p{1.3cm}
    >{\centering\arraybackslash}p{1cm}
    >{\centering\arraybackslash}p{1cm}
    >{\centering\arraybackslash}p{1cm}
    |
    >{\centering\arraybackslash}p{1cm}
    >{\centering\arraybackslash}p{1cm}
    >{\centering\arraybackslash}p{1cm}
}
\toprule
\multirow{2}{*}{Paper} & \multicolumn{3}{c}{Disease diagnosis (\%)} & \multicolumn{3}{c}{Disease progression (\%)} \\
\cmidrule(lr){2-4} \cmidrule(lr){5-7}
 & Precision & Recall & F1 & Precision & Recall & F1 \\
\midrule
R2Gen        & 44.1 & 25.7 & 32.5 & 33.6 & 33.7 & 33.6 \\
MedVersa     & 59.1 & 47.7 & 52.8 & 35.3 & 32.5 & 33.9 \\
L\_R2Gen     & 54.3 & 34.4 & 42.1 & \textbf{38.5} & 39.0 & 38.8 \\
MLRG         & 59.0 & 47.9 & 52.8 & 33.0 & 39.0 & 35.7 \\
HC-LLM       & 50.6 & 38.0 & 43.4 & 34.4 & 42.5 & 38.0 \\
Maira2       & \textbf{61.5} & \textbf{55.3} & \textbf{58.2} & 35.2 & \textbf{50.4} & \textbf{41.4} \\
Libra        & 60.7 & 54.8 & 57.6 & 35.1 & 49.3 & 41.0 \\
\bottomrule
\end{tabular}
\end{threeparttable}
\end{table*}

\begin{table*}[!htbp]
\centering

\caption{Comparison of report generation model performance on the L-MIMIC dataset across different classes of disease progression metrics based on disease progression metrics.}
\label{apptab:class}
\begin{tabular}{
   p{1.3cm} 
>{\centering\arraybackslash}p{1cm}
>{\centering\arraybackslash}p{1cm}
>{\centering\arraybackslash}p{1cm} 
>{\centering\arraybackslash}p{1cm}
>{\centering\arraybackslash}p{1cm}
>{\centering\arraybackslash}p{1cm} 
>{\centering\arraybackslash}p{1cm}
>{\centering\arraybackslash}p{1cm}
>{\centering\arraybackslash}p{1cm} 
}
\toprule
\multirow{2}{*}{Paper} & \multicolumn{3}{c}{No change (\%)} & \multicolumn{3}{c}{Improved (\%)} & \multicolumn{3}{c}{Worsened (\%)} \\
\cmidrule(lr){2-4} \cmidrule(lr){5-7} \cmidrule(lr){8-10}
 & Precision & Recall & F1 & Precision & Recall & F1 & Precision & Recall & F1 \\
\midrule
R2Gen       & 44.9 & 44.3 & 44.6 & 0.2 & 0.2 & 0.2 & 2.8 & 2.7 & 2.7 \\
MedVersa    & 43.4 & 39.5 & 41.4 & 9.9 & 8.9 & 9.4 & 14.4 & 13.6 & 14.0 \\
L\_R2Gen    & \textbf{51.8} & 51.2 & \textbf{51.5} & 2.2 & 2.3 & 2.3 & 2.0 & 2.1 & 2.1 \\
MLRG        & 41.6 & 47.6 & 44.4 & 9.4 & 11.9 & 10.5 & 11.0 & 14.4 & 12.5 \\
HC-LLM      & 44.6 & 53.5 & 48.6 & 10.0 & 12.8 & 11.2 & 4.7 & 6.1 & 5.3 \\
Maira2      & 42.5 & \textbf{59.8} & 49.7 & \textbf{14.8} & \textbf{21.9} & \textbf{17.7} & \textbf{15.5} & \textbf{22.1} & \textbf{18.3} \\
Libra       & 43.8 & 59.5 & 50.5 & 11.7 & 17.1 & 13.9 & 12.5 & 18.4 & 14.9 \\
\bottomrule
\end{tabular}
\end{table*}

\subsection{Impact of LLM annotation errors on the performance rankings of report generation models}\label{app:rankings}

\begin{figure*}[htbp]
    \centering

    \begin{subfigure}[b]{0.48\textwidth}
        \centering
        \begin{overpic}[width=\linewidth]{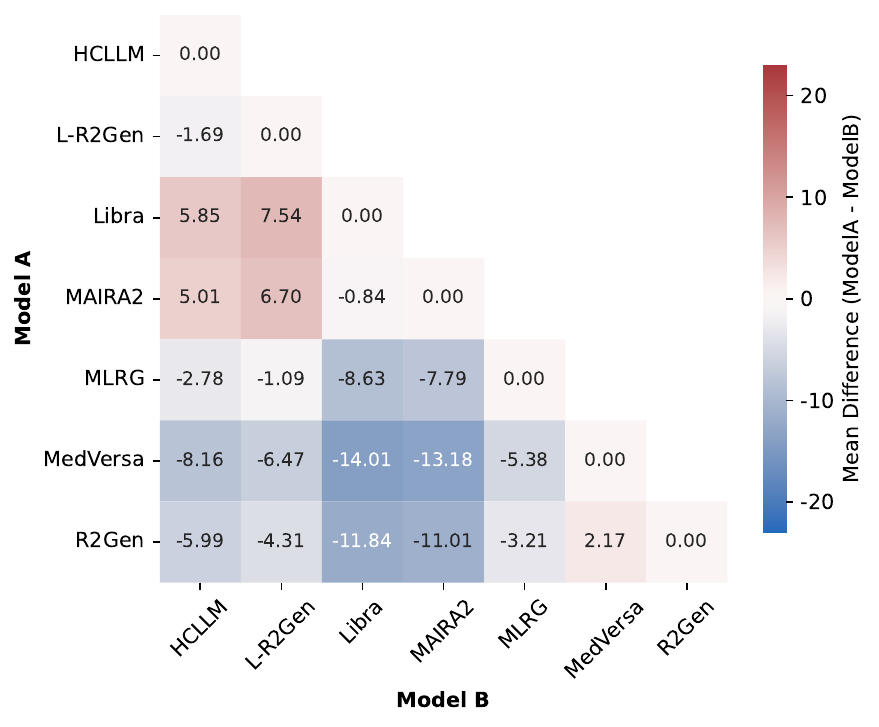}
            \put(-2,70){\textbf{a}}
        \end{overpic}
    \end{subfigure}
    \hfill
    \begin{subfigure}[b]{0.48\textwidth}
        \centering
        \begin{overpic}[width=\linewidth]{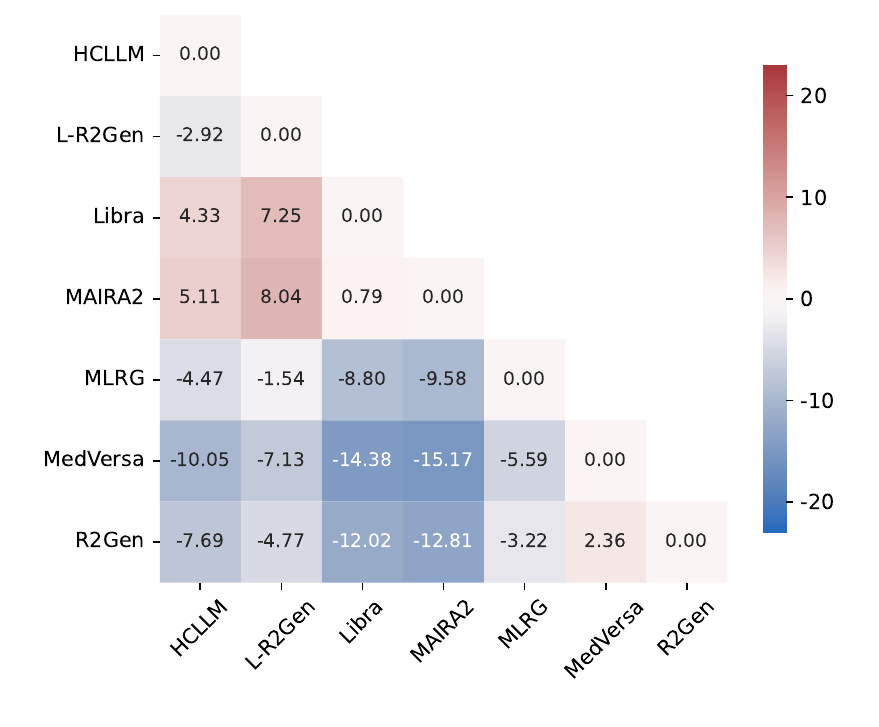}
            \put(-2,70){\textbf{b}}
        \end{overpic}
    \end{subfigure}

    \vspace{0.5em}

    \begin{subfigure}[b]{0.48\textwidth}
        \centering
        \begin{overpic}[width=\linewidth]{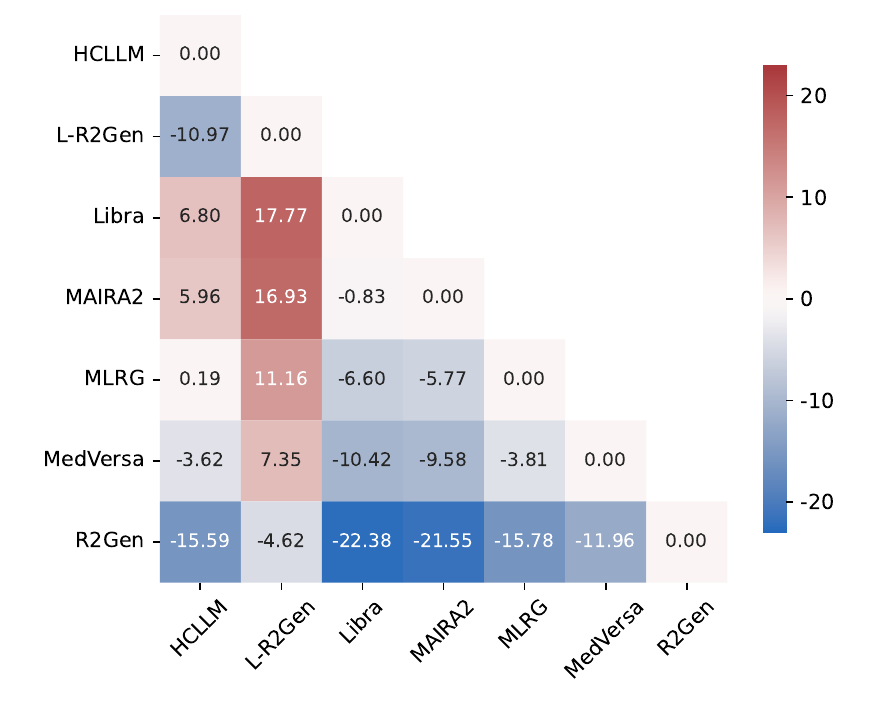}
            \put(-2,70){\textbf{c}}
        \end{overpic}
    \end{subfigure}
    \hfill
    \begin{subfigure}[b]{0.48\textwidth}
        \centering
        \begin{overpic}[width=\linewidth]{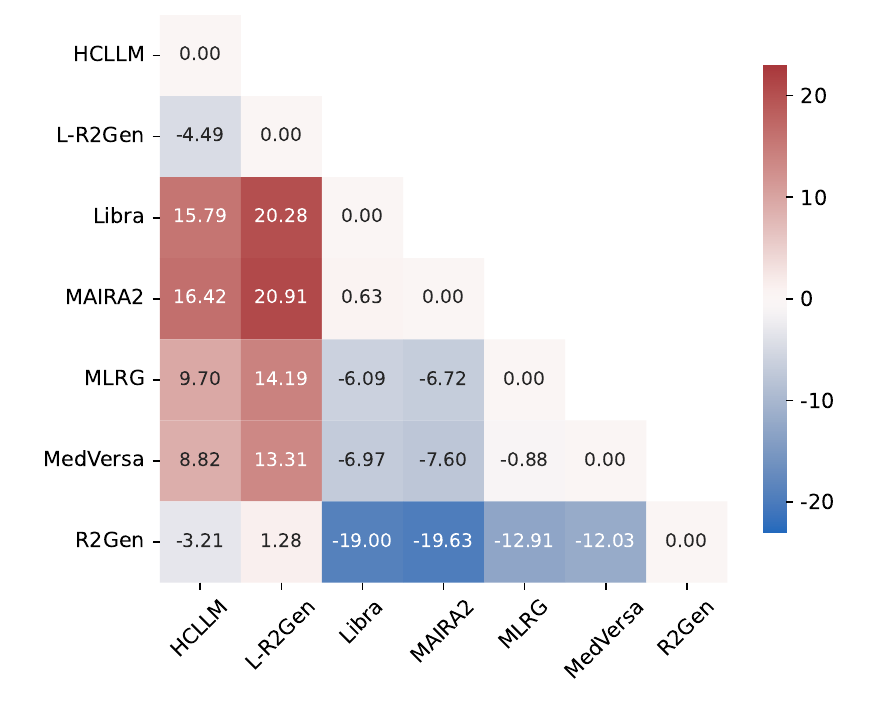}
            \put(-2,70){\textbf{d}}
        \end{overpic}
    \end{subfigure}

    \caption{Pairwise comparison of mean performance differences among seven automated report generation models. (a) micro-averaged F1. (b) F1 score for the no change class. (c) F1 score for the improved class. (d) F1 score for the worsened class.}
    \label{appfig:diff_heatmaps}
\end{figure*}

The results reported in Tables \ref{table:boot_evaluation}, \ref{tab:language}, \ref{tab:clinical}, and \ref{apptab:class}  were all obtained using the same LLM to evaluate both the ground-truth reports and the generated reports. Additionally, we supplemented the bootstrap analysis with pairwise comparisons based on mean performance differences (Figure \ref{appfig:diff_heatmaps}), and the results are consistent with the pairwise empirical win probabilities presented in Figure \ref{fig:winprob_heatmaps} of the main text. This setup may introduce bias stemming from the LLM's own preferences, as the annotations used for evaluation are generated by the same LLM, potentially favoring models whose outputs better align with the LLM's inherent preferences.

To investigate whether LLM-based annotation noise could affect the relative ranking of report generation models, we performed three complementary robustness analyses. These experiments were designed to evaluate (1) the stability of LLM annotations under stochastic inference, (2) the extent to which individual LLM annotator bias affects model ranking, and (3) the agreement between LLM-generated annotations and human expert annotations. 

\subsubsection{Robustness to stochastic annotation variability}

To verify whether the inherent stochasticity of LLM-based annotation affects model ranking stability, we conducted five independent evaluation runs across all seven baseline models under stochastic sampling settings (Section \ref{sec:med_decoding}). We evaluated the inter-run ranking consistency using Kendall's coefficient of concordance ($W$) for overall agreement and the average pairwise Spearman rank correlation coefficient ($\bar{\rho}$) across all $\binom{5}{2} = 10$ pairs of runs. As reported in Table~\ref{apptab:5run}, our evaluation framework demonstrated exceptional stability across all metrics. 

Specifically, Kendall's $W$ reached 1.0 for most metrics, indicating complete agreement in model rankings across different evaluation runs. The corresponding significance tests showed that all agreement coefficients were statistically significant ($p<0.001$), confirming that the observed ranking consistency was unlikely to occur by chance. For the class-specific medical correctness metrics, the agreement remained highly consistent, with $W$ and $\bar{\rho}$ values exceeding 0.95. These results demonstrate that the stochasticity introduced by LLM-based annotation has a negligible effect on model ranking stability and support the robustness and reproducibility of our evaluation framework.

\begin{table*}[!htbp]
\centering

\caption{Comparison of report generation model performance on 400 randomly sampled reports across longitudinal metrics over five evaluation runs (mean $\pm$ SD). B4, RL, and ME denote BLEU-4, ROUGE-L, and METEOR, respectively. A\_F1, micro-average F1 score; N\_F1, I\_F1, and W\_F1, F1 scores for the no-change, improved, and worsened classes, respectively. $W$ and average $\bar{\rho}$ denote Kendall's coefficient of concordance and the average Spearman rank correlation coefficient across the five runs, respectively. $^{***}$ indicate statistical significance at 
$p<0.001$.}
\label{apptab:5run}

\begin{tabular}{
  p{1.8cm} 
>{\centering\arraybackslash}p{1.7cm}
>{\centering\arraybackslash}p{1.7cm}
>{\centering\arraybackslash}p{1.7cm}
>{\centering\arraybackslash}p{1.7cm}
>{\centering\arraybackslash}p{1.7cm}
>{\centering\arraybackslash}p{1.7cm}
>{\centering\arraybackslash}p{1.7cm}
}
\toprule
\multirow{2}{*}{Paper} & \multicolumn{3}{c}{Language-based metrics}  & \multicolumn{4}{c}{Medical correctness-based metrics (\%)} \\
\cmidrule(lr){2-4} \cmidrule(lr){5-8}
 & B4 & RL & ME & A\_F1 & N\_F1 & I\_F1 & W\_F1 \\
\midrule
R2Gen     & 2.8 $\pm$ 0.0 & 17.5 $\pm$ 0.0 & 6.1 $\pm$ 0.0  & 46.6 $\pm$ 0.3 & 56.6 $\pm$ 0.3 & 0.0 $\pm$ 0.0  & 3.9 $\pm$ 0.0 \\
MedVersa  & 3.0 $\pm$ 0.0 & 17.9 $\pm$ 0.0 & 7.2 $\pm$ 0.0  & 41.5 $\pm$ 0.4 & 49.8 $\pm$ 0.4 & 14.9 $\pm$ 0.1 & 20.1 $\pm$ 0.7 \\
L\_R2Gen  & 4.3 $\pm$ 0.0 & 19.1 $\pm$ 0.0 & 7.1 $\pm$ 0.0  & 50.5 $\pm$ 0.5 & 60.8 $\pm$ 0.4 & 0.3 $\pm$ 0.8  & 9.3 $\pm$ 0.9 \\
MLRG      & 7.7 $\pm$ 0.0 & 23.0 $\pm$ 0.0 & 10.2 $\pm$ 0.0 & 44.0 $\pm$ 0.4 & 53.7 $\pm$ 0.6 & 15.7 $\pm$ 1.0 & 18.0 $\pm$ 0.6 \\
HC-LLM    & 6.9 $\pm$ 0.0 & 20.7 $\pm$ 0.0 & 10.0 $\pm$ 0.0 & 49.2 $\pm$ 0.4 & 61.5 $\pm$ 0.4 & 14.5 $\pm$ 0.6 & 5.9 $\pm$ 0.7 \\
Maira2    & 9.8 $\pm$ 0.0 & 26.0 $\pm$ 0.1 & 12.1 $\pm$ 0.0 & 53.5 $\pm$ 0.3 & 62.3 $\pm$ 0.4 & 32.0 $\pm$ 1.4 & 26.9 $\pm$ 0.8 \\
Libra     & 13.7 $\pm$ 0.0 & 28.2 $\pm$ 0.0 & 14.0 $\pm$ 0.0 & 54.4 $\pm$ 0.4 & 63.9 $\pm$ 0.4 & 22.6 $\pm$ 0.5 & 20.6 $\pm$ 0.8 \\
\midrule
$W$  & 1.0$^{***}$ & 1.0$^{***}$ & 1.0$^{***}$ & 1.0$^{***}$ & 1.0$^{***}$ & 0.962$^{***}$ & 0.989$^{***}$ \\
Avg. $\bar{\rho}$ & 1.0$^{***}$ & 1.0$^{***}$ & 1.0$^{***}$ & 1.0$^{***}$ & 1.0$^{***}$ & 0.953$^{***}$ & 0.986$^{***}$ \\
\bottomrule
\end{tabular}
\end{table*}

\subsubsection{Inter-LLM annotation consistency}

A potential concern of LLM-based evaluation is that a single LLM annotator may introduce model-specific preferences or systematic biases, which could affect the relative ranking of report generation models. To assess the impact of such biases, given the strong performance of Gemini 3 Flash (Section \ref{app:gemini}), we replaced the ground-truth annotator from Qwen2.5-32B with Gemini Flash 3 while keeping the generated report annotations unchanged. Since the generated reports from all models were still evaluated under the same Qwen-based criteria, any substantial change in model ranking would suggest that the original evaluation was sensitive to the interpretation bias of the ground-truth annotator. Conversely, high ranking consistency between the two annotation settings would indicate that the evaluation framework is robust to potential biases from an individual LLM annotator.

To quantify the consistency between the two annotation settings, we evaluated all report generation models under both settings using conventional language-generation metrics (BLEU-4, ROUGE-L, and METEOR) as well as medical correctness metrics, including the micro-averaged F1 score and class-specific F1 scores for the no-change, improved, and worsened categories, based on the longitudinal descriptions. Since our primary interest is whether changing the LLM annotator affects the relative ordering of report generation models rather than their absolute metric values, we further computed Spearman's rank correlation coefficient ($\rho$) and Kendall's rank correlation coefficient ($\tau$) to measure the agreement between model rankings under the two annotation settings.

Table \ref{tab:class} summarizes the evaluation results under the two annotation settings. Although replacing the ground-truth annotator from Gemini Flash 3 to Qwen2.5-32B leads to slight changes in the absolute values of both language-generation and medical correctness metrics, the relative ranking of report generation models remains highly consistent. Specifically, all language-generation metrics (BLEU-4, ROUGE-L, and METEOR) achieve perfect rank agreement, with Spearman's rank correlation coefficient ($\rho$) and Kendall's rank correlation coefficient ($\tau$) both equal to 1.000 ($p<0.001$). Similarly, the medical correctness metrics also exhibit strong ranking consistency, with $\rho$ ranging from 0.964 to 1.000 (all $p<0.001$) and $\tau$ ranging from 0.905 to 1.000 (all $p\le0.01$). The statistically significant correlations indicate that the observed ranking agreement is unlikely to have occurred by chance, while the consistently high values of $\rho$ and $\tau$ demonstrate that changing the LLM annotator has little effect on the relative ordering of report generation models. The biases introduced by different LLM annotators are unlikely to significantly affect the relative ranking of report generation models.

\begin{table*}[!htbp]
\centering

\caption{Performance of report generation models on 400 randomly sampled reports, evaluated using longitudinal metrics. G and Q denote ground-truth reports annotated by Gemini Flash 3 and Qwen2.5-32B, respectively; all generated reports are annotated by Qwen2.5-32B. B4, RL, and ME denote BLEU-4, ROUGE-L, and METEOR, respectively. A\_F1 denotes the micro-averaged F1 score, while N\_F1, I\_F1, and W\_F1 denote the F1 scores for the no-change, improved, and worsened classes, respectively. Numbers in parentheses indicate model rankings for each metric. $\rho$ and $\tau$ denote Spearman's and Kendall's rank correlation coefficients, respectively. $^{***}$ and $^{**}$ indicate statistical significance at 
$p<0.001$ and $p<0.01$, respectively.}
\label{tab:class}

\begin{tabular}{
   p{1.2cm} 
>{\centering\arraybackslash}p{0.7cm}
>{\centering\arraybackslash}p{0.7cm}
>{\centering\arraybackslash}p{0.7cm}
>{\centering\arraybackslash}p{0.7cm}
>{\centering\arraybackslash}p{0.7cm}
>{\centering\arraybackslash}p{0.7cm}
>{\centering\arraybackslash}p{0.7cm}
>{\centering\arraybackslash}p{0.7cm}
>{\centering\arraybackslash}p{0.7cm}
>{\centering\arraybackslash}p{0.7cm}
>{\centering\arraybackslash}p{0.7cm}
>{\centering\arraybackslash}p{0.7cm}
>{\centering\arraybackslash}p{0.7cm}
>{\centering\arraybackslash}p{0.7cm}
}
\toprule
\multirow{3}{*}{Paper} & \multicolumn{6}{c}{Language-based metrics}  & \multicolumn{8}{c}{Medical correctness-based metrics (\%)} \\
\cmidrule(lr){2-7} \cmidrule(lr){8-15}
 & \multicolumn{2}{c}{B4} & \multicolumn{2}{c}{RL} & \multicolumn{2}{c}{ME} & \multicolumn{2}{c}{A\_F1} & \multicolumn{2}{c}{N\_F1} & \multicolumn{2}{c}{I\_F1} & \multicolumn{2}{c}{W\_F1} \\
\cmidrule(lr){2-3} \cmidrule(lr){4-5}\cmidrule(lr){6-7} \cmidrule(lr){8-9}\cmidrule(lr){10-11} \cmidrule(lr){12-13}\cmidrule(lr){14-15}
 &G&Q&G&Q&G&Q&G&Q&G&Q&G&Q&G&Q \\
\midrule
R2Gen &2.0(7)&2.8(7)&16.4(7)&17.5(7)&5.7(7)&6.1(7)&45.8(5)&46.7(5)&57.4(5)&56.7(5)&0.0(7)&0.0(6)&4.2(7)&3.9(7) \\ MedVersa &2.0(7)&3.0(6)&16.8(6)&17.9(6)&6.7(5)&7.2(5)&40.7(7)&41.2(7)&50.4(7)&49.5(7)&13.3(5)&14.8(5)&19.6(2)&20.6(2) \\ L\_R2Gen &3.1(5)&4.2(5)&17.8(5)&19.1(5)&6.5(6)&7.1(6)&49.3(4)&50.7(4)&60.5(4)&60.9(4)&4.5(6)&0.0(6)&12.3(5)&10.4(5) \\ MLRG &6.1(3)&7.7(3)&22.0(3)&23.0(3)&9.5(3)&10.2(3)&41.8(6)&44.0(6)&52.3(6)&53.5(6)&16.1(3)&16.8(3)&16.5(4)&17.9(4) \\ HC-LLM &5.8(4)&6.9(4)&20.1(4)&20.7(4)&9.3(4)&10.0(4)&48.9(5)&49.3(5)&62.6(2)&61.4(3)&14.8(4)&15.1(4)&7.2(6)&6.6(6) \\ Maira2 &8.0(2)&9.8(2)&25.0(2)&26.1(2)&11.4(2)&12.1(2)&51.2(2)&53.3(2)&60.9(3)&62.3(2)&28.2(1)&31.2(1)&28.0(1)&26.2(1) \\ Libra &11.7(1)&13.6(1)&26.9(1)&28.2(1)&13.1(1)&14.0(1)&52.5(1)&53.6(1)&63.1(1)&63.5(1)&25.3(2)&21.6(2)&18.0(3)&18.5(3) \\
\bottomrule
$\rho$ 
& \multicolumn{2}{c}{1.000$^{***}$} 
& \multicolumn{2}{c}{1.000$^{***}$} 
& \multicolumn{2}{c}{1.000$^{***}$} 
& \multicolumn{2}{c}{1.000$^{***}$} 
& \multicolumn{2}{c}{0.964$^{***}$} 
& \multicolumn{2}{c}{0.991$^{***}$} 
& \multicolumn{2}{c}{1.000$^{***}$} \\

$\tau$ 
& \multicolumn{2}{c}{1.000$^{***}$} 
& \multicolumn{2}{c}{1.000$^{***}$} 
& \multicolumn{2}{c}{1.000$^{***}$} 
& \multicolumn{2}{c}{1.000$^{***}$} 
& \multicolumn{2}{c}{0.905$^{**}$} 
& \multicolumn{2}{c}{0.976$^{**}$} 
& \multicolumn{2}{c}{1.000$^{***}$} \\
\bottomrule
\end{tabular}
\end{table*}

Furthermore, as discussed in Section \ref{sec:error_ana}, many annotation errors stem from ambiguous or insufficient contextual information. In such cases, using the same model to annotate both GT and generated reports may improve annotation consistency by ensuring more consistent interpretation criteria across the evaluation process.

\subsubsection{Agreement between LLM and human annotations}

To address the concern that LLM-based evaluation might inherently favor reports with LLM-friendly phrasing over those with genuine clinical accuracy, we conducted an independent human validation study. A randomly selected subset of 100 reports was independently annotated by clinical experts following the exact same protocol used by the Qwen2.5-32B evaluator.

We compared the model rankings derived from human annotations against those from the LLM using Spearman's ($\rho$) and Kendall's ($\tau$) rank correlation coefficients. As shown in Table~\ref{tab:human}, model rankings evaluated by Qwen2.5-32B correlated positively and statistically significantly with human clinical judgments across all evaluated metrics. 

For language-based metrics, human and LLM evaluators demonstrated high ranking alignment ($\rho \ge 0.964$, $\tau \ge 0.905$, $p < 0.01$), with BLEU-4 and ROUGE-L yielding identical model rankings ($\rho = 1.000$, $\tau = 1.000$, $p < 0.001$). For medical correctness-based metrics, the rank correlations remained positive and statistically significant ($p < 0.05$), with $\rho$ ranging from $0.847$ to $0.990$ and $\tau$ from $0.683$ to $0.972$. Specifically, high rank consistency was observed for the improved class ($I\_F1$: $\rho = 0.990$, $\tau = 0.972$, $p < 0.01$) and no-change class ($N\_F1$: $\rho = 0.929$, $\tau = 0.810$, $p < 0.05$), whereas a moderate correlation was obtained for the worsened class ($W\_F1$: $\rho = 0.847$, $\tau = 0.683$, $p < 0.05$). 

Furthermore, it is important to note that automated annotation methods, including both rule-based approaches and conventional deep learning models, may introduce varying degrees of errors and biases. Consequently, the model rankings produced by these methods may not perfectly align with human-annotated rankings. This issue is not exclusive to LLM-based models; nevertheless, the use of automated tools remains highly beneficial for evaluation tasks. Given that manual annotation is prohibitively labor-intensive, these tools offer critical advantages in efficiency and scalability, especially when handling the large volume of reports generated during iterative model development. Previous longitudinal annotation approaches either relied on overly simplistic predefined rules (e.g., TEM \cite{1bannur2023learning}) or failed to release the complete annotation pipeline (e.g., ImaGenome Silver \cite{ImaGenomewu2021chest}), limiting their utility and reproducibility. The primary goal of this work is to propose a flexible and extensible LLM-based framework for longitudinal information annotation. Consequently, our proposed framework is compatible with multiple LLMs and can be readily adapted as foundation models evolve.

The bias of large language models and the differences between them and human evaluations have become a heavily investigated research topic \cite{zheng2023judging}. While this is frequently discussed as a standalone subject in the literature, we dedicate this chapter to a preliminary investigation into the consistency between LLM-based annotations and human evaluations.

\begin{table*}[!htbp]
\centering

\caption{Performance comparison of report generation models on 100 randomly sampled reports using longitudinal evaluation metrics. H/Q denotes evaluations provided by human annotators and Qwen2.5-32B, respectively, for both ground-truth and generated reports. B4, RL, and ME denote BLEU-4, ROUGE-L, and METEOR, respectively. A\_F1 denotes the micro-average F1 score; N\_F1, I\_F1, and W\_F1 denote the F1 scores for the no-change, improved, and worsened classes, respectively. Numbers in parentheses indicate model rankings for each metric. $\rho$ and $\tau$ denote Spearman's and Kendall's rank correlation coefficients between H and Q, respectively. $^{***}$, $^{**}$, and $^{*}$ indicate statistical significance at 
$p<0.001$, $p<0.01$, and $p<0.05$, respectively.}
\label{tab:human}

\begin{tabular}{
   p{1.3cm} 
>{\centering\arraybackslash}p{0.7cm}
>{\centering\arraybackslash}p{0.7cm}
>{\centering\arraybackslash}p{0.7cm}
>{\centering\arraybackslash}p{0.7cm}
>{\centering\arraybackslash}p{0.7cm}
>{\centering\arraybackslash}p{0.7cm}
>{\centering\arraybackslash}p{0.7cm}
>{\centering\arraybackslash}p{0.7cm}
>{\centering\arraybackslash}p{0.7cm}
>{\centering\arraybackslash}p{0.7cm}
>{\centering\arraybackslash}p{0.7cm}
>{\centering\arraybackslash}p{0.7cm}
>{\centering\arraybackslash}p{0.7cm}
>{\centering\arraybackslash}p{0.7cm}
}
\toprule
\multirow{3}{*}{Paper} & \multicolumn{6}{c}{Language-based metrics}  & \multicolumn{8}{c}{Medical correctness-based metrics (\%)} \\
\cmidrule(lr){2-7} \cmidrule(lr){8-15}
 & \multicolumn{2}{c}{B4} & \multicolumn{2}{c}{RL} & \multicolumn{2}{c}{ME} & \multicolumn{2}{c}{A\_F1} & \multicolumn{2}{c}{N\_F1} & \multicolumn{2}{c}{I\_F1} & \multicolumn{2}{c}{W\_F1} \\
\cmidrule(lr){2-3} \cmidrule(lr){4-5}\cmidrule(lr){6-7} \cmidrule(lr){8-9}\cmidrule(lr){10-11} \cmidrule(lr){12-13}\cmidrule(lr){14-15}
 &H&Q&H&Q&H&Q&H&Q&H&Q&H&Q&H&Q \\
\midrule
R2Gen & 0.7(7) & 0.7(7) & 12.9(7) & 13.5(7) & 4.0(7) & 4.0(7) & 42.2(5) & 42.3(7) & 51.0(6) & 50.0(7) & 0.0(5) & 0.0(5) & 12.5(4) & 16.7(4) \\ 
MedVersa & 1.1(6) & 1.7(6) & 15.3(6) & 16.5(6) & 5.8(6) & 6.3(5) & 36.4(7) & 43.1(6) & 45.2(7) & 51.4(6) & 15.0(2) & 11.8(2) & 9.3(5) & 22.2(3) \\ 
L\_R2Gen & 2.4(5) & 2.2(5) & 15.8(5) & 16.9(5) & 6.1(5) & 5.8(6) & 42.6(3) & 46.1(3) & 51.9(4) & 55.3(4) & 0.0(5) & 0.0(5) & 6.3(6) & 0.0(6) \\ 
MLRG & 4.9(3) & 5.2(3) & 20.0(3) & 20.8(3) & 8.7(3) & 8.6(3) & 42.0(6) & 43.5(5) & 51.2(5) & 53.1(5) & 6.1(4) & 6.5(4) & 17.0(3) & 11.1(5) \\ 
HC-LLM & 4.2(4) & 5.1(4) & 17.8(4) & 18.3(4) & 7.7(4) & 8.1(4) & 42.6(4) & 45.8(4) & 54.1(3) & 58.2(3) & 0.0(5) & 0.0(5) & 0.0(7) & 0.0(6) \\ 
Maira2 & 8.6(2) & 9.0(2) & 23.9(2) & 24.0(2) & 11.7(2) & 11.7(2) & 52.9(1) & 59.3(1) & 62.1(2) & 69.3(1) & 32.7(1) & 32.6(1) & 26.7(1) & 29.6(1) \\ 
Libra & 11.5(1) & 11.9(1) & 27.0(1) & 27.8(1) & 12.7(1) & 12.9(1) & 51.4(2) & 50.8(2) & 62.6(1) & 60.1(2) & 11.1(3) & 11.8(2) & 20.0(2) & 25.0(2) \\
\midrule
$\rho$ & \multicolumn{2}{c}{1.000$^{***}$} & \multicolumn{2}{c}{1.000$^{***}$} & \multicolumn{2}{c}{0.964$^{***}$} & \multicolumn{2}{c}{0.893$^{**}$} & \multicolumn{2}{c}{0.929$^{**}$} & \multicolumn{2}{c}{0.990$^{***}$} & \multicolumn{2}{c}{0.847$^{*}$} \\
$\tau$ & \multicolumn{2}{c}{1.000$^{***}$} & \multicolumn{2}{c}{1.000$^{***}$} & \multicolumn{2}{c}{0.905$^{**}$} & \multicolumn{2}{c}{0.810$^{*}$} & \multicolumn{2}{c}{0.810$^{*}$} & \multicolumn{2}{c}{0.972$^{**}$} & \multicolumn{2}{c}{0.683$^{*}$} \\
\bottomrule
\end{tabular}
\end{table*}

\end{appendices}
\end{document}